\documentclass[journal, compsoc]{IEEEtran}

\usepackage{url}            
\usepackage{amsmath,amsfonts,amssymb}       
\usepackage{multirow}
\usepackage{algorithm,algorithmic}
\usepackage{xcolor}
\usepackage[normalem]{ulem}
\usepackage{arydshln}
\usepackage{caption}
\usepackage{enumitem}
\usepackage{wrapfig}
\usepackage{microtype}
\usepackage{graphicx}
\usepackage{subfigure}
\usepackage{booktabs} 
\usepackage{cutwin}
\usepackage{bbm}
\usepackage[pangram]{blindtext}
\usepackage[calc]{adjustbox}

\usepackage{tabularx}
\usepackage{color, colortbl}
\usepackage{kotex}
\usepackage{lipsum}

\usepackage{enumitem} 

\makeatletter
\def\hlinewd#1{%
	\noalign{\ifnum0=`}\fi\hrule \@height #1 %
	\futurelet\reserved@a\@xhline}
\makeatother

\renewcommand{\algorithmiccomment}[1]{\bgroup\hfill$\triangleright$~#1\egroup}

\definecolor{crimson}{rgb}{0.86, 0.08, 0.24}
\definecolor{orange-red}{rgb}{1.0, 0.27, 0.0}

\newcommand{\bsy}{\boldsymbol}

\newcommand{\newor}{%
  \mathbin{%
    {\vee}\mspace{-2.9mu}
  }%
}
\usepackage{bm}
\newcommand{\bs}[1] {\bm{#1}}

\usepackage{svg}
\DeclareMathOperator*{\minimize}{minimize}
\DeclareMathOperator*{\logicalor}{\vee}

\usepackage{gensymb}
\usepackage{marvosym}

\usepackage{mathtools,tikz}
\DeclareRobustCommand\sampleline[1]{%
  \tikz\draw[#1] (0,0) (0,\the\dimexpr\fontdimen22\textfont2\relax)
  -- (2em,\the\dimexpr\fontdimen22\textfont2\relax);%
}

\definecolor{Red}{rgb}{1,0,0}
\definecolor{Blue}{rgb}{0,0,0.8}
\definecolor{Green}{rgb}{0,0.7,0.2}
\definecolor{airforceblue}{rgb}{0.36, 0.54, 0.66}
\definecolor{ao(english)}{rgb}{0.0, 0.5, 0.0}
\definecolor{azure(colorwheel)}{rgb}{0.0, 0.5, 1.0}
\definecolor{crimson}{rgb}{0.86, 0.08, 0.24}
\definecolor{darkcerulean}{rgb}{0.03, 0.27, 0.49}
\definecolor{cobalt}{rgb}{0.0, 0.28, 0.67}
\definecolor{rosegold}{rgb}{0.72, 0.43, 0.47}
\definecolor{orange-red}{rgb}{1.0, 0.27, 0.0}
\definecolor{mountainmeadow}{rgb}{0.19, 0.73, 0.56}
\definecolor{malachite}{rgb}{0.04, 0.85, 0.32}
\definecolor{darkblue}{rgb}{0.0, 0.0, 0.55}

\definecolor{customblue}{rgb}{0.2, 0.35, 0.8}
\definecolor{gg}{gray}{0.9}

\usepackage[capitalize]{cleveref}
\crefname{section}{Sec.}{Secs.}
\Crefname{section}{Section}{Sections}
\Crefname{table}{Table}{Tables}
\crefname{table}{Tab.}{Tabs.}

\usepackage{makecell}

\newcommand{\eat}[1]{{}}
\creflabelformat{equation}{#2\textup{#1}#3}  

\begin{document}
%
\title{Forget-free Continual Learning with Soft-Winning SubNetworks}

\author{Haeyong~Kang\thanks{Email: haeyong.kang@kaist.ac.kr},
        Jaehong~Yoon, 
        Sultan Rizky Madjid, 
        Sung Ju Hwang,
        and~Chang~D.~Yoo$^{\ast}$\thanks{$^\ast$ Corresponding Author.} 
}

\markboth{Preprint, March~2023}
{Shell \MakeLowercase{\textit{et al.}}: Bare Demo of IEEEtran.cls for Computer Society Journals}

\IEEEtitleabstractindextext{%
\begin{abstract}
Inspired by \emph{Regularized Lottery Ticket Hypothesis (RLTH)}, which states that competitive smooth (non-binary) subnetworks exist within a dense network in continual learning tasks, we investigate two proposed architecture-based continual learning methods which sequentially learn and select adaptive binary- (WSN) and non-binary Soft-Subnetworks (SoftNet) for each task. WSN and SoftNet jointly learn the regularized model weights and task-adaptive non-binary masks of subnetworks associated with each task whilst attempting to select a small set of weights to be activated (winning ticket) by reusing weights of the prior subnetworks. Our proposed WSN and SoftNet are inherently immune to catastrophic forgetting as each selected subnetwork model does not infringe upon other subnetworks in Task Incremental Learning (TIL). In TIL, binary masks spawned per winning ticket are encoded into one N-bit binary digit mask, then compressed using Huffman coding for a sub-linear increase in network capacity to the number of tasks. Surprisingly, in the inference step, SoftNet generated by injecting small noises to the backgrounds of acquired WSN (holding the foregrounds of WSN) provides excellent forward transfer power for future tasks in TIL. SoftNet shows its effectiveness over WSN in regularizing parameters to tackle the overfitting, to a few examples in Few-shot Class Incremental Learning (FSCIL). 
\end{abstract}

}

\maketitle

\IEEEdisplaynontitleabstractindextext

\IEEEpeerreviewmaketitle

\section{Introduction}

\IEEEPARstart{C}{ontinual} Learning (CL), also known as Lifelong Learning \cite{ThrunS1995,rusu2016progressive,zenke2017continual, hassabis2017neuroscience}, is a learning paradigm where a series of tasks are learned sequentially. The principle objective of continual learning is to replicate human cognition, characterized by the ability to learn new concepts or skills incrementally throughout one's lifespan. An optimal continual learning system could facilitate a positive forward and backward transfer, leveraging the knowledge gained from previous tasks to solve new ones while also updating its understanding of previous tasks with the new knowledge. However, building successful continual learning systems is challenging due to the occurrence of \textit{catastrophic forgetting} or \textit{catastrophic interference}~\cite{McCloskey1989}, a phenomenon where the performance of the model on previous tasks significantly deteriorates when it learns new tasks. This can make it challenging to retain the knowledge acquired from previous tasks, ultimately leading to a decline in overall performance. To tackle the catastrophic forgetting problem in continual learning, numerous approaches have been proposed, which can be broadly classified as follows: (1) \textbf{Regularization-based methods}~\cite{Kirkpatrick2017, chaudhry2020continual, Jung2020, titsias2019functional, mirzadeh2020linear} aim to keep the learned information of past tasks during continual training aided by sophisticatedly designed regularization terms, (2) \textbf{Rehearsal-based methods}~\cite{rebuffi2017icarl, riemer2018learning, chaudhry2018efficient, chaudhry2019continual, Saha2021} utilize a set of real or synthesized data from the previous tasks and revisit them, and (3) \textbf{Architecture-based methods}~\cite{mallya2018piggyback, Serra2018, li2019learn, wortsman2020supermasks, kang2022forget, kang2022soft} propose to minimize the inter-task interference via newly designed architectural components. 

\begin{figure*}[ht]
    \centering
    \setlength{\tabcolsep}{-0pt}{%
    \begin{tabular}{cccc}
    \hspace{0.0in}
    \includegraphics[height=4.3cm, trim={0.1cm 0.02cm 0.1cm 0.08cm},clip]{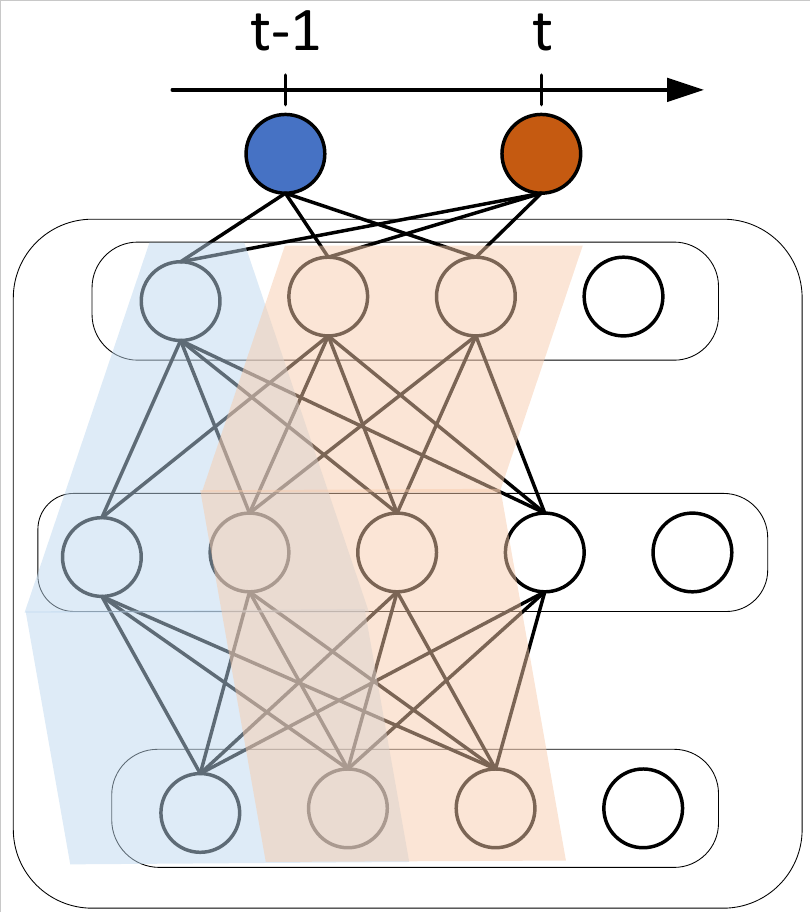} &\hspace{0.2in}
    \includegraphics[height=4.3cm, trim={0.1cm 0.02cm 0.1cm 0.08cm},clip]{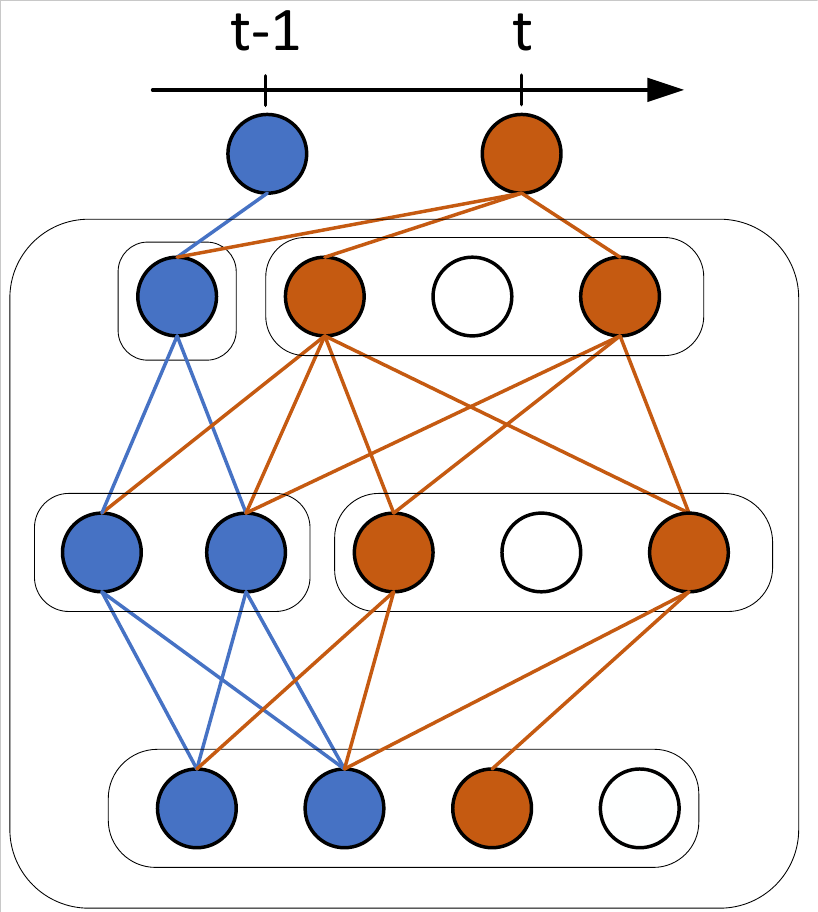} &\hspace{0.2in}
    \includegraphics[height=4.3cm, trim={0.1cm 0.02cm 0.1cm 0.08cm},clip]{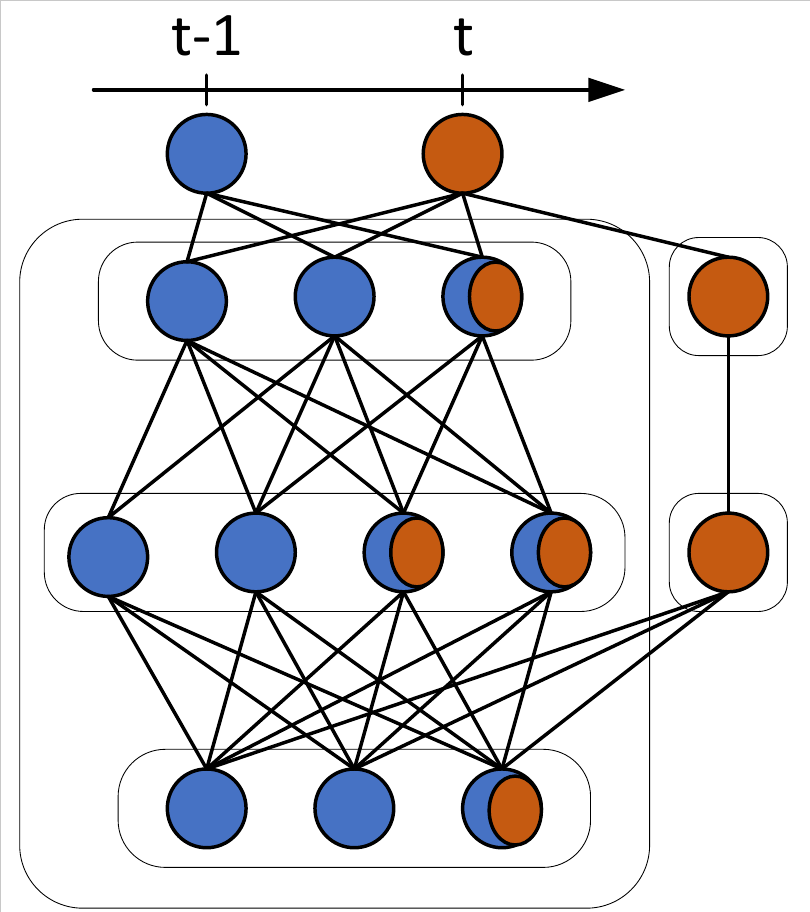} &\hspace{0.2in}
    \includegraphics[height=4.3cm, trim={0.1cm 0.02cm 0.1cm 0.08cm},clip]{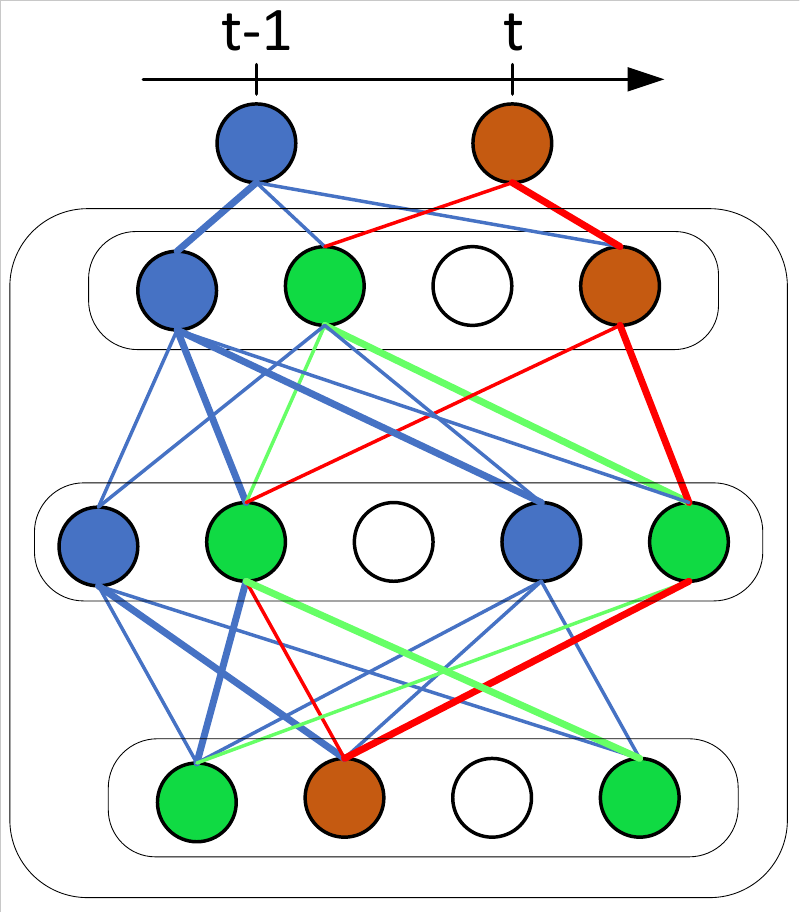} \\ 
    
    \hspace{0.0in}
    \makecell{\small (a) Fixed Backbone \\ \small (Piggyback, SupSup) } &\hspace{0.2in}
    \makecell{\small (b) Biased \small Transfer \\ \small (PackNet, CLNP)} &\hspace{0.2in}
    \makecell{\small (c) Selective Reuse Expansion \\ \small beyond Dense Network (APD)} &\hspace{0.2in}
    \makecell{\small (d) Selective Reuse Expansion \\ \small within Network (WSN)} \\
    \end{tabular}
    }
    \vspace{-0.05in}
    \caption{\textbf{Concept Comparison:} (a) Piggyback \cite{mallya2018piggyback}, and SupSup \cite{wortsman2020supermasks} find the optimal binary mask on a fixed backbone network a given task (b) PackNet \cite{mallya2018packnet} and CLNP \cite{golkar2019continual} forces the model to reuse all features and weights from previous subnetworks which causes bias in the transfer of knowledge (c) APD \cite{Yoon2020} selectively reuse and dynamically expand the dense network (d) Our WSN selectively reuse and dynamically expand subnetworks within a dense network. \textcolor{Green}{Green edges} are reused weights.}
    \label{fig:concept_comparison}
    \vspace{-0.1in}
\end{figure*}

Despite the remarkable success of recent works on rehearsal- and architecture-based continual learning, a majority of recent methods request external memory as new tasks arrive, making the model difficult to scale to larger and more complex tasks. Rehearsal-based CL requires additional storage to store the replay buffer or generative models, and architecture-based methods leverage additional model capacity to account for new tasks. These trends lead to an essential question: how can we build a memory-efficient CL model that does not exceed the backbone network's capacity or even requires a much smaller capacity? Several studies have shown that deep neural networks are over-parameterized~\cite{Denil2013, Han2016learning_both_weights_struct, Li2016pruning_convnets} and thus removing redundant/unnecessary weights can achieve on-par or even better performance than the original dense network. More recently, Lottery Ticket Hypothesis (LTH)~\cite{frankle2018lottery} demonstrates the existence of sparse subnetworks, named \emph{winning tickets}, that preserve the performance of a dense network. However, searching for optimal winning tickets during continual learning with iterative pruning methods requires repetitive pruning and retraining for each arriving task, which could be more practical. Furthermore, leveraged by Regularized Lottery Ticket Hypothesis (RLTH)~\cite{kang2022soft}, subnetworks exemplified that it could overfit a few task data, potentially limiting their effectiveness on new tasks or datasets.


To tackle the issues of external replay buffer, capacity, and over-fitting, we suggest a novel regularized CL method which finds the high-performing \textit{Regularized Winning SubNetwork} referred to as Soft-Subnetwork (\textbf{SoftNet})~\cite{kang2022soft} given tasks without the need for retraining and rewinding. As a baseline of SoftNet, non-regularized Winning Subnetworks referred to as \textbf{WSN}~\cite{kang2022forget} are restated, as shown in \Cref{fig:concept_comparison} (d). Also, we set previous pruning-based CL approaches \cite{mallya2018piggyback, wortsman2020supermasks} (see \Cref{fig:concept_comparison} (a)) to baselines of architectures, which obtain task-specific subnetworks given a pre-trained backbone network. Our WSN incrementally learns model weights and task-adaptive binary masks (the subnetworks) within the neural network. To allow the forward transfer when a model learns on a new task, we reuse the learned subnetwork weights for the previous tasks, however selectively, as opposed to using all the weights \cite{mallya2018packnet} (see \Cref{fig:concept_comparison} (b)), that may lead to biased transfer. Further, the WSN eliminates the threat of catastrophic forgetting during continual learning by freezing the subnetwork weights for the previous tasks and does not suffer from the negative transfer, unlike \cite{YoonJ2018iclr} (see \Cref{fig:concept_comparison} (c)), whose subnetwork weights for the previous tasks can be updated when training on the new tasks.

LTH often leverages the magnitudes of the weights as a pruning criterion to find the optimal subnetworks. However, in CL, relying only on the weight magnitude may be ineffective since the weights are shared across classes, and thus training on the new tasks will change the weights trained for previous tasks (reused weights). The update of reused weights will trigger an avalanche effect where weights selected to be part of the subnetworks for later tasks will always be better in the eyes of the learner, which will result in the catastrophic forgetting of the knowledge for the prior tasks. Thus, in CL, the learner must train on the new tasks without changing the reused weights. To find the optimal subnetworks, we decouple the information of the learning parameter and the network structure into two separate learnable parameters, namely, \emph{weights} and \emph{weight scores}. The weight scores are binary masks with the same shapes as the weights. Now, subnetworks are found by selecting the weights with the top-$k$ percent weight ranking scores. More importantly, decoupling the weights and the structure allows us to find the optimal subnetwork online without iterative retraining, pruning, and rewinding. There is one more thing to consider to find the optimal subnetworks induced by binary masks in CL. According to CL tasks, the subnetworks tend to overfit the few samples, i.e., Few-Shot Class Incremental Learning (FSCIL). Therefore, we could find regularized subnetworks yielded by smooth (soft) masks. To this end, the proposed methods are designed to jointly learn the weights and the structure of the optimal regularized subnetworks, whose overall size is smaller than a dense network.

\noindent
Our contributions can be summarized as follows:
\begin{itemize}[leftmargin=*]
    \item Inspired by Regularized Lottery Ticket Hypothesis (RLTH), we propose novel forget-free continual learning methods referred to as WSN and SoftNet, which learn a compact subnetwork for each task while keeping the weights selected by the previous tasks intact. 
    
    \item Our proposed WSN and SoftNet do not perform explicit pruning for learning the subnetwork. Our methods eliminate catastrophic forgetting and enable forward transfer from previous tasks to new ones in Task Incremental Learning (TIL).  

    \item Our WSN obtains compact subnetworks using Huffman coding with a sub-linear increase in the network capacity, outperforming existing continual learning methods regarding accuracy-capacity trade-off and forward / backward transfer in TIL. 
    
    \item Our SoftNet trains two different types of subnetworks for solving the FSCIL problem, alleviating the continual learner from forgetting previous sessions and overfitting simultaneously, outperforming strong baselines on public benchmark tasks.
\end{itemize}

\section{Related Works}

\noindent
\textbf{Continual Learning.} Continual learning~\cite{McCloskey1989, ThrunS1995, KumarA2012icml, LiZ2016eccv}, also known as lifelong learning, is the challenge of learning a sequence of tasks continuously while utilizing and preserving previously learned knowledge to improve performance on new tasks. Several major approaches have been proposed to tackle the challenges of continual learning, such as catastrophic forgetting. One such approach is \textit{regularization-based methods}~\cite{Kirkpatrick2017,chaudhry2020continual, Jung2020, titsias2019functional, mirzadeh2020linear}, which aim to reduce catastrophic forgetting by imposing regularization constraints that inhibit changes to the weights or nodes associated with past tasks. \textit{Rehearsal-based approaches}~\cite{rebuffi2017icarl, chaudhry2018efficient, chaudhry2019continual, Saha2021, deng2021flattening} store small data summaries to the past tasks and replay them during training to retain the acquired knowledge. Some methods in this line of work~\cite {ShinH2017nips, aljundi2019online} accommodate the generative model to construct the pseudo-rehearsals for previous tasks. \textit{Architecture-based approaches}~\cite{mallya2018piggyback, Serra2018, li2019learn, wortsman2020supermasks, kang2022forget, kang2022soft} use the additional capacity to expand~\cite{xu2018reinforced, YoonJ2018iclr} or isolate~\cite{rusu2016progressive} model parameters, preserving learned knowledge and preventing forgetting. Both rehearsal and architecture-based methods have shown remarkable efficacy in suppressing catastrophic forgetting but require additional capacity for the task-adaptive parameters~\cite{wortsman2020supermasks} or the replay buffers. \\

\noindent
\textbf{Pruning-based Continual Learning.} While most works aim to increase the performance of continual learners by adding memory, some researchers have focused on building memory and computationally efficient continual learners by using pruning-based constraints. CLNP \cite{golkar2019continual} is one example of a method that selects important neurons for a given task using $\ell_1$ regularization to induce sparsity and freezes them to maintain performance. Neurons that are not selected are reinitialized for future task training. Another method, Piggyback \cite{mallya2018piggyback}, trains task-specific binary masks on the weights given a pre-trained model. However, this method does not allow for knowledge transfer among tasks, and its performance highly depends on the quality of the backbone model. HAT~\cite{Serra2018} proposes task-specific learnable attention vectors to identify important weights per task. The masks are formulated to layerwise cumulative attention vectors during continual learning. A recent method, LL-Tickets~\cite{Chen2021lifelonglottery}, shows a sparse subnetwork called lifelong tickets that performs well on all tasks during continual learning. The method searches for more prominent tickets from current ones if the obtained tickets cannot sufficiently learn the new task while maintaining performance on past tasks. However, LL-Tickets require external data to maximize knowledge distillation with learned models for prior tasks, and the ticket expansion process involves retraining and pruning steps. WSN~\cite{kang2022forget} is another method that jointly learns the model and task-adaptive binary masks on subnetworks associated with each task. It attempts to select a small set of weights activated (winning ticket) by reusing weights of the prior subnetworks. \\

\noindent 
\textbf{Soft-subnetwork.} 
Recent studies have shown that context-dependent gating of sub-spaces~\cite{he2018overcoming}, parameters~\cite{mallya2018packnet, he2019task, mazumder2021few}, or layers~\cite{serra2018overcoming} of a single deep neural network is effective in addressing catastrophic forgetting during continual learning. Moreover, combining context-dependent gating with constraints that prevent significant changes in model weights, such as SI~\cite{zenke2017continual} and EWC~\cite{Kirkpatrick2017}, can lead to further performance improvements, as shown by \textit{Masse et al.}~\cite{masse2018alleviating}. Flat minima, which can be seen as acquiring sub-spaces, have also been proposed to address catastrophic forgetting. Previous studies have demonstrated that a flat minimizer is more robust to random perturbations~\cite{hinton1993keeping, hochreiter1994simplifying, jiang2019fantastic}. In a recent study by \textit{Shi et al.}~\cite{shi2021overcoming}, obtaining flat loss minima in the base session, which refers to the first task session with a sufficient number of training instances, was found to be necessary to alleviate catastrophic forgetting in Few-Shot Class Incremental Learning (FSCIL). They achieved this by shifting the model weights on the obtained flat loss contour. Our work investigates the performance of two proposed architecture-based continual learning methods: WSN / Soft-Subnetworks (SoftNet). We select sub-networks~\cite{frankle2018lottery, Liu2018Hier, you2019drawing, zhou2019deconstructing, wortsman2019discovering, Ramanujan2020, kang2022forget, chijiwa2022metaticket} and optimize the regularized sub-network parameters in a sub-space~\cite{kang2022soft} in Task Incremental Learning (TIL) and Few-Shot Class Incremental Learning (FSCIL) settings.

\section{Soft-Winning SubNetworks}
In this section, we present our pruning-based continual learning methods, \textit{Winning SubNetworks} (WSN) \cite{kang2022forget} and \textit{Soft-Winning SubNetworks} (SoftNet) \cite{kang2022soft}. In WSN, the neural network searches for the task-adaptive winning tickets and updates only the weights that have not been trained on the previous tasks. After training on each task, the subnetwork parameters of the model are frozen to ensure that the proposed method is inherently immune to catastrophic forgetting. The WSN method is designed to selectively transfer previously learned knowledge to future tasks (i.e., forward transfer), which can significantly reduce the training time needed for convergence during sequential learning. This feature is especially critical for large-scale learning problems where a continual learner trains on multiple tasks sequentially, leading to significant time and computational savings. Originally, SoftNet was proposed to address the issues of forgetting previous sessions and overfitting a few samples of new sessions. To achieve this, the method trains two types of subnetworks concurrently. \\

\begin{figure*}[ht]
    \centering
    \setlength{\tabcolsep}{-0pt}{%
    \begin{tabular}{cccc}
    \includegraphics[height=5.6cm, trim={0.2cm 0.1cm 0.1cm 0.3cm},clip]{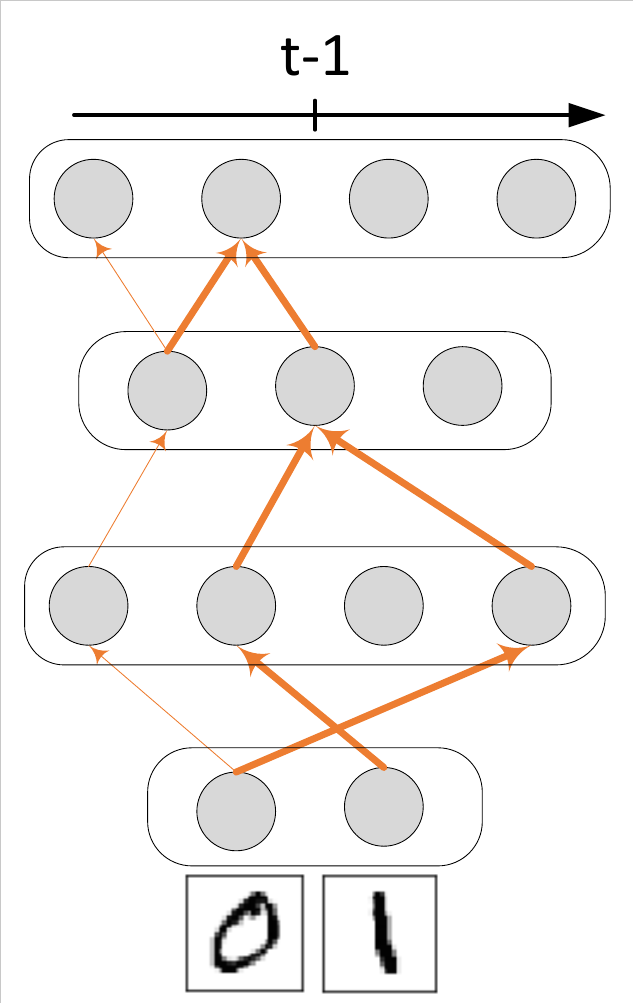} & \hspace{0.2in}
    \includegraphics[height=5.6cm, trim={0.2cm 0.1cm 0.1cm 0.3cm},clip]{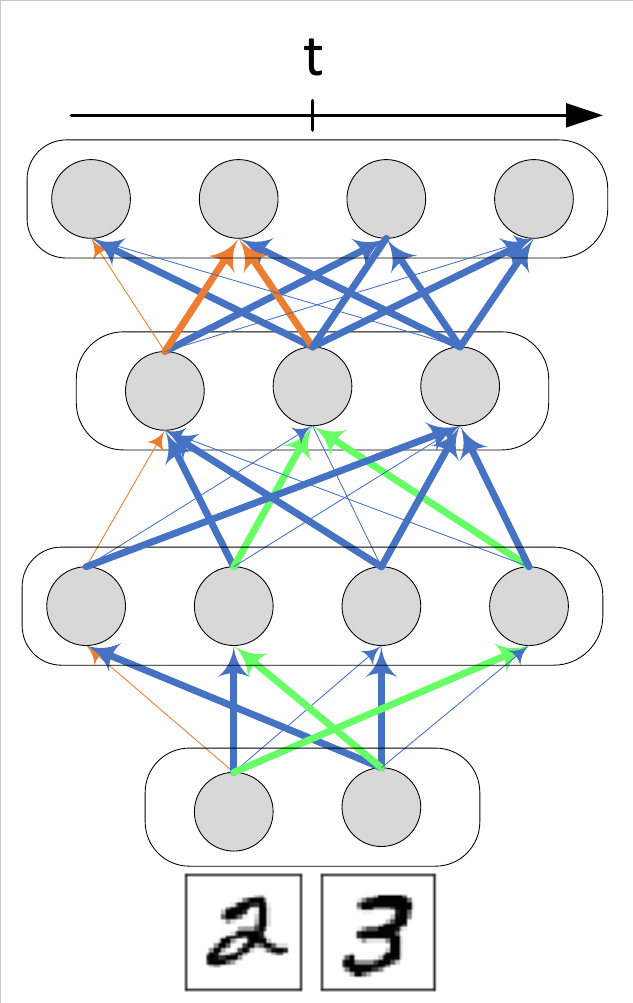} & \hspace{0.2in}
    \includegraphics[height=5.6cm, trim={0.2cm 0.1cm 0.1cm 0.3cm},clip]{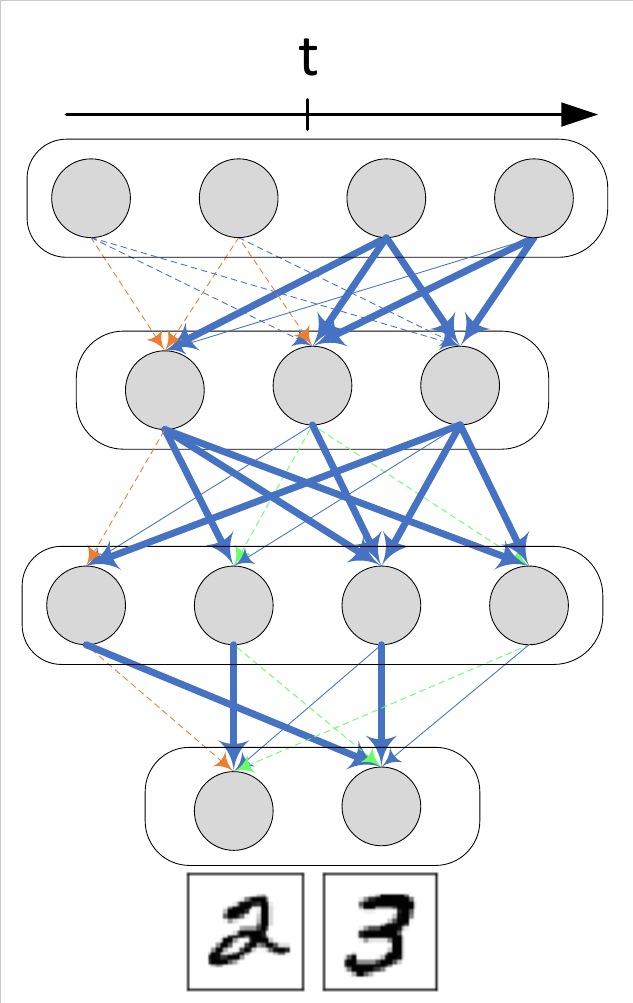} & \hspace{0.2in}
    \includegraphics[height=5.6cm, trim={0.2cm 0.1cm 0.1cm 0.3cm},clip]{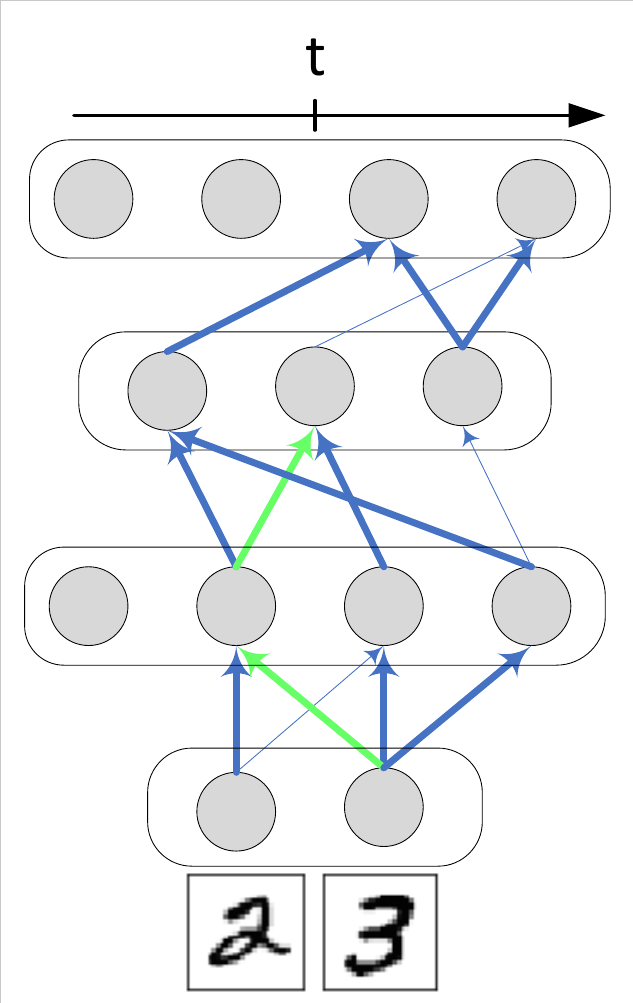} \\ 
    \hspace{0.2in}
    \makecell{\small (a) selected \textcolor{orange}{weights} $\hat{\bsy\theta}_{t-1}$ \\ \small at prior task} &\hspace{0.25in}
    \makecell{\small (b) forward pass \\ \small using \textcolor{Green}{reused weights}} &\hspace{0.25in}
    \makecell{\small (c) backward pass \\ \small only on \textcolor{blue}{non-used weights}} &\hspace{0.25in}
    \makecell{\small (d) selected \textcolor{blue}{weights} $\hat{\bsy\theta}_{t}$ \\ \small with subsets of \textcolor{Green}{reused weights}} \\

    \end{tabular}
    }
    \vspace{-0.1in}
    \caption{\textbf{An illustration of Winning SubNetworks (WSN):} (a) The top-c\% \textcolor{orange}{weights} $\hat{\bsy\theta}_{t-1}$ at prior task are obtained, (b) In the forward pass of a new task, WSN \textcolor{Green}{reuses weights} selected from prior tasks, (c) In the backward pass, WSN updates only \textcolor{blue}{non-used weights}, and (d) after several iterations of (b) and (c), we acquire again the top-c\% \textcolor{blue}{weights} $\hat{\bsy\theta}_{t}$ including subsets of \textcolor{Green}{reused weights} for the new task.}
  \label{fig:concept_figure}
  \vspace{-0.2in}
\end{figure*}

\noindent 
\textbf{Problem Statement.} Consider a supervised learning setup where $T$ tasks arrive to a learner sequentially. We denote that $\mathcal{D}_t=\{\textbf{x}_{i,t}, y_{i,t}\}_{i=1}^{n_t}$ is the dataset of task $t$, composed of $n_t$ pairs of raw instances and corresponding labels. We assume a neural network $f(\cdot;\bsy\theta)$, parameterized by the model weights $\bsy\theta$ and standard continual learning scenario aims to learn a sequence of tasks by solving the following optimization procedure at each step $t$: 
\begin{equation}
\bsy \theta^{\ast}=\minimize_{\bsy\theta} \frac{1}{n_t}\sum^{n_t}_{i=1}\mathcal{L}(f(\textbf{x}_{i,t};\bsy\theta), y_{i,t}),
\label{eq:task_loss}
\end{equation}
where $\mathcal{L}(\cdot, \cdot)$ is a classification objective loss such as cross-entropy loss. $\mathcal{D}_t$ for task $t$ is only accessible when learning task $t$. Note rehearsal-based continual learning methods allow memorizing a small portion of the dataset to replay. We further assume that task identity is given in the training and testing stages. 

Continual learners frequently use over-parameterized deep neural networks to ensure enough capacity for learning future tasks. This approach often leads to the discovery of subnetworks that perform as well as or better than the original network. Given the neural network parameters $\bsy\theta$, the binary attention mask $\mathbf{m}^*_t$ that describes the optimal subnetwork for task $t$ such that $|\mathbf{m}^*_t|$ is less than the model capacity $c$  follows as:
\begin{equation}
\begin{split}
    \mathbf{m}^*_t &= \underset{\mathbf{m}_t\in\{0,1\}^{|\bsy\theta|}}{\minimize} \frac{1}{n_t}\sum^{n_t}_{i=1}\mathcal{L}\big(f(\textbf{x}_{i,t};\bsy\theta\odot \mathbf{m}_t), y_{i,t}\big) - \mathcal{J}  \\
    &\quad\quad\quad\quad\quad\quad \text{subject to~}|\mathbf{m}^*_t|\leq c,
\end{split}
\label{eq:subnetwork}
\end{equation}
where task loss $\mathcal{J}=\mathcal{L}\big(f(\textbf{x}_{i,t};\bsy\theta), y_{i,t}\big)$ and $c\ll|\bsy\theta|$ (used as the selected proportion $\%$ of model parameters in the following section). In the optimization section, we describe how to obtain $\mathbf{m}^*_t$ using a single learnable weight score $\mathbf{s}$ that is subject to updates while minimizing task loss jointly for each task. 

\subsection{Winning SubNetworks (WSN)}\label{sub_sec:wsn}
Let each weight be associated with a learnable parameter we call \textit{weight score} $\mathbf{s}$, which numerically determines the importance of the weight associated with it; that is, a weight with a higher weight score is seen as more important. We find a sparse subnetwork $\hat{\bsy\theta}_t$ of the neural network and assign it as a solver of the current task $t$. We use subnetworks instead of the dense network as solvers for two reasons: (1) Lottery Ticket Hypothesis \cite{frankle2018lottery} shows the existence of a subnetwork that performs as well as the whole network, and (2) subnetwork requires less capacity than dense networks, and therefore it inherently reduces the size of the expansion of the solver.

Motivated by such benefits, we propose a novel \textit{Winning SubNetworks} (WSN\footnote{WSN code is available at \url{https://github.com/ihaeyong/WSN.git}}), which is the joint-training method for continual learning that trains on task - while selecting an important subnetwork given the task $t$ as shown in Fig. \ref{fig:concept_figure}. The illustration of WSN explains how to acquire binary weights within a dense network step by step. We find $\hat{\bsy\theta}_t$ by selecting the $c$\% weights with the highest weight scores $\mathbf{s}$, where $c$ is the target layerwise capacity ratio in \%. A task-dependent binary weight represents the selection of weights $\mathbf{m}_t$ where a value of $1$ denotes that the weight is selected during the forward pass and $0$ otherwise. Formally, $\mathbf{m}_t$ is obtained by applying a indicator function $\mathbbm{1}_c$ on $\mathbf{s}$ where $\mathbbm{1}_c(s)=1$ if $\mathbf{s}$ belongs to top-$c\%$ scores and $0$ otherwise. Therefore, the subnetwork $\hat{\bsy\theta}_t$ for task $t$ is obtained by $\hat{\bsy\theta}_t = \bsy\theta \odot \mathbf{m}_t$.  

\subsection{Soft-Subnetworks (SoftNet)}
Several works have addressed overfitting issues in continual learning from different perspectives, including NCM~\cite{hou2019learning}, BiC~\cite{wu2019large}, OCS~\cite{yoon2022online}, and FSLL~\cite{mazumder2021few}. To mitigate the overfitting issue in subnetworks, we use a simple yet efficient method named \emph{SoftNet} proposed by \cite{kang2022soft}. The following new paradigm, referred to as \emph{Regularized Lottery Ticket Hypothesis}~\cite{kang2022soft} which is inspired by the \emph{Lottery Ticket Hypothesis}~\cite{frankle2018lottery} has become the cornerstone of SoftNet: 

\noindent
\textbf{Regularized Lottery Ticket Hypothesis (RLTH).} \textit{A randomly-initialized dense neural network contains a regularized subnetwork that can retain the prior class knowledge while providing room to learn the new class knowledge through isolated training of the subnetwork.} \\

\noindent
\textbf{SoftNet}. Based on RLTH, we propose a method, referred to as \textbf{Soft}-Sub\textbf{Net}works (\textbf{SoftNet}\footnote{SoftNet code is available at \url{https://github.com/ihaeyong/SoftNet-FSCIL.git}}). SoftNet jointly learns the randomly initialized dense model, and soft mask $\bm{m} \in [0, 1]^{|\bm \theta|}$ on Soft-subnetwork on each task training; the soft mask consists of the major part of the model parameters $m=1$ and the minor ones $m<1$ where $m=1$ is obtained by the top-$c\%$ of model parameters and $m<1$ is obtained by the remaining ones ($100 - \text{top-}c\%$) sampled from the uniform distribution $U(0, 1)$. Here, it is critical to select minor parameters $m<1$ in a given dense network. 

\begin{figure}[ht]
    \centering
    \setlength{\tabcolsep}{-6pt}{%
    \begin{tabular}{cc}
    \includegraphics[width=0.25\textwidth]{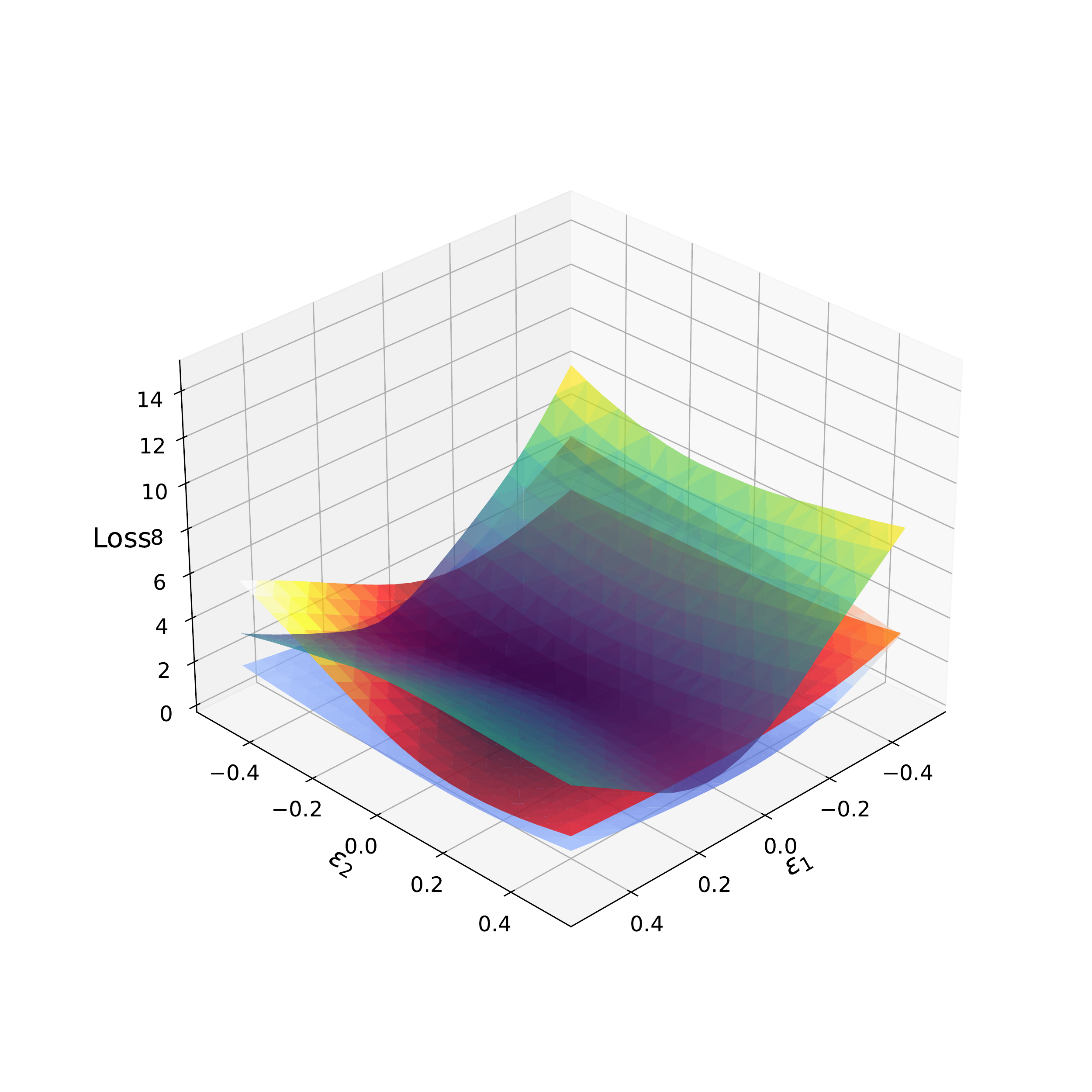} &
    \includegraphics[width=0.25\textwidth]{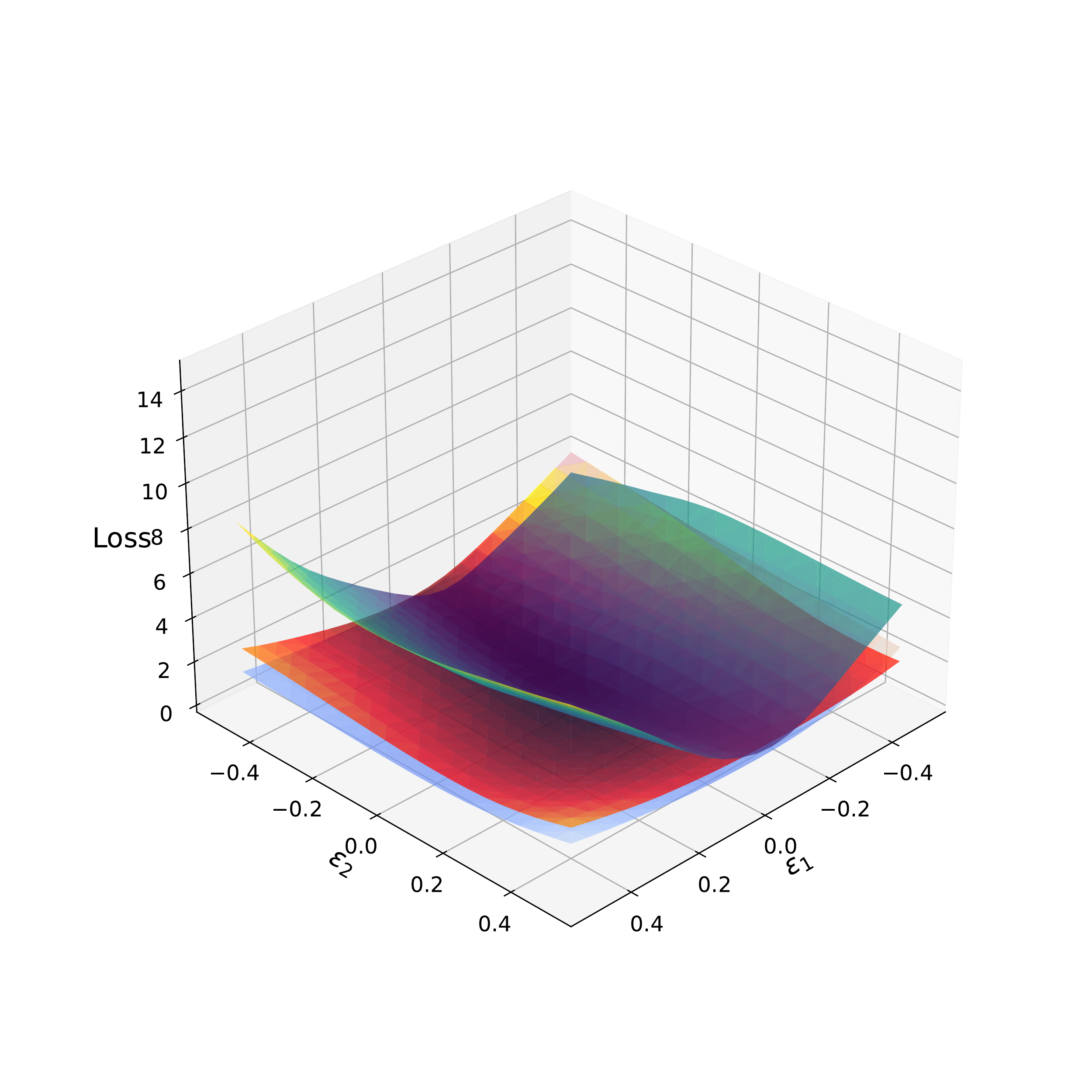} \\ 
    
    \small (a) epoch=70, $c=10.0\%$ & \small (b) epoch=100, $c=10.0\%$  \\
    \end{tabular}
    }
    \caption{\small \textbf{Loss landscapes of} \textcolor{teal}{Dense Network}, \textcolor{red}{HardNet (WSN)}, \textbf{and} \textcolor{cyan}{SoftNet}: Subnetworks provide a more flat global minimum than dense neural networks. To demonstrate the loss landscapes, we trained a simple three-layered, fully connected model (fc-4-25-30-3) on the Iris Flower dataset (which is three classification problem) for 100 epochs \cite{hart1968condensed,dasarathy1980nosing}.}
    \label{fig:main_plot_loss_lens}
    \vspace{-0.12in}
\end{figure}

\subsection{Convergence of Subnetworks}

\textbf{Convergences of HardNet (WSN) / SoftNet.} To interpret the convergence of SoftNet, we follow the Lipschitz-continuous objective gradients~\cite{boyd2004convex, bottou2018optimization}: the objective function of dense networks $R: \mathbb{R}^d \rightarrow \mathbb{R}$ is continuously differentiable and the gradient function of $R$, namely, $\nabla R: \mathbb{R}^d \rightarrow \mathbb{R}^d$, \textit{Lipschitz continuous with Lipschitz constant} $L > 0$, i.e., 

\begin{equation} \label{eq:dense_lip}
\begin{split}
    || \nabla R(\bm{\theta}) - \nabla R(\bm{\theta}') ||_2 & \leq  L || \bm{\theta} - \bm{\theta}' || \;\; \\ 
    & \text{ for all } \{\bm{\theta}, \bm{\theta}'\} \subset \mathbb{R}^d.
\end{split} 
\end{equation}

\noindent
\textbf{Proposition.} Subnetwork achieves a faster rate than dense networks. To prove this, following the same formula, we define the Lipschitz-continuous objective gradients of subnetworks as follows:

\begin{equation} \label{eq:subnet_lip}
\begin{split}
    || \nabla R(\bm{\theta} \odot \bm{m}) - \nabla R(\bm{\theta}' \odot \bm{m}) ||_2 &\leq  L || (\bm{\theta} - \bm{\theta}') \odot \bm{m} || \\ 
    & \text{ for all } \{\bm{\theta}, \bm{\theta}'\} \subset \mathbb{R}^d.
\end{split} 
\end{equation}
where $\bs{m}$ is a binary mask. In comparision of Eq. \ref{eq:dense_lip} and \ref{eq:subnet_lip}, we use the theoretical analysis~\cite{ye2020good} where subnetwork achieve a faster rate of $R(\bm{\theta} \odot \bm{m})=\mathcal{O}(1/||\bm{m}||^2_1)$ at most. The comparison is as follows:
\begin{equation}
\begin{split}
    \frac{|| \nabla R(\bm{\theta} \odot \bm{m}) - \nabla R(\bm{\theta}' \odot \bm{m}) ||_2}{|| (\bm{\theta} - \bm{\theta}') \odot \bm{m} ||}  &< \\
    \frac{|| \nabla R(\bm{\theta}) - \nabla R(\bm{\theta}') ||_2}{|| \bm{\theta} - \bm{\theta}' ||} &\leq L
\end{split}
\end{equation}
The smaller the value is, the flatter the solution (loss landscape) has. The equation is established from the relationship $R(\bm{\theta} \odot \bm{m}) \ll R^\ast(\bm{\theta})$, where $R^\ast(\bm{\theta}$ denotes the best possible loss achievable by convex combinations of all parameters despite $|| (\bm{\theta} - \bm{\theta}') \odot \bm{m} ||<|| \bm{\theta} - \bm{\theta}' ||$. \\

\noindent 
\textbf{Corollary.}
Furthermore, we have the following inequality if $|| R(\bm{\theta} \odot \bm{m}_{hard})-R(\bm{\theta} \odot \bm{m}_{soft})|| \simeq 0$ and $||\bm{m}_{hard}|| < ||\bm{m}_{soft}||$:
\begin{equation} 
\begin{split}
    & \frac{|| \nabla R(\bm{\theta} \odot \bm{m}_{hard}) - \nabla R(\bm{\theta}' \odot \bm{m}_{hard}) ||_2}{|| (\bm{\theta} - \bm{\theta}') \odot \bm{m}_{hard} ||} \geq \\ & \frac{|| \nabla R(\bm{\theta} \odot \bm{m}_{soft}) - \nabla R(\bm{\theta}' \odot \bm{m}_{soft}) ||_2}{|| (\bm{\theta} - \bm{\theta}') \odot \bm{m}_{soft}||}
\end{split}
\end{equation}
where the equality holds \textit{iff} $||\bm{m}_{hard}|| = ||\bm{m}_{soft}||$. We prepare the loss landscapes of Dense Network, HardNet (WSN), and SoftNet as shown in \Cref{fig:main_plot_loss_lens} as an example to support the above subnetwork's inequality.

\subsection{Optimization of Winning SubNetworks (WSN)}\label{sub_sec:opt_wsn}
To jointly learn the model weights and task-adaptive binary masks of subnetworks associated with each task, given an objective $\mathcal{L}(\cdot)$, we optimize $\bsy\theta$ and $\mathbf{s}$ with:
\begin{equation}
\label{eq:optimizer}
    \minimize_{\bsy\theta, \mathbf{s}} \mathcal{L} (\bsy\theta \odot \mathbf{m}_t; \mathcal{D}_t).
\end{equation}
However, this vanilla optimization procedure presents two problems: (1) updating all $\bsy\theta$ when training for new tasks will cause interference to the weights allocated for previous tasks, and (2) the indicator function always has a gradient value of $0$; therefore, updating the weight scores $\mathbf{s}$ with its loss gradient is not possible. To solve the first problem, we selectively update the weights by allowing updates only on the weights not selected in the previous tasks. To do that, we use an \textit{accumulate binary mask} $\mathbf{M}_{t-1}=\logicalor_{i=1}^{t-1} \mathbf{m}_i$ when learning task $t$, then for an optimizer with learning rate $\eta,$ the $\bsy\theta$ is updated as follows: 
\begin{equation}
\label{eq:param_update}
\bsy\theta \leftarrow \bsy\theta - \eta \left(\frac{\partial \mathcal{L}}{\partial \bsy\theta} \odot (\mathbf{1}-\mathbf{M}_{t-1})\right),
\end{equation} 
effectively freezing the weights of the subnetworks selected for the previous tasks. To solve the second problem, we use Straight-through Estimator \cite{Hinton2012, Bengio2013, Ramanujan2020} in the backward pass since $\mathbf{m}_t$ is obtained by top-$c\%$ scores. Specifically, we ignore the derivatives of the indicator function and update the weight score as follows: 
\begin{equation}
\label{eq:ste}
\quad \mathbf{s} \leftarrow \mathbf{s} - \eta\left(\frac{\partial \mathcal{L}}{\partial \mathbf{s}}\right).
\end{equation}

The use of separate weight scores $\mathbf{s}$ as the basis for selecting subnetwork weights makes it possible to reuse some of the weights from previously chosen weights $\bsy\theta \odot \mathbf{m}_t$ in solving the current task $t$, which can be viewed as \textit{transfer learning}. Likewise, previously chosen weights that are irrelevant to the new tasks are not selected; instead, weights from the set of not-yet-chosen weights are selected to meet the target network capacity for each task, which can be viewed as \textit{finetuning} from tasks $\{1,...,t-1\}$ to task $t$. Our WSN optimizing procedure is summarized in pseudo algorithm \ref{alg:algorithm1}.

\begin{algorithm}[ht]
  \caption{Winning SubNetworks (WSN) for TIL.}\label{alg:algorithm1}
    \small
    \begin{algorithmic}[1]
    \INPUT $\{\mathcal{D}_t\}_{t=1}^{\mathcal{T}}$, model weights $\bsy\theta$, score weights $\mathbf{s}$, binary mask $\mathbf{M}_0=\mathbf{0}^{|\bsy\theta|}$, layer-wise capacity $c$
    \STATE Randomly initialize $\bsy\theta$ and $\mathbf{s}$.
    \FOR{task $t = 1, . . . , \mathcal{T}$}
        \FOR{batch $\mathbf{b}_t\sim\mathcal{D}_t$} 
            \STATE Obtain soft-mask $\mathbf{m}_t$ of the top-$c\%$ scores $\mathbf{s}$ at each layer 
            \STATE Compute $\mathcal{L}\left( \bsy\theta \odot \mathbf{m}_t;\mathbf{b}_t \right)$
            \STATE $\bsy\theta \leftarrow \bsy\theta - \eta \left(\frac{\partial \mathcal{L}}{\partial \bsy\theta} \odot (\mathbf{1}-\mathbf{M}_{t-1})\right)$ \COMMENT{Weight update}     
            \STATE $\mathbf{s} \leftarrow \mathbf{s} - \eta(\frac{\partial \mathcal{L}}{\partial \mathbf{s}})$ \COMMENT{Weight score update}
        \ENDFOR
        \STATE $\mathbf{M}_{t} \leftarrow \mathbf{M}_{t-1} \newor \mathbf{m}_t$ \COMMENT{Accumulate binary mask}
    \ENDFOR
  \end{algorithmic}
\end{algorithm}

\subsection{Winning Ticket Encoding}
A binary mask is needed to store task-specific weights for each task in WSN. However, the main issue is that as the number of tasks increases in deep-learning models, the number of binary masks required also increases. We use a compression algorithm to compress all binary task masks to address this capacity issue and achieve forget-free continual learning. To accomplish this, we first convert a sequence of binary masks into a single accumulated decimal mask and then convert each integer into an ASCII code to represent a unique symbol. Next, we apply Huffman encoding~\cite{huffman1952method}, a lossless compression algorithm, to the symbols. We empirically observed that Huffman encoding compresses 7-bit binary maps with a compression rate of approximately 78\% and decompresses them without any bit loss. Furthermore, experimental results demonstrate that the compression rate is sub-linearly increasing with the size of binary bits.

\subsection{SoftNet via Complementary Winning Tickets}
Similar to WSN's optimization discussed in \Cref{sub_sec:opt_wsn}, let each weight be associated with a learnable parameter we call \textit{weight score} $\bm{s}$, which numerically determines the importance of the associated weight. In other words, we declare a weight with a higher score more important. At first, we find a subnetwork ${\bm \theta}^\ast = \bm{\theta} \odot \bm{m}^\ast_t$ of the dense neural network and then assign it as a solver of the current session $t$. The subnetworks associated with each session jointly learn the model weight $\bm{\theta}$ and binary mask $\bm{m}_t$. Given an objective $\mathcal{L}_t$, we optimize $\bm{\theta}$ as follows:
\begin{equation}
    \bm{\theta}^\ast, \bm{m}^\ast_t = \minimize_{\bm{\theta}, \bm{s}} \mathcal{L}_t (\bm{\theta} \odot \bm{m}_t; \mathcal{D}_t).
    \label{eq:softnet_loss}
\end{equation}
where $\bm{m}_t$ is obtained by applying an indicator function $\mathbbm{1}_c$ on weight scores $\bm{s}$. Note $\mathbbm{1}_c(s)=1$ if $s$ belongs to top-c$\%$ scores and $0$ otherwise.

In the optimization process for FSCIL, however, we consider two main problems: (1) Catastrophic forgetting: updating all $\bm{\theta} \odot \bm{m}_{t-1}$ when training for new sessions will cause interference with the weights allocated for previous tasks; thus, we need to freeze all previously learned parameters $\bm{\theta} \odot \bm{m}_{t-1}$; (2) Overfitting: the subnetwork also encounters overfitting issues when training an incremental task on a few samples, as such, we need to update a few parameters irrelevant to previous task knowledge., i.e., $\bm{\theta} \odot (\bm{1}-\bm{m}_{t-1})$.  

To acquire the optimal subnetworks that alleviate the two issues, we define a soft-subnetwork by dividing the dense neural network into two parts-one is the major subnetwork $\bm{m}_\text{major}$, and another is the minor subnetwork $\bm{m}_\text{minor}$. The defined Soft-SubNetwork (\textcolor{cyan}{SoftNet}) follows as:
\begin{equation}
    \bm{m}_\text{soft} = \bm{m}_\text{major} \oplus \bm{m}_\text{minor},
    \label{eq:soft_mask}
\end{equation}
where $\bm{m}_\text{major}$ is a binary mask and $\bm{m}_\text{minor} \sim U(0,1)$ and $\oplus$ represents an element-wise summation. As such, a soft-mask is given as $\bs{m}_t^\ast \in [0,1]^{|\bm{\theta}|}$ in Eq.\ref{eq:softnet_loss}. In the all-experimental FSCIL setting, $\bm{m}_\text{major}$ maintains the base task knowledge $t=1$ while $\bm{m}_\text{minor}$ acquires the novel task knowledge $t \geq 2$. Then, with base session learning rate $\alpha,$ the $\bsy\theta$ is updated as follows: $\bm{\theta} \leftarrow \bm{\theta} - \alpha \left(\frac{\partial \mathcal{L}}{\partial \bm{\theta}} \odot \bm{m}_\text{soft}\right)$ effectively regularize the weights of the subnetworks for incremental learning. The subnetworks are obtained by the indicator function that always has a gradient value of $\bm{0}$; therefore, updating the weight scores $\bm{s}$ with its loss gradient is impossible. We update the weight scores by using Straight-through Estimator \cite{Hinton2012, Bengio2013, Ramanujan2020} in the backward pass. Specifically, we ignore the derivatives of the indicator function and update the weight score $\bm{s} \leftarrow \bm{s} - \alpha \left(\frac{\partial \mathcal{L}}{\partial \bm{s}} \odot \bm{m}_\text{soft} \right)$, where $\bm{m}_{\text{soft}}=\bm{1}$ for exploring the optimal subnetwork for base session training. Our Soft-subnetwork optimizing procedure is summarized in Algorithm \ref{alg:algorithm2}. Once a single soft-subnetwork $\bs{m}_{\text{soft}}$ is obtained in the base session, then we use the soft-subnetwork for the entire new sessions without updating.

\begin{algorithm}[ht]
  \caption{Soft-Subnetworks (SoftNet) for FSCIL.}\label{alg:algorithm2}
    \small
    \begin{algorithmic}[1]
    \INPUT $\{\mathcal{D}^t\}_{t=1}^{\mathcal{T}}$, model weights $\bm\theta$, and score weights $\bm{s}$, layer-wise capacity $c$ \\
    \STATE \textcolor{blue}{// Training over base classes $t = 1$} \\ 
    \STATE Randomly initialize $\bm\theta$ and $\bm{s}$. \\
    \FOR{epoch $e = 1, 2, \cdots$}
        \STATE Obtain softmask $\bm{m}_\text{soft}$ of $\bm{m}_{major}$ and $\bm{m}_{minor} \sim U(0,1)$ at each layer \\
        \FOR{batch $\bm{b}_t\sim \mathcal{D}^t$} 
            \STATE Compute $\mathcal{L}_{base}\left( \bm\theta \odot \bm{m}_\text{soft};\bm{b}_t \right)$ by Eq.~\ref{eq:softnet_loss}
            
            \STATE $\bm\theta \leftarrow \bm\theta - \alpha \left(\frac{\partial \mathcal{L}}{\partial \bm\theta} \odot \bm{m}_\text{soft}\right)$ 
            \STATE $\bm{s} \leftarrow \bm{s} - \alpha \left(\frac{\partial \mathcal{L}}{\partial \bm{s}} \odot \bm{m}_\text{soft}\right)$ 
        \ENDFOR
    \ENDFOR \\
    \STATE \textcolor{blue}{// Incremental learning $t \geq 2$} \\
    \STATE \text{Combine the training data $\mathcal{D}^t$} \\
    \STATE \text{~~~and the exemplars saved in previous few-shot sessions} \\
    \FOR{epoch $e = 1, 2, \cdots$}
        \FOR{batch $\bm{b}_t\sim \mathcal{D}^t$} 
            \STATE Compute $\mathcal{L}_{m}\left( \bm\theta \odot \bm{m}_\text{soft};\bm{b}_t \right)$ by Eq.~\ref{eq:proto_loss} 
            
            \STATE $\bm{\theta} \leftarrow \bm{\theta} - \beta \left(\frac{\partial \mathcal{L}}{\partial \bm{\theta}} \odot \bm{m}_{\textcolor{red}{minor}} \right)$ 
        \ENDFOR
    \ENDFOR \\
    \OUTPUT model parameters $\bm{\theta}$, $\bm{s}$, and $\bm{m}_{\text{soft}}$.
  \end{algorithmic}
\end{algorithm}

\subsection{Soft-SubNetwork for Incremental Learning}
We now describe the overall procedure of our soft-pruning-based incremental learning/inference method under the following FSCIL settings. This includes the training phase with a normalized informative measurement, as outlined in prior work~\cite{shi2021overcoming}, and the inference phase.

\noindent 
\textbf{Few-shot Class Incremental Learning} (FSCIL) aims to learn new sessions with only a few examples continually. A FSCIL model learns a sequence of $T$ training sessions $\{\mathcal{D}^1,\cdots, \mathcal{D}^T\}$, where $\mathcal{D}^t=\{z_i^t = (\bm{x}_i^t, y_i^t)\}^{n_t}_i$ is the training data of session $t$ and $\bm{x}_i^t$ is an example of class $y_i^t \in \mathcal{O}^t$. In FSCIL, the base session $\mathcal{D}^1$ usually contains a large number of classes with sufficient training data for each class. In contrast, the subsequent sessions ($t \geq 2$) will only contain a small number of classes with a few training samples per class, e.g., the $t^{\mathrm{th}}$ session $\mathcal{D}^t$ is often presented as a \textit{N}-way \textit{K}-shot task. In each training session $t$, the model can access only the training data $\mathcal{D}^t$ and a few examples stored in the previous session. When the training of session $t$ is completed, we evaluate the model on test samples from all classes $\mathcal{O} = \bigcup_{i=1}^t \mathcal{O}^i$, where $\mathcal{O}^i \bigcap \mathcal{O}^{j\neq i} = \emptyset$ for $\forall i, j \leq T$.

\noindent 
\textbf{Base Training} $(t = 1)$. In the base learning session, we optimize the soft-subnetwork parameter $\bm\theta$ (including a fully-connected layer as a classifier) and weight score $\bm s$ with cross-entropy loss jointly using the training examples of $\mathcal{D}^1$. 

\noindent 
\textbf{Incremental Training} $(t \geq 2)$. In the incremental few-shot learning sessions $(t \geq 2)$, leveraged by $\bm{\theta} \odot \bm{m}_{\text{soft}}$, we fine-tune few minor parameters $\bm{\theta} \odot \bm{m}_{\text{minor}}$ of the soft-subnetwork to learn new classes. Since $\bm{m}_{\text{minor}} < \bm{1}$, the soft-subnetwork alleviates the overfitting of a few samples. Furthermore, instead of Euclidean distance \cite{shi2021overcoming}, we employ a metric-based classification algorithm with cosine distance to finetune the few selected parameters. In some cases, Euclidean distance fails to give the real distances between representations, especially when two points with the same distance from prototypes do not fall in the same class. In contrast, representations with a low cosine distance are located in the same direction from the origin, providing a normalized informative measurement. We define the loss function as: 
\begin{equation}
\begin{split}
    & \mathcal{L}_m (z; \bm{\theta} \odot \bm{m}_{soft}) =\\
    & -\sum_{z \in \mathcal{D}} \sum_{o \in \mathcal{O}} \mathbbm{1}(y=o) \log \left( \frac{e^{-d(\bm{p}_o, f(\bm{x};\; \bm{\theta} \odot \bm{m}_{soft}))}}{\sum_{o_k \in \mathcal{O}} e^{-d(\bm{p}_{o_k}, f(\bm{x};\; \bm{\theta} \odot \bm{m}_{soft}))}} \right)
\end{split}
    \label{eq:proto_loss}
\end{equation}
where $d\left(\cdot, \cdot\right)$ denotes cosine distance, $\bm{p}_o$ is the prototype of class $o$, $\mathcal{O} = \bigcup_{i=1}^t \mathcal{O}^i$ refers to all encountered classes, and $\mathcal{D} = \mathcal{D}^t \bigcup \mathcal{P}$ denotes the union of the current training data $\mathcal{D}^t$ and the exemplar set $\mathcal{P} = \left\{\bm{p}_2 \cdots, \bm{p}_{t-1}\right\}$, where $\mathcal{P}_{t_e} \left(2 \leq t_e < t\right)$ is the set of saved exemplars in session $t_e$. Note that the prototypes of new classes are computed by $\bm{p}_o = \frac{1}{N_o} \sum_i \mathbbm{1}(y_i = o) f(\bm{x}_i; \bm{\theta} \odot \bm{m}_{soft})$ and those of base classes are saved in the base session, and $N_o$ denotes the number of the training images of class $o$. We also save the prototypes of all classes in $\mathcal{O}^t$ for later evaluation.  \\

\noindent 
\textbf{Inference for Incremental Soft-Subnetwork.} In each session, the inference is also conducted by a simple nearest class mean (NCM) classification algorithm \cite{mensink2013distance, shi2021overcoming} for fair comparisons. Specifically, all the training and test samples are mapped to the embedding space of the feature extractor $f$, and Euclidean distance $d_u(\cdot, \cdot)$ is used to measure the similarity between them. The classifier gives the $k$th prototype index $o_k^{\ast} = \arg\min_{o \in \mathcal{O}} d_u(f(\bm{x}; \bm{\theta} \odot \bm{m}_{soft}), \bm{p}_o)$ as output.

\section{Experiments}
We now validate our method on several benchmark datasets against relevant continual learning baselines on Task-Incremental Learning (TIL) and Few-shot Class Incremental Learning (FSCIL).

\subsection{Task-incremental Learning (TIL)}
First, we consider task-incremental continual learning with a multi-head configuration for all experiments in the paper. We follow the experimental setups in recent works \cite{Saha2021, Yoon2020, deng2021flattening}.

\noindent 
\textbf{Datasets and architectures.} We use six different popular sequential datasets for CL problems with five different neural network architectures as follows: 
\textbf{1) Permuted MNIST (PMNIST):} A variant of MNIST~\cite{LeCun1998} where each task has a deterministic permutation to the input image pixels.
\textbf{2) 5-Datasets:} A mixture of 5 different vision datasets~\cite{Saha2021}: CIFAR-10~\cite{Krizhevsky2009}, MNIST~\cite{LeCun1998}, SVHN~\cite{Netzer2011SVHN}, FashionMNIST~\cite{Xiao2017Fashion}, and notMNIST~\cite{Yaroslav2011notMNIST}. 
\textbf{3) Omniglot Rotation:} An OCR images dataset composed of 100 tasks, each including 12 classes. We further preprocess the raw images by generating their rotated version in $90 \degree,~180\degree, \text{~and~} 270\degree$, followed by \cite{Yoon2020}. \textbf{4) CIFAR-100 Split} \cite{Krizhevsky2009}\textbf{:} A visual object dataset constructed by randomly dividing 100 classes of CIFAR-100 into ten tasks with ten classes per task. \textbf{5) CIFAR-100 Superclass:} We follow the setting from \cite{Yoon2020} that divides CIFAR-100 dataset into 20 tasks according to the 20 superclasses, and each superclass contains five different but semantically related classes. \textbf{6) TinyImageNet}~\cite{Stanford}\textbf{:} A variant of ImageNet~\cite{krizhevsky2012imagenet} containing 40 of 5-way classification tasks with the image size by $64 \times 64\times 3$.

We use two-layered MLP with $100$ neurons per layer for PMNIST, variants of LeNet~\cite{LeCun1998} for the experiments on Omniglot Rotation and CIFAR-100 Superclass experiments. Also, we use a modified version of AlexNet similar to \cite{Serra2018, Saha2021} for the CIFAR-100 Split dataset and a reduced ResNet-18 similar to \cite{chaudhry2019continual, Saha2021} for 5-Datasets. For TinyImageNet, we also use the same network architecture as \cite{gupta2020maml, deng2021flattening}, which consists of 4 Conv layers and three fully connected layers. \\

\noindent 
\textbf{Baselines.} We compare our WSN with strong CL baselines; regularization-based methods: \textbf{HAT}~\cite{Serra2018} and \textbf{EWC}~\cite{Kirkpatrick2017}, rehearsal-based methods: \textbf{GPM}~\cite{Saha2021}, and a pruning-based method: \textbf{PackNet}~\cite{mallya2018packnet} and \textbf{SupSup}~\cite{wortsman2020supermasks}. \textbf{PackNet} and \textbf{SupSup} is set to the baseline to show the effectiveness of re-used weights. We also compare with naive sequential training strategy, referred to as \textbf{FINETUNE}. \textbf{Multitask Learning (MTL)} and \textbf{Single-task Learning (STL)} are not a CL method. MTL trains on multiple tasks simultaneously, and STL trains on single tasks independently.\\

\noindent
We summarize the architecture-based baselines as follows:
\begin{enumerate}[itemsep=0em, topsep=-1ex, itemindent=0em, leftmargin=1.2em, partopsep=0em]
    \item \textbf{PackNet}~\cite{mallya2018packnet}: iterative pruning and network re-training architecture for packing multiple tasks into a single network.
    \item \textbf{SupSup}~\cite{wortsman2020supermasks}: finding supermasks (subnetworks) within a randomly initialized network for each task in continual learning.
    \item \textbf{WSN / \textcolor{magenta}{SoftNet}}~(ours): jointly training model and finding task-adaptive subnetworks of novel/prior parameters for continual learning. Note that, in the inference step, \textcolor{magenta}{SoftNet} is acquired empirically by injecting small noises $U(0,\text{1e-3})$ to the backgrounds $\bm{0}$ of acquired task WSN while holding the foregrounds $\bm{1}$ of WSN. \\ 
\end{enumerate}

\begin{table*}[ht]
\small
\vspace{-0.1in}
\begin{center}
\caption{Performance comparison of the proposed method and baselines on various benchmark datasets. We report the mean and standard deviation of the average accuracy (ACC), average capacity (CAP), and average forward / backward transfer (FWT / BWT) across five independent runs. The best results are highlighted in bold. Values with $\dagger$ and $\ast$ denote reported performances from \cite{Saha2021} and \cite{Yoon2020}. We consider PackNet~\cite{mallya2018packnet} and SupSup~\cite{wortsman2020supermasks} as the baselines.}

\resizebox{\textwidth}{!}{
\begin{tabular}{lccccccccc}
\toprule
\multicolumn{1}{c}{\textbf{Method}}&\multicolumn{3}{c}{\textbf{Permuted MNIST}}&
\multicolumn{3}{c}{\textbf{5 Datasets}}&
\multicolumn{3}{c}{\textbf{Omniglot Rotation}} \\

\midrule
&ACC (\%)&CAP (\%)&FWT / BWT&
ACC (\%)&CAP (\%)&FWT / BWT&
ACC (\%)&CAP (\%)&FWT / BWT \\
\midrule
STL&
{97.37}~\scriptsize($\pm$ 0.01)&{$~$ 1,000.0}&- / -&
{93.44}~\scriptsize($\pm$ 0.12)&{$~$ 500.0}&- / -&
~~{82.13}~\scriptsize($\pm$ 0.08)$^\ast$ &{$~$ 10,000.0}& - / - \\
 \midrule
FINETUNE&
{78.22}~\scriptsize($\pm$ 0.84)&{$~$ 100.0}& - / {-0.21}~\scriptsize($\pm$ 0.01)&
{80.06}~\scriptsize($\pm$ 0.74)&{$~$ 100.0}& ~~ - / {-0.17}~\scriptsize($\pm$ 0.01)&
{44.48}~\scriptsize($\pm$ 1.68)&{$~$ 100.0}& - / {-0.45}~\scriptsize($\pm$ 0.02) \\

EWC~\cite{Kirkpatrick2017}&
{92.01}~\scriptsize($\pm$ 0.56)&~~{100.0}& - / {-0.03}~\scriptsize($\pm$ 0.00)&
~~{88.64}~\scriptsize($\pm$ 0.26)$^\dagger$&~~~~{100.0}$^\dagger$& ~~ - / {-0.04}~\scriptsize($\pm$ 0.01)$^\dagger$&~~{68.66}~\scriptsize($\pm$ 1.92)$^\ast$&~~~~{100.0}$^\ast$& - / {-} \\

HAT~\cite{Serra2018}&
{-}&{-}& - / {-}&
~~{91.32}~\scriptsize($\pm$ 0.18)$^\dagger$&~~~~{100.0}$^\dagger$&~~ - / {-0.03}~\scriptsize($\pm$ 0.00)$^\dagger$&
{-}&{-}& - / {-} \\

GPM~\cite{Saha2021}&
{94.96}~\scriptsize($\pm$ 0.07)&{$~$ 100.0}& - / {-0.02}~\scriptsize($\pm$ 0.01)&
~~{91.22}~\scriptsize($\pm$ 0.20)$^\dagger$&{$~$ 100.0}&~~ - / {-0.01}~\scriptsize($\pm$ 0.00)$^\dagger$&
{85.24}~\scriptsize($\pm$ 0.37)&{$~$ 100.0}& - / {-0.01}~\scriptsize($\pm$ 0.00) \\

\midrule
PackNet~\cite{mallya2018packnet}&
{96.37}~\scriptsize($\pm$ 0.1)&{$~$ 96.38}&{~-9.52}~\scriptsize($\pm$ 0.6) / \cellcolor{gg}\textbf{0.0}&
{92.81}~\scriptsize($\pm$ 0.1)&{$~$ 82.86}&{~-9.73}~\scriptsize($\pm$ 0.5) / \cellcolor{gg}\textbf{0.0}&
{30.70}~\scriptsize($\pm$ 1.5)&{$~$ 399.2}&{~-9.02}~\scriptsize($\pm$ 0.8) / \cellcolor{gg}\textbf{0.0} \\ 

SupSup~\cite{wortsman2020supermasks}&
{96.31}~\scriptsize($\pm$ 0.1)&{$~$ 122.89}~\scriptsize($\pm$ 0.1)&{~-9.63}~\scriptsize($\pm$ 0.7) / \cellcolor{gg}\textbf{0.0}&
{93.28}~\scriptsize($\pm$ 0.2)&{$~$ 104.27}~\scriptsize($\pm$ 0.2)&{~-9.52}~\scriptsize($\pm$ 0.6) / \cellcolor{gg}\textbf{0.0}&
{58.14}~\scriptsize($\pm$ 2.4)&{$~$ 407.12}~\scriptsize($\pm$ 0.2)&{~-8.87}~\scriptsize($\pm$ 0.6) / \cellcolor{gg}\textbf{0.0} \\ 

\midrule
WSN, $c=~~3\%$&
{94.84}~\scriptsize($\pm$ 0.1)&\textbf{{$~~~$ 19.87}~\scriptsize($\pm$ 0.2)}& {-10.25}~\scriptsize($\pm$ 0.0) / \cellcolor{gg}\textbf{0.0} &
{90.57}~\scriptsize($\pm$ 0.7)&\textbf{{$~~~$ 12.11}~\scriptsize($\pm$ 0.1)}& {~-9.62}~\scriptsize($\pm$ 0.6) / \cellcolor{gg}\textbf{0.0} & 
{80.68}~\scriptsize($\pm$ 2.6)&\textbf{{$~~~$ 75.87}~\scriptsize($\pm$ 1.2)}&{-8.33}~\scriptsize($\pm$ 0.1) / \cellcolor{gg}\textbf{0.0} \\ 

WSN, $c=~~5\%$&
{95.65}~\scriptsize($\pm$ 0.0)&{$~~~$ 26.49}~\scriptsize($\pm$ 0.2) & {~~-9.96}~\scriptsize($\pm$ 0.7) / \cellcolor{gg}\textbf{0.0}&
{91.61}~\scriptsize($\pm$ 0.2)&{$~~~$ 17.26}~\scriptsize($\pm$ 0.3)& {~-9.86}~\scriptsize($\pm$ 0.4) / \cellcolor{gg}\textbf{0.0}& 
\textbf{{87.28}~\scriptsize($\pm$ 0.7)}&{$~~~$ 79.85}~\scriptsize($\pm$ 1.2)& {-8.42}~\scriptsize($\pm$ 0.1) / \cellcolor{gg}\textbf{0.0} \\ 

WSN, $c=10\%$&
{96.14}~\scriptsize($\pm$ 0.0)&{$~~~$ 40.41}~\scriptsize($\pm$ 0.6)& {-10.14}~\scriptsize($\pm$ 1.4) / \cellcolor{gg}\textbf{0.0}&
{92.67}~\scriptsize($\pm$ 0.1)&{$~~~$ 28.01}~\scriptsize($\pm$ 0.3)&~\textbf{-9.33}~\scriptsize($\pm$ \textbf{0.4}) / \cellcolor{gg}\textbf{0.0}& 
{83.10}~\scriptsize($\pm$ 1.6)&{$~~~$ 83.08}~\scriptsize($\pm$ 1.6)& \textbf{-8.29}~\scriptsize($\pm$ \textbf{0.1}) / \cellcolor{gg}\textbf{0.0} \\ 

WSN, $c=30\%$&
\textbf{{96.41}~\scriptsize($\pm$ 0.1)}&{$~~~$ 77.73}~\scriptsize($\pm$ 0.4)& {~~-9.52}~\scriptsize($\pm$ 0.4) / {\cellcolor{gg}\textbf{0.0}}& 
{93.22}~\scriptsize($\pm$ 0.3)&{$~~~$ 62.30}~\scriptsize($\pm$ 0.7)& {~-9.68}~\scriptsize($\pm$ 0.6) / \cellcolor{gg}\textbf{0.0}& 
{81.89}~\scriptsize($\pm$ 1.2)&{$~$ ~~102.2}~\scriptsize($\pm$ 0.9) & {-8.34}~\scriptsize($\pm$ 0.1) / \cellcolor{gg}\textbf{0.0} \\ 

WSN, $c=50\%$&
{96.24}~\scriptsize($\pm$ 0.1)&{$~~~$ 98.10}~\scriptsize($\pm$ 0.3)& ~~\textbf{-9.39}~\scriptsize($\pm$ \textbf{0.8}) / \cellcolor{gg}\textbf{0.0}& 
\textbf{{93.41}~\scriptsize($\pm$ 0.1)}&{$~~~$ 86.10}~\scriptsize($\pm$ 0.6) & {~-9.51}~\scriptsize($\pm$ 0.4) / \cellcolor{gg}\textbf{0.0}& 
{79.80}~\scriptsize($\pm$ 2.2)&{$~$ ~~121.2}~\scriptsize($\pm$ 0.5) & {-8.33}~\scriptsize($\pm$ 0.1) / \cellcolor{gg}\textbf{0.0} \\ 


WSN, $c=70\%$&
{96.29}~\scriptsize($\pm$ 0.0)&{$~~~$ 102.56}~\scriptsize($\pm$ 0.1)& {~~-9.67}~\scriptsize($\pm$ 0.7) / \cellcolor{gg}\textbf{0.0}&  
{{92.38}~\scriptsize($\pm$ 0.5)}&{$~~~$ 98.73}~\scriptsize($\pm$ 0.4) & {~-9.53}~\scriptsize($\pm$ 0.4) / \cellcolor{gg}\textbf{0.0}& 
{79.02}~\scriptsize($\pm$ 2.5)&{~~~~140.6}~\scriptsize($\pm$ 0.3) & {-8.32}~\scriptsize($\pm$ 0.1) / \cellcolor{gg}\textbf{0.0} \\

\midrule 

\textcolor{magenta}{SoftNet}, $c=~~3\%$&
{94.84}~\scriptsize($\pm$ 0.1)&\textbf{{$~~~$ 19.87}~\scriptsize($\pm$ 0.2)}& \textbf{23.76}~\scriptsize($\pm$ \textbf{1.8}) / \cellcolor{gg}\textbf{0.0} &
{90.57}~\scriptsize($\pm$ 0.7)&\textbf{{$~~~$ 12.11}~\scriptsize($\pm$ 0.1)}& {36.08}~\scriptsize($\pm$ 7.1) /  \cellcolor{gg}\textbf{0.0} & 
{80.68}~\scriptsize($\pm$ 2.6)&\textbf{{$~~~$ 75.87}~\scriptsize($\pm$ 1.2)}&{51.52}~\scriptsize($\pm$ 4.3) / \cellcolor{gg}\textbf{0.0} \\ 

\textcolor{magenta}{SoftNet}, $c=~~5\%$&
{95.65}~\scriptsize($\pm$ 0.0)&{$~~~$ 26.49}~\scriptsize($\pm$ 0.2)&{23.44}~\scriptsize($\pm$ 2.0) / \cellcolor{gg}\textbf{0.0}&
{91.61}~\scriptsize($\pm$ 0.2)&{$~~~$ 17.26}~\scriptsize($\pm$ 0.3)& {40.19}~\scriptsize($\pm$ 4.6) / \cellcolor{gg}\textbf{0.0}& 
\textbf{{87.28}~\scriptsize($\pm$ 0.7)}&{$~~~$ 79.85}~\scriptsize($\pm$ 1.2)& \textbf{56.54}~\scriptsize($\pm$ \textbf{4.1}) / \cellcolor{gg}\textbf{0.0} \\ 

\textcolor{magenta}{SoftNet}, $c=10\%$&
{96.14}~\scriptsize($\pm$ 0.0)&{$~~~$ 40.41}~\scriptsize($\pm$ 0.6)& {22.12}~\scriptsize($\pm$ 1.6) / \cellcolor{gg}\textbf{0.0}&
{92.67}~\scriptsize($\pm$ 0.1)&{$~~~$ 28.01}~\scriptsize($\pm$ 0.3)&{38.06}~\scriptsize($\pm$ 9.6) / \cellcolor{gg}\textbf{0.0}& 
{83.10}~\scriptsize($\pm$ 1.6)&{$~~~$ 83.08}~\scriptsize($\pm$ 1.6)& {54.14}~\scriptsize($\pm$ 3.5) / \cellcolor{gg}\textbf{0.0} \\ 

\textcolor{magenta}{SoftNet}, $c=30\%$&
\textbf{{96.41}~\scriptsize($\pm$ 0.1)}&{$~~~$ 77.73}~\scriptsize($\pm$ 0.4)& {22.12}~\scriptsize($\pm$ 1.6) / {\cellcolor{gg}\textbf{0.0}}& 
{93.22}~\scriptsize($\pm$ 0.3)&{$~~~$ 62.30}~\scriptsize($\pm$ 0.7)& {40.83}~\scriptsize($\pm$ 5.7) / \cellcolor{gg}\textbf{0.0}& 
{81.89}~\scriptsize($\pm$ 1.2)&{$~$ ~~102.2}~\scriptsize($\pm$ 0.9) & {53.03}~\scriptsize($\pm$ 1.5) / \cellcolor{gg}\textbf{0.0} \\ 

\textcolor{magenta}{SoftNet}, $c=50\%$&
{96.24}~\scriptsize($\pm$ 0.1)&{$~~~$ 98.10}~\scriptsize($\pm$ 0.3)& {21.91}~\scriptsize($\pm$ 1.9) / \cellcolor{gg}\textbf{0.0}& 
\textbf{{93.41}~\scriptsize($\pm$ 0.1)}&{$~~~$ 86.10}~\scriptsize($\pm$ 0.6) & {40.81}~\scriptsize($\pm$ 7.2) / \cellcolor{gg}\textbf{0.0}& 
{79.80}~\scriptsize($\pm$ 2.2)&{$~$ ~~121.2}~\scriptsize($\pm$ 0.5) & {50.15}~\scriptsize($\pm$ 0.7) / \cellcolor{gg}\textbf{0.0} \\ 

\textcolor{magenta}{SoftNet}, $c=70\%$&
{96.29}~\scriptsize($\pm$ 0.0)&{$~~~$ 99.21}~\scriptsize($\pm$ 0.1)& {20.57}~\scriptsize($\pm$ 1.2) / \cellcolor{gg}\textbf{0.0}& 
{{92.38}~\scriptsize($\pm$ 0.5)}&{$~~~$ 98.73}~\scriptsize($\pm$ 0.4) & \textbf{41.04}~\scriptsize($\pm$ \textbf{5.2}) / \cellcolor{gg}\textbf{0.0}& 
{79.02}~\scriptsize($\pm$ 2.5)&{~~~~140.6}~\scriptsize($\pm$ 0.3) & {49.62}~\scriptsize($\pm$ 1.7) / \cellcolor{gg}\textbf{0.0} \\

\midrule
MTL&
~~{96.70~\scriptsize($\pm$ 0.02)$^\dagger$}&~~{100.0}&- / {-}&
~~{91.54~\scriptsize($\pm$ 0.28)$^\dagger$}&~~{100.0}&- / {-}&
{81.23~\scriptsize($\pm$ 0.52)}&~~{100.0}& - / {-} \\
\bottomrule
\end{tabular}}
\label{tab:main_table_new}
\vspace{-0.2in}
\end{center}
\end{table*}

\noindent
\textbf{Experimental settings.} \label{exp_setting} 
As we directly implement our method from the official code of \cite{Saha2021}, we provide the values for HAT and GPM reported in \cite{Saha2021}. For Omniglot Rotation and Split CIFAR-100 Superclass, we deploy the proposed architecture in multi-head settings with hyperparameters as reported in \cite{Yoon2020}. All our experiments run on a single-GPU setup of NVIDIA V100. We provide more details of the datasets, architectures, and experimental settings, including the hyperparameter configurations for all methods in \Cref{app_sec:exper_detail}.   

\begin{figure}[ht]
    \centering
    \vspace{-0.1in}
    \renewcommand{\arraystretch}{1}
    \setlength{\tabcolsep}{0pt}{%
    \begin{tabular}{cc}
    \includegraphics[width=.5\columnwidth]{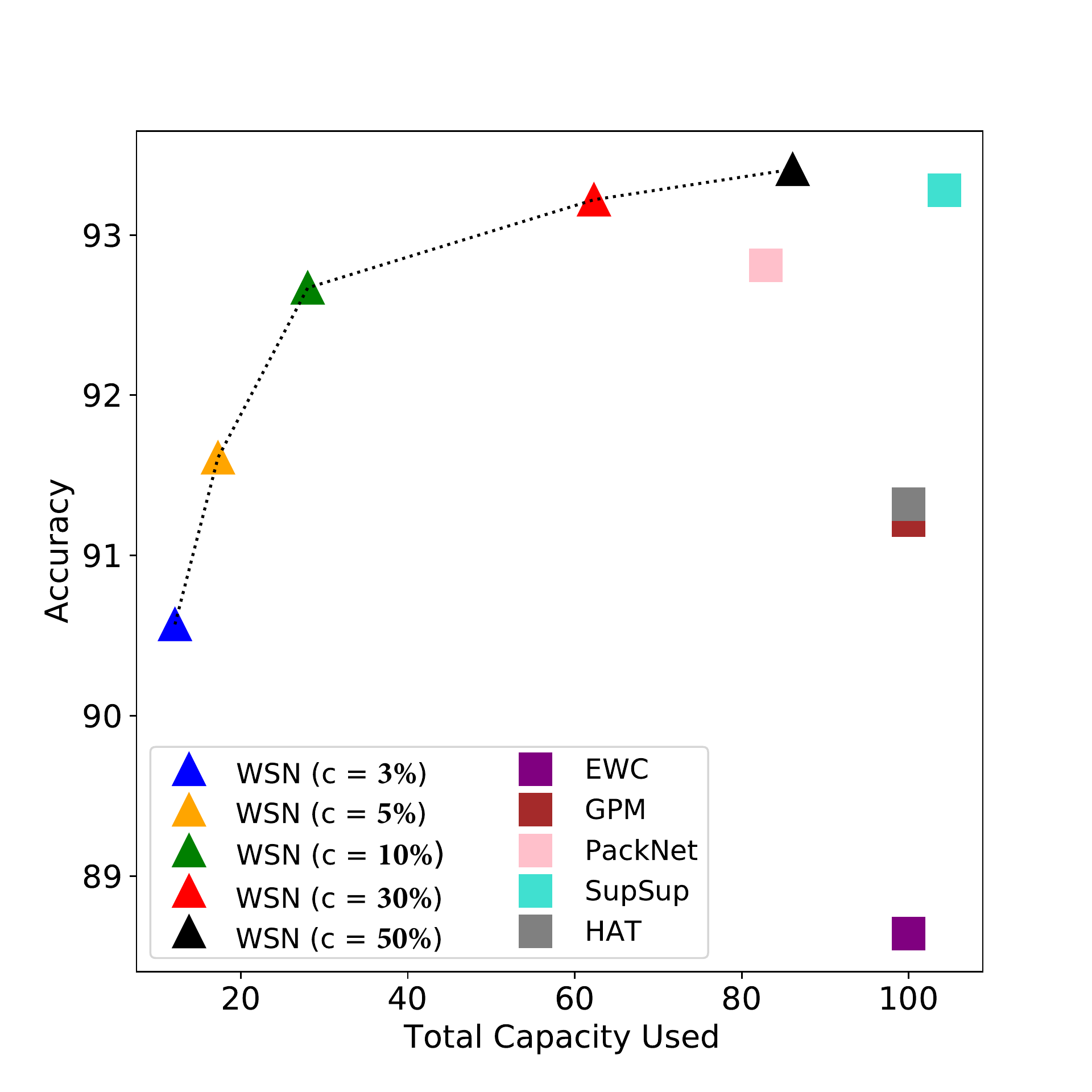} &
    \includegraphics[width=.5\columnwidth]{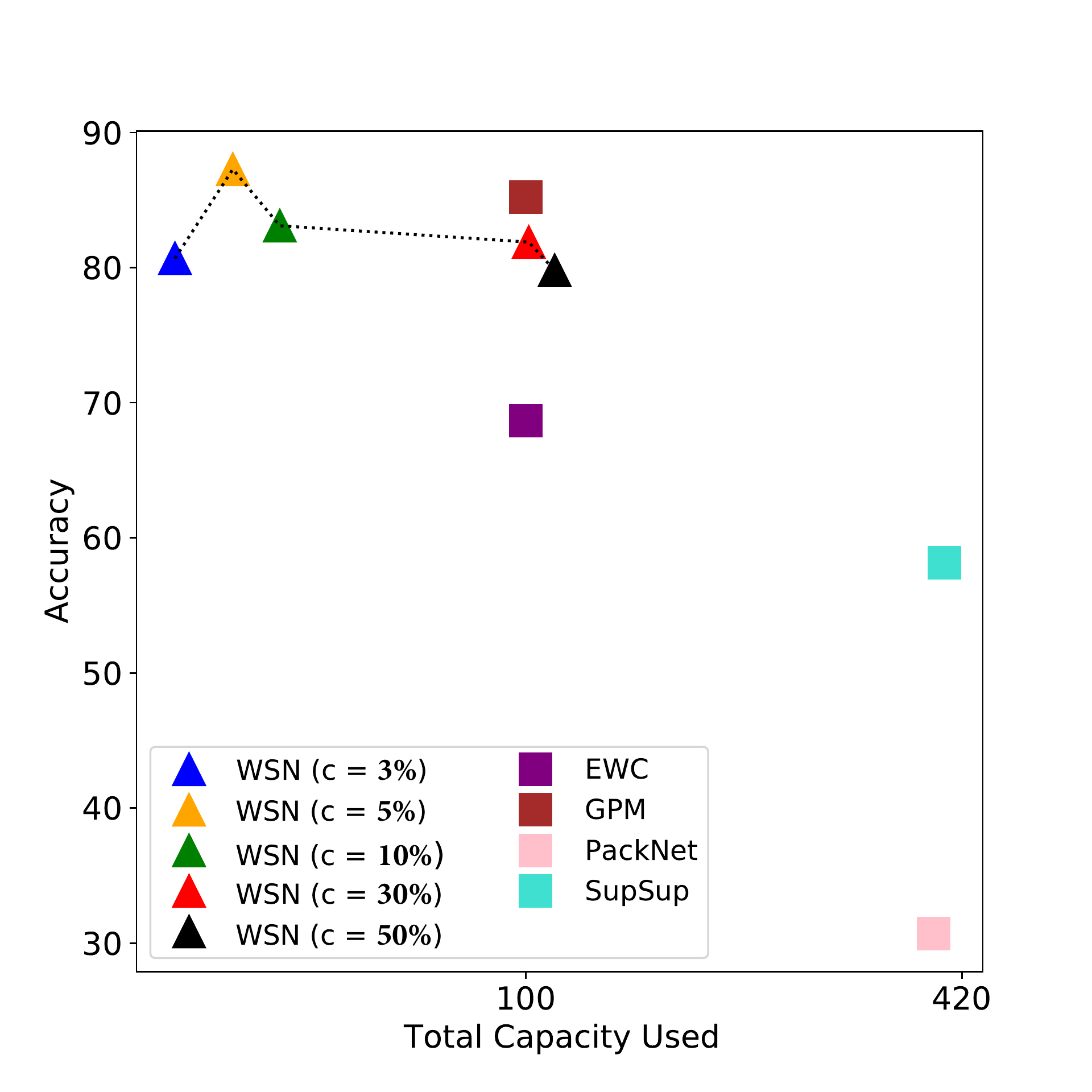} \\
    \small (a) 5-Dataset - ResNet-18 & \small (b) Omniglot Rotation - LeNet\\
    \end{tabular}}
    \caption{\footnotesize \textbf{Accuracy over Total Capacity Usage}: The performances of Hard-WSN gets better than others as total model capacities increase, and then, saturated approximately at 80 \%.}
    
    \label{fig:acc_vs_cap}
\vspace{-0.1in}
\end{figure}

\noindent 
\textbf{Performance metrics.} We evaluate all methods based on the following four metrics: 
\begin{enumerate}[itemsep=0em, topsep=-1ex, itemindent=0em, leftmargin=1.2em, partopsep=0em]
\item {\textbf{Accuracy (ACC)}} measures the average of the final classification accuracy on all tasks: $\mathrm{ACC}=\frac{1}{T} \sum_{i=1}^{T} A_{T, i}$, where $A_{T,i}$ is the test accuracy for task $i$ after training on task $T$.

\item {\textbf{Capacity (CAP)}} measures the total percentage of non-zero weights plus the prime masks for all tasks as follows: $\mathrm{Capacity}= (1-\mathcal{S}) + \frac{(1-\alpha) T}{32}$, where we assume all task weights of 32-bit precision. $\mathcal{S}$ is the sparsity of $\mathbf{M}_{T}$ and the average compression rate $\alpha \approx 0.78$ that we acquired through 7bit Huffman encoding, which depends on the size of bit-binary maps; the compression rate $\alpha$ also depends on compression methods typically.   

\item {\textbf{Forward Transfer (FWT)}}  measures how much the representations that we learned so far help learn new tasks, namely: $FWT = \frac{1}{T} \sum_{i=2}^{T} A_{i-1, i} - R_i$ where $R_i$ is the accuracy of a randomly initialized network on task $i$.

\item {\textbf{Backward Transfer (BWT)}} measures the forgetting during continual learning. Negative BWT means that learning new tasks causes the forgetting on past tasks: $\mathrm{BWT}=\frac{1}{T-1}\sum_{i=1}^{T-1} A_{T, i}-A_{i, i}$.

\end{enumerate}

\subsection{Few-shot Class Incremental Learning (FSCIL)}
We introduce experimental setups - Few-Shot Class Incremental Learning (FSCIL) settings to provide soft-subnetworks' effectiveness. We empirically evaluate and compare our soft-subnetworks with state-of-the-art methods and vanilla subnetworks in the following subsections. \\

\noindent
\textbf{Datasets.} To validate the effectiveness of the soft-subnetwork, we follow the standard FSCIL experimental setting. We randomly select 60 classes as the base and 40 as new classes for CIFAR-100 and miniImageNet. In each incremental learning session, we construct 5-way 5-shot tasks by randomly picking five classes and sampling five training examples for each class. \\

\noindent
\textbf{Baselines.} We mainly compare our SoftNet with architecture-based methods for FSCIL: FSLL~\cite{mazumder2021few} that selects important parameters for each session, and HardNet, representing a binary subnetwork. Furthermore, we compare other FSCIL methods such as iCaRL~\cite{rebuffi2017icarl}, Rebalance~\cite{hou2019learning}, TOPIC~\cite{tao2020few}, IDLVQ-C~\cite{chen2020incremental}, and F2M~\cite{shi2021overcoming}. We also include a joint training method~\cite{shi2021overcoming} that uses all previously seen data, including the base and the following few-shot tasks for training as a reference. Furthermore, we fix the classifier re-training method (cRT)~\cite{kang2019decoupling} for long-tailed classification trained with all encountered data as the approximated upper bound. \\

\noindent
\textbf{Experimental settings.} The experiments are conducted with NVIDIA GPU RTX8000 on CUDA 11.0. We also randomly split each dataset into multiple sessions. We run each algorithm ten times for each dataset and report their mean accuracy. We adopt ResNet18~\cite{he2016deep} as the backbone network. For data augmentation, we use standard random crop and horizontal flips. In the base session training stage, we select top-$c\%$ weights at each layer and acquire the optimal soft-subnetworks with the best validation accuracy. In each incremental few-shot learning session, the total number of training epochs is $6$, and the learning rate is $0.02$. We train new class session samples using a few minor weights of the soft-subnetwork (Conv4x layer of ResNet18) obtained by the base session learning. We specify further experiment details in Appendix. 

\section{Results On Task-Incremental Learning}
\subsection{Comparisons with Baselines}
We evaluate our algorithm on three-standard benchmark datasets: Permuted MNIST, 5-Datasets, and Omniglot Rotation. We set PackNet and SupSup to baselines as non-reused weight methods and compared WSN with the baselines, including other algorithms as shown in \Cref{tab:main_table_new}. Our WSN outperformed all the baselines in three measurements, achieving the best average accuracy of 96.41\%, 93.41\%, and 87.28\% while using the least capacity compared to the other existing methods, respectively. Moreover, our WSN was proved to be a forget-free model like PackNet. Compared with PackNet, however, WSN showed the effectiveness of reused weights for continual learning and its scalability in the Omniglot Rotation experiments with the least capacities by a large margin. \Cref{fig:acc_vs_cap} provides the accuracy over total capacity usage. Our WSN's accuracy is higher than PackNet's, approximately at 80\% total capacity usage on 5-Dataset, and WSN outperformed others at 80\% total capacity usage on Omniglot Rotation. The lower performances of PackNet might attribute to Omniglot Rotation dataset statistics since, regardless of random rotations, tasks could share visual features in common such as circles, curves, and straight lines. Therefore, non-reused methods might be brutal to train a new task model unless prior weights were transferred to the current model, a.k.a. forward transfer learning. To show the WSN's power of forward transferring knowledge, SoftNet was prepared by initializing zero-part of masks to small random perturbations. Then, we observed that, in the inference step, SoftNet showed great power to transfer knowledge while maintaining WSN's performances of ACC and CAP.    

\subsection{Comparisons with the SOTA in TIL}
\begin{table*}[ht]
\begin{center}
\caption{Performance comparisons of the proposed method and other state-of-the-art including baselines - PackNet~\cite{mallya2018packnet} and SupSup~\cite{wortsman2020supermasks} - on various benchmark datasets. We report the mean and standard deviation of the average accuracy (ACC), average capacity (CAP), and average forward / backward transfer (FWT / BWT) across $5$ independent runs with five seeds under the same experimental setup \cite{deng2021flattening}. The best results are highlighted in bold. Also, $\dagger$ $~$ denotes results reported from \cite{deng2021flattening}.}

\resizebox{\textwidth}{!}{
\begin{tabular}{lccccccccc}
\toprule
\multicolumn{1}{c}{\textbf{Method}}&\multicolumn{3}{c}{\textbf{CIFAR-100 Split}}&
\multicolumn{3}{c}{\textbf{CIFAR-100 Superclass}}&
\multicolumn{3}{c}{\textbf{TinyImageNet}} \\

\midrule
&ACC (\%)&CAP (\%)& FWT / BWT (\%)&
ACC (\%)&CAP (\%)& FWT / BWT (\%)&
ACC (\%)&CAP (\%)&FWT / BWT (\%) \\
\midrule


La-MaML \cite{gupta2020maml} & 71.37~\scriptsize($\pm$ 0.7)$^\dagger$ & 100.0 & - / -5.39~\scriptsize($\pm$ 0.5)$^\dagger$ 
& 54.44~\scriptsize($\pm$ 1.4)$^\dagger$ & 100.0  & - / -6.65~\scriptsize($\pm$ 0.9)$^\dagger$ & 66.90~\scriptsize($\pm$ 1.7)$^\dagger$ & 100.0  & - / -9.13~\scriptsize($\pm$ 0.9)$^\dagger$  \\


GPM \cite{Saha2021} & 73.18~\scriptsize($\pm$ 0.5)$^\dagger$ & 100.0 & - / -1.17~\scriptsize($\pm$ 0.3)$^\dagger$ & 57.33~\scriptsize($\pm$ 0.4)$^\dagger$ & 100.0  & - /  -0.37~\scriptsize($\pm$ 0.1)$^\dagger$ & 67.39~\scriptsize($\pm$ 0.5)$^\dagger$ & 100.0  & - / ~\textbf{1.45}~\scriptsize($\pm$ \textbf{0.2})$^\dagger$  \\


FS-DGPM \cite{deng2021flattening} & 74.33~\scriptsize($\pm$ 0.3)$^\dagger$ & 100.0 & 
 - / -2.71~\scriptsize($\pm$ 0.2)$^\dagger$ & 58.81~\scriptsize($\pm$ 0.3)$^\dagger$  & 100.0  & - / -2.97~\scriptsize($\pm$ 0.4)$^\dagger$ & 70.41~\scriptsize($\pm$ 1.3)$^\dagger$  & 100.0  & - / -2.11~\scriptsize($\pm$ 0.9)$^\dagger$ \\
 
\midrule




PackNet \cite{mallya2018packnet} & 72.39~\scriptsize($\pm$ 0.4) & ~~96.38 \scriptsize($\pm$ 0.0) & {~0.12}~\scriptsize($\pm$ 0.6) / \cellcolor{gg}\textbf{0.0} & 58.78~\scriptsize($\pm$ 0.5) & 126.65 \scriptsize($\pm$ 0.0) & {~~0.56}~\scriptsize($\pm$ 0.8) / \cellcolor{gg}\textbf{0.0} & 55.46~\scriptsize($\pm$ 1.2) & 188.67 \scriptsize($\pm$ 0.0) & {~-0.44}~\scriptsize($\pm$ 0.5) / \cellcolor{gg}\textbf{0.0} \\

SupSup \cite{wortsman2020supermasks} & 75.47~\scriptsize($\pm$ 0.3) & 129.00 \scriptsize($\pm$ 0.1) & {~0.06}~\scriptsize($\pm$ 0.5) / \cellcolor{gg}\textbf{0.0} & 61.70~\scriptsize($\pm$ 0.3) & 162.49~\scriptsize($\pm$ 0.0) & {~-0.50}~\scriptsize($\pm$ 0.6) / \cellcolor{gg}\textbf{0.0} & 59.60~\scriptsize($\pm$ 1.1) & 214.52~\scriptsize($\pm$ 0.9) & {~-0.82}~\scriptsize($\pm$ 0.6) / \cellcolor{gg}\textbf{0.0} \\


\midrule

WSN, $c=~~3\%$ & 72.23~\scriptsize($\pm$ 0.3) & ~~\textbf{18.56}~\scriptsize($\pm$ \textbf{0.3}) & ~\textbf{0.13}~\scriptsize($\pm$ \textbf{0.2}) / \cellcolor{gg}\textbf{0.0} & 54.99~\scriptsize($\pm$ 0.7) & ~~\textbf{22.30}~\scriptsize($\pm$ \textbf{0.2}) & ~\textbf{0.15}~\scriptsize($\pm$ \textbf{0.9}) / \cellcolor{gg}\textbf{0.0} & 68.72~\scriptsize($\pm$ 1.6) & ~~\textbf{37.19}~\scriptsize($\pm$ \textbf{0.2})  & -0.49~\scriptsize($\pm$ 0.3) / \cellcolor{gg}\textbf{0.0}  \\

WSN, $c=~~5\%$ & 73.56~\scriptsize($\pm$ 0.2) & ~~25.09~\scriptsize($\pm$ 0.4) & ~0.05~\scriptsize($\pm$ 0.5) / \cellcolor{gg}\textbf{0.0} & 57.99~\scriptsize($\pm$ 1.3) & ~~27.37~\scriptsize($\pm$ 0.3)  & ~0.59~\scriptsize($\pm$ 0.6) / \cellcolor{gg}\textbf{0.0} & 71.22~\scriptsize($\pm$ 0.9) & ~~41.98~\scriptsize($\pm$ 0.5)  & -0.73~\scriptsize($\pm$ 0.3) / \cellcolor{gg}\textbf{0.0}  \\

WSN, $c=10\%$ & 75.46~\scriptsize($\pm$ 0.1) & ~~39.87~\scriptsize($\pm$ 0.6) & ~0.07~\scriptsize($\pm$ 0.3) /  \cellcolor{gg}\textbf{0.0} & 60.45~\scriptsize($\pm$ 0.4) & ~~38.55~\scriptsize($\pm$ 0.2)  & -0.35~\scriptsize($\pm$ {0.5}) / \cellcolor{gg}\textbf{0.0} & \textbf{71.96}~\scriptsize($\pm$ \textbf{1.4}) & ~~48.65~\scriptsize($\pm$ 3.0)  & -0.63~\scriptsize($\pm$ 0.4) / \cellcolor{gg}\textbf{0.0}  \\

WSN, $c=30\%$ & 77.12~\scriptsize($\pm$ 0.3) & ~~80.26~\scriptsize($\pm$ 1.5) & -0.02~\scriptsize($\pm$ 0.5) / \cellcolor{gg}\textbf{0.0} & 61.47~\scriptsize($\pm$ 0.3) & ~~63.47~\scriptsize($\pm$ 1.3) & -0.49~\scriptsize($\pm$ {0.5}) / \cellcolor{gg}\textbf{0.0} & 70.92~\scriptsize($\pm$ 1.3) & ~~73.44~\scriptsize($\pm$ 2.4)  & -\textbf{0.32}~\scriptsize($\pm$ \textbf{0.1}) / \cellcolor{gg}\textbf{0.0}  \\

WSN, $c=50\%$ & 77.46~\scriptsize($\pm$ 0.4) & ~~99.13~\scriptsize($\pm$ 0.5) & -0.04~\scriptsize($\pm$ 0.3) / \cellcolor{gg}\textbf{0.0} & \textbf{61.79}~\scriptsize($\pm$ \textbf{0.2}) & ~~80.93~\scriptsize($\pm$ 1.6) & -0.26~\scriptsize($\pm$ {0.7}) / \cellcolor{gg}\textbf{0.0} & 69.06~\scriptsize($\pm$ 0.8) & ~~92.03~\scriptsize($\pm$ 1.8)  & -0.33~\scriptsize($\pm$ 0.1) / \cellcolor{gg}\textbf{0.0}  \\

WSN, $c=70\%$ & \textbf{78.08}~\scriptsize($\pm$ \textbf{0.3}) & 105.77~\scriptsize($\pm$ 0.2) & -0.06~\scriptsize($\pm$ 0.1) / \cellcolor{gg}\textbf{0.0} & 61.24~\scriptsize($\pm$ 0.2) & ~~97.40~\scriptsize($\pm$ 1.0) & -0.16~\scriptsize($\pm$ {1.2}) / \cellcolor{gg}\textbf{0.0} & 67.18~\scriptsize($\pm$ 1.6) & 108.82~\scriptsize($\pm$ 1.3) & -0.37~\scriptsize($\pm$ 0.4) / \cellcolor{gg}\textbf{0.0}  \\

\midrule

\textcolor{magenta}{SoftNet}, $c=~~3\%$ & {72.23}~\scriptsize($\pm$ {0.3}) & ~~\textbf{18.56}~\scriptsize($\pm$ \textbf{0.3}) & \textbf{52.07}~\scriptsize($\pm$ \textbf{0.5}) / \cellcolor{gg}\textbf{0.0}  & 55.00~\scriptsize($\pm$ 0.7) & ~~\textbf{22.30}~\scriptsize($\pm$ \textbf{0.2}) & \textbf{34.88}~\scriptsize($\pm$ \textbf{1.1}) / \cellcolor{gg}\textbf{0.0} & 68.72~\scriptsize($\pm$ 1.6) & ~~\textbf{37.19}~\scriptsize($\pm$ \textbf{0.2})  & 49.98~\scriptsize($\pm$ 1.4) / \cellcolor{gg}\textbf{0.0}  \\

\textcolor{magenta}{SoftNet}, $c=~~5\%$ & {73.57}~\scriptsize($\pm$ {0.2}) & ~~25.09~\scriptsize($\pm$ 0.4) & 51.70~\scriptsize($\pm$ 0.5) /  \cellcolor{gg}\textbf{0.0} & 57.99~\scriptsize($\pm$ 1.3) & ~~27.37~\scriptsize($\pm$ 0.3) & 34.76~\scriptsize($\pm$ 1.1) / \cellcolor{gg}\textbf{0.0} & 71.22~\scriptsize($\pm$ 0.9) & ~~41.98~\scriptsize($\pm$ 0.5) & \textbf{50.83}~\scriptsize($\pm$ 0.6) / \cellcolor{gg}\textbf{0.0}  \\

\textcolor{magenta}{SoftNet}, $c=10\%$ & {75.47}~\scriptsize($\pm$ {0.1}) & ~~39.87~\scriptsize($\pm$ 0.6) & 51.58~\scriptsize($\pm$ 0.6) / \cellcolor{gg}\textbf{0.0} & 60.46~\scriptsize($\pm$ 0.4) & ~~38.55~\scriptsize($\pm$ 0.2) & 32.21~\scriptsize($\pm$ 0.9) / \cellcolor{gg}\textbf{0.0} & \textbf{71.97}~\scriptsize($\pm$ \textbf{1.4}) & ~~48.65~\scriptsize($\pm$ 3.0) & 48.86~\scriptsize($\pm$ 1.6) / \cellcolor{gg}\textbf{0.0}  \\

\textcolor{magenta}{SoftNet}, $c=30\%$ & {77.12}~\scriptsize($\pm$ {0.3}) & ~~80.26~\scriptsize($\pm$ 1.5) & 51.64~\scriptsize($\pm$ 0.9) / \cellcolor{gg}\textbf{0.0} & 61.47~\scriptsize($\pm$ 0.3) & ~~63.47~\scriptsize($\pm$ 1.3) & 32.04~\scriptsize($\pm$ 1.3) / \cellcolor{gg}\textbf{0.0} & 70.92~\scriptsize($\pm$ 1.4) & ~~73.44~\scriptsize($\pm$ 2.4) & 47.54~\scriptsize($\pm$ 1.0) / \cellcolor{gg}\textbf{0.0}  \\

\textcolor{magenta}{SoftNet}, $c=50\%$ & {77.46}~\scriptsize($\pm$ {0.4}) & ~~99.13~\scriptsize($\pm$ 0.5) & 52.06~\scriptsize($\pm$ 1.3) / \cellcolor{gg}\textbf{0.0} & \textbf{61.80}~\scriptsize($\pm$ \textbf{0.2}) & ~~80.93~\scriptsize($\pm$ 1.6) & 30.40~\scriptsize($\pm$ 0.7) / \cellcolor{gg}\textbf{0.0} & 69.06~\scriptsize($\pm$ 0.8) & ~~92.03~\scriptsize($\pm$ 1.8) & 47.80~\scriptsize($\pm$ 1.1) / \cellcolor{gg}\textbf{0.0}  \\

\textcolor{magenta}{SoftNet}, $c=70\%$ & \textbf{78.08}~\scriptsize($\pm$ \textbf{0.3}) & ~105.77~\scriptsize($\pm$ 0.2) & 54.07~\scriptsize($\pm$ 0.5) / \cellcolor{gg}\textbf{0.0} & 61.24~\scriptsize($\pm$ 0.2) & ~~97.40~\scriptsize($\pm$ 1.0) & 30.94~\scriptsize($\pm$ \textbf{1.4}) / \cellcolor{gg}\textbf{0.0} & 67.19~\scriptsize($\pm$ 1.6) & 108.82~\scriptsize($\pm$ 1.3) & 48.27~\scriptsize($\pm$ 0.9) / \cellcolor{gg}\textbf{0.0}  \\



\midrule 


Multitask & ~~79.75~\scriptsize($\pm$ 0.4)$^\dagger$ & 100.0 & - / - & ~~61.00~\scriptsize($\pm$ 0.2)$^\dagger$ & 100.0 & - / - & ~~77.10~\scriptsize($\pm$ 1.1)$^\dagger$ & 100.0 & - / - \\

\bottomrule
\end{tabular}}
\label{tab:main_sota_table}
\end{center}\vspace{-0.2in}
\end{table*}

\begin{figure}[t]
    \centering
    \setlength{\tabcolsep}{-7pt}{%
    \begin{tabular}{cc}

    \includegraphics[height=2.96cm,trim={0.3cm 0.1cm 0.1cm 0.1cm},clip]{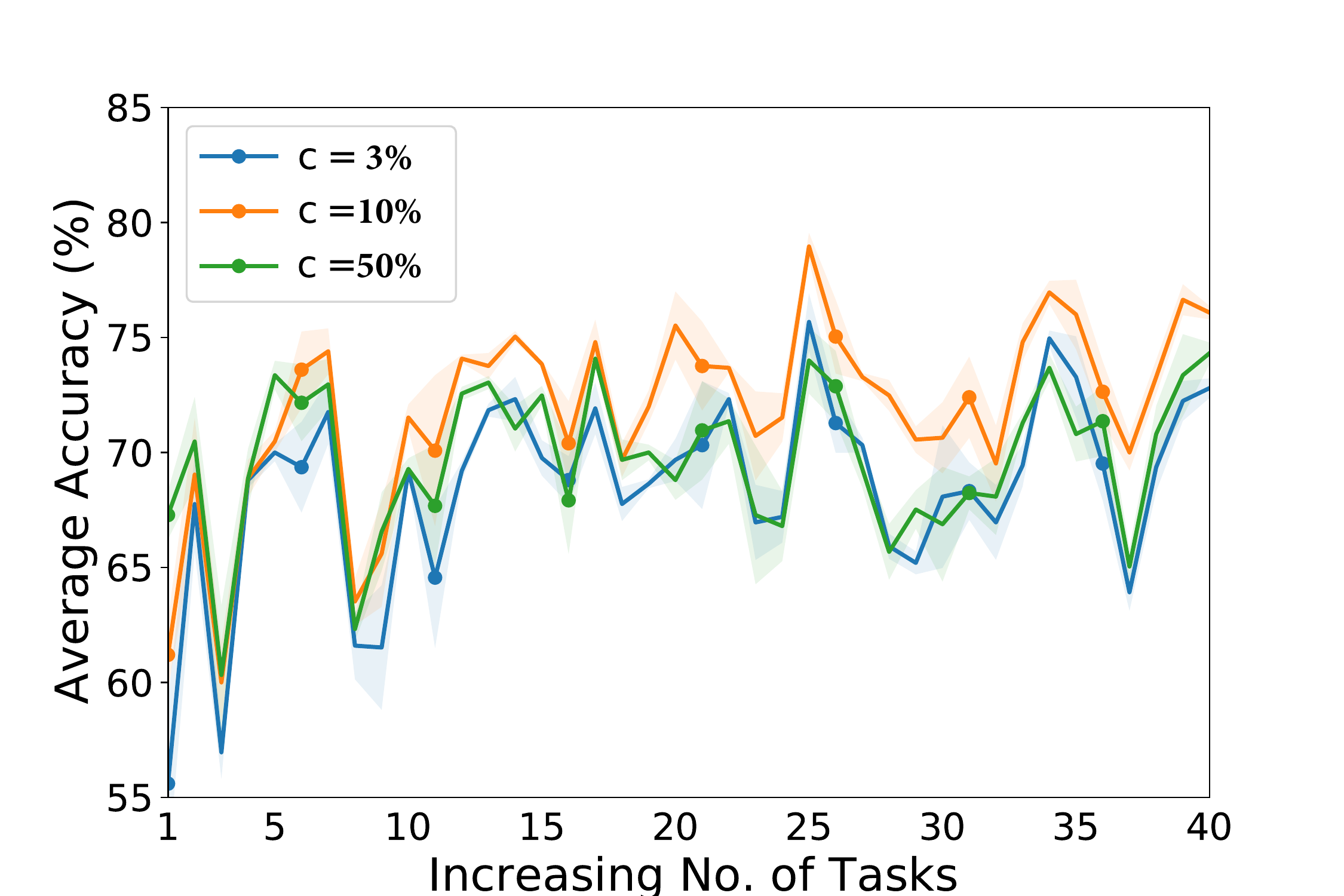} &\;\; 
    \includegraphics[height=2.96cm,trim={0.1cm 0.1cm 0.1cm 0.1cm},clip]{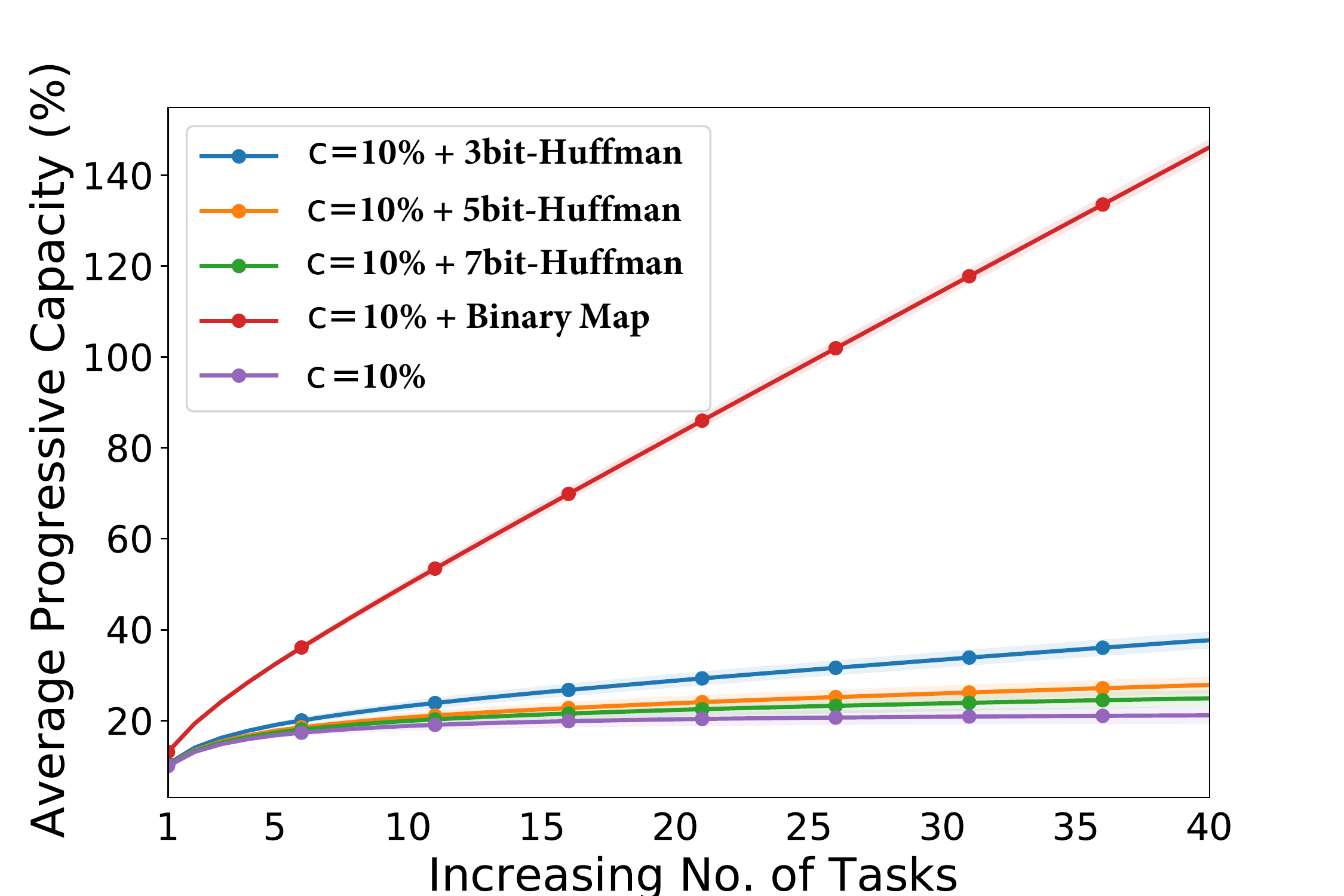} \\
    
    \small (a) Per Task Accuracy over $c$ & \small (b) Model Capacity
    \end{tabular}}
    \caption{\textbf{WSN's Performances and Compressed Capacities}: Sequence of TinyImageNet Dataset Experiments: (a) Setting $c = 10 \%$ shows generalized performances over others and (b) Within 40 tasks, the 7-bits compressed model increase its capacity the least.}
    \label{fig:main_tinyimg_plots}
    \vspace{-0.1in}
\end{figure}
\begin{figure*}[ht]
    \centering
    \setlength{\tabcolsep}{-10pt}{%
    \begin{tabular}{cccc}

    \includegraphics[height=3.3cm,trim={0.1cm 0.1cm 0.1cm 0.1cm},clip]{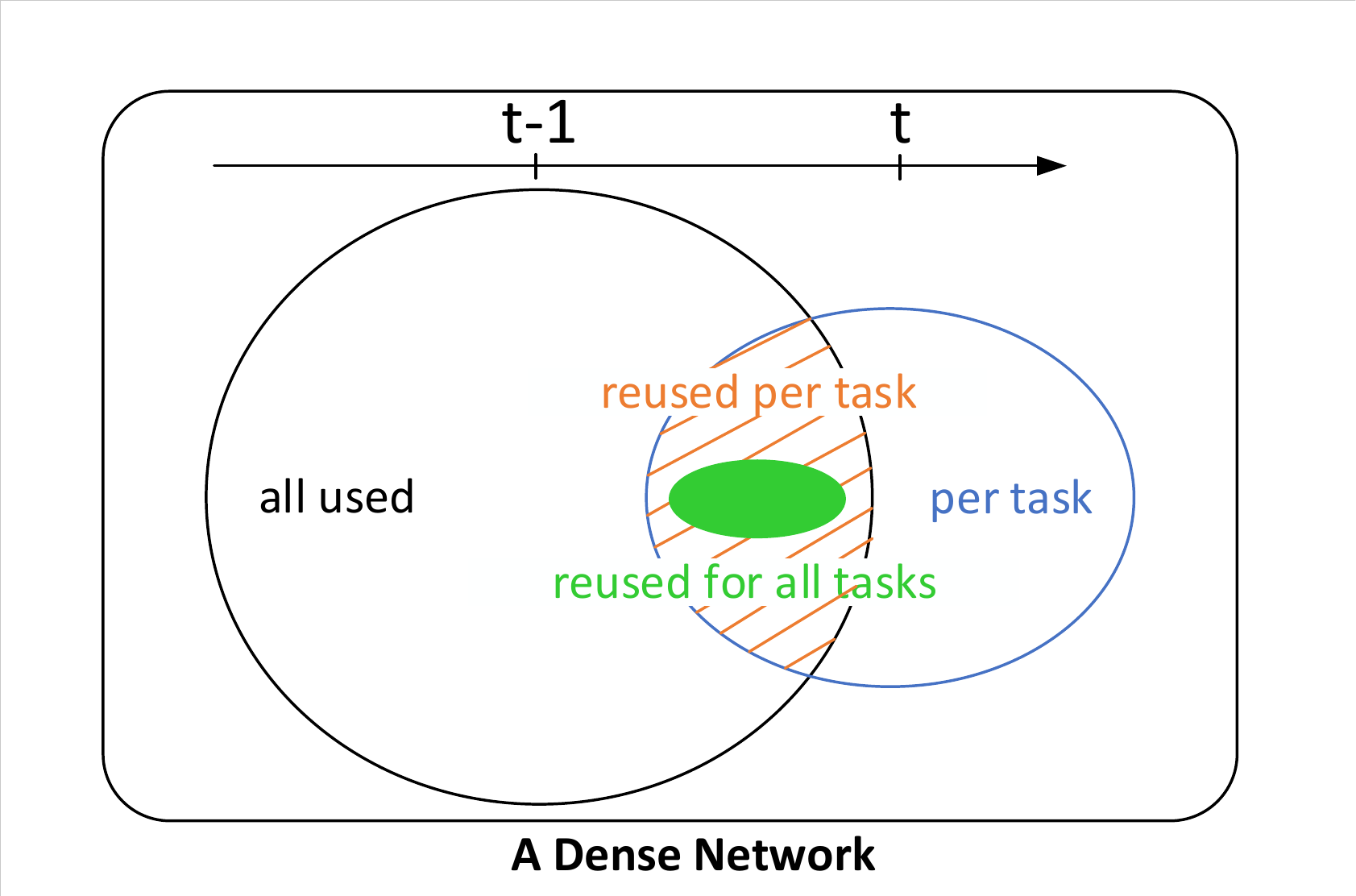} & ~~~ 
    \includegraphics[height=3.3cm,trim={0.1cm 0.1cm 0.1cm 0.1cm},clip]{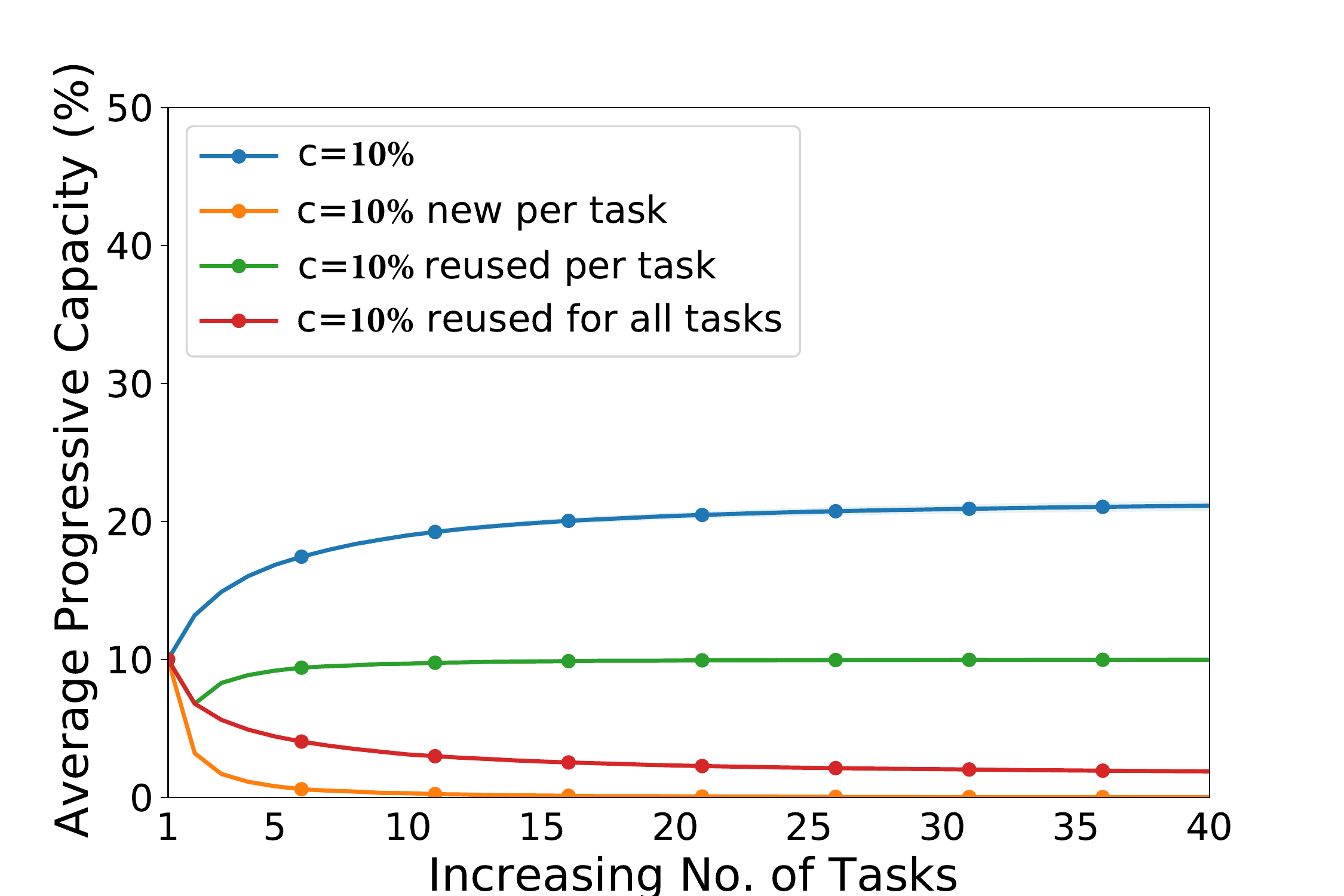} & ~~~ 
    \includegraphics[height=3.3cm,trim={0.1cm 0.1cm 0.1cm 0.1cm},clip]{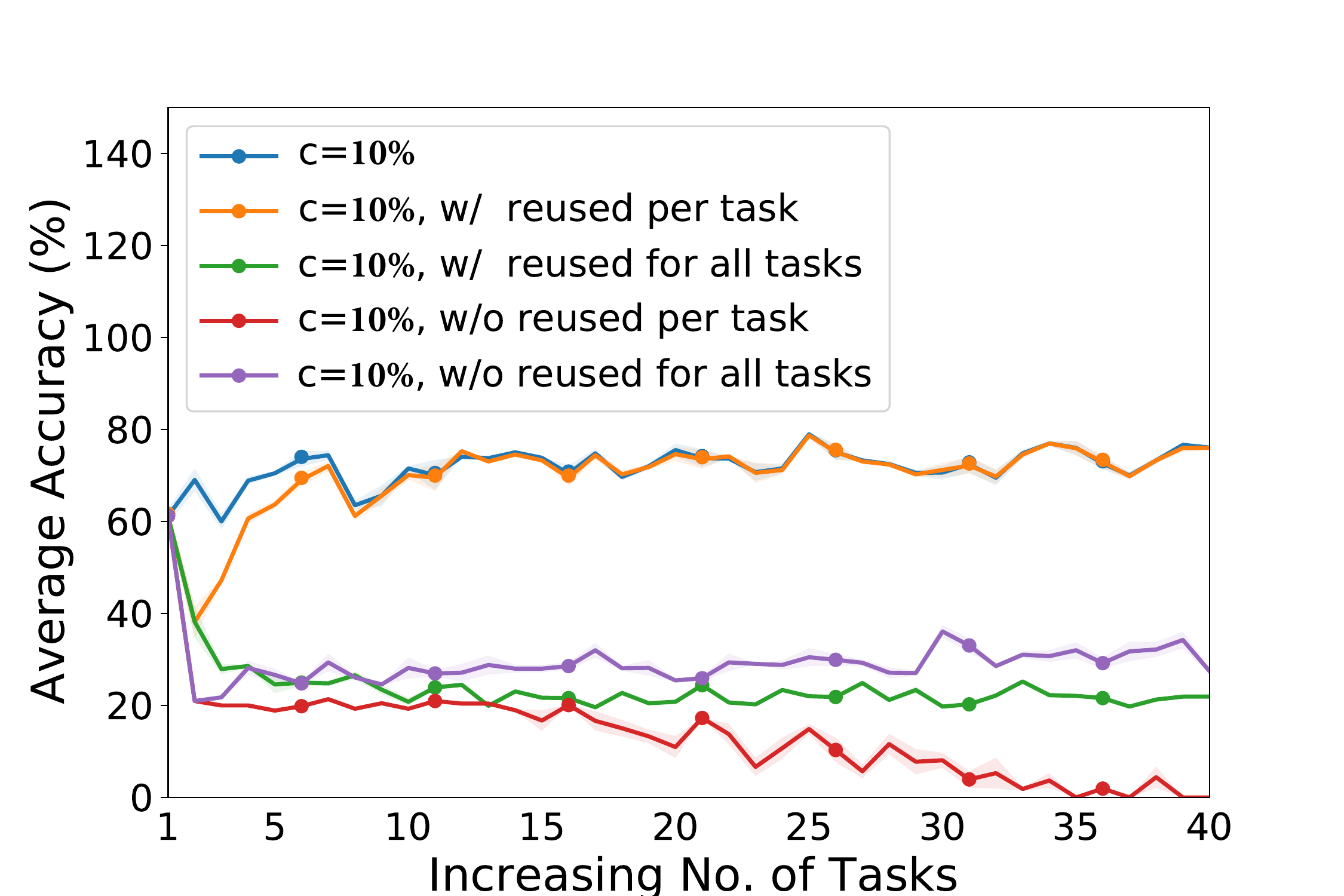} & ~~~
    \includegraphics[height=3.3cm,trim={0.1cm 0.1cm 0.1cm 0.1cm},clip]{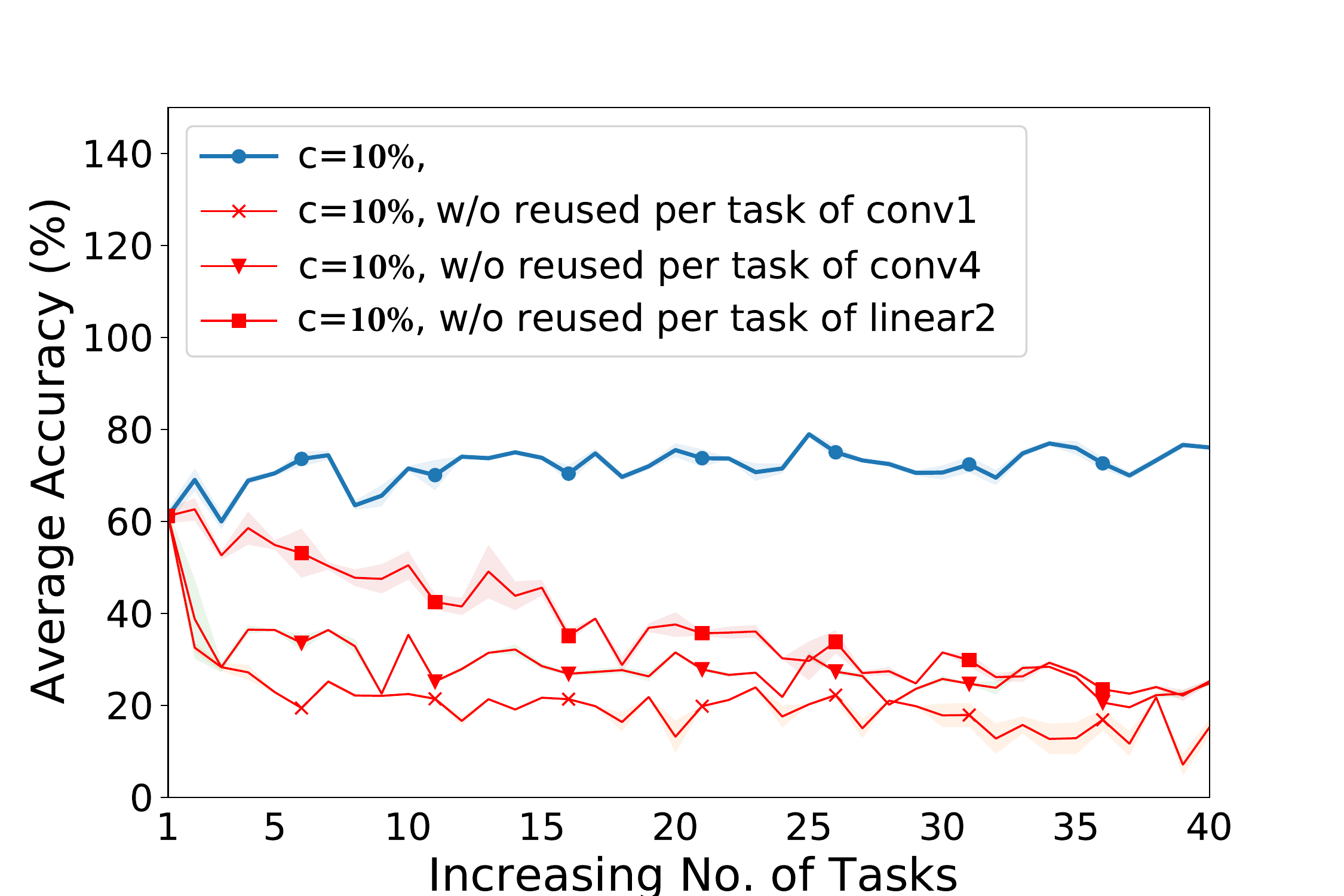} \\ 
    
    
    
    \small (a) A Diagram of Capacities 
    & \small (b) Capacities over $c$ 
    & \small (c) Acc. of Reused Weights 
    & \small (d) Acc. w/o Reused Weights \\
    \vspace{-0.2in}
    \end{tabular}
    }
    
    \caption{\textbf{Layer-wise Analysis of WSN} on TinyImageNet Dataset Experiments: (a) Weights reusability within a dense network, (b) Capacities except to binary maps are determined by $c = 10\%$, (c) The most significant forgetting occurs from weights without reused per task; the significance of weights reused for all tasks gets lower, and from task 7, all used weights seem to be enough to infer all tasks, and (d) In the inference step, we inspected a layer-wise forgetting caused by removing (setting masking value 1 to 0) reused weights per task on the trained model and observed performance drops significantly at Conv1 layer.}
    
    \label{fig:main_tinyimg_all_acc_plots}
    \vspace{-0.15in}
\end{figure*}

\begin{figure}[t]
    \vspace{-0.1in}
    \centering
    \begin{adjustbox}{width=\columnwidth}
    \includegraphics[height=6.0cm, trim={1.5cm 0.1cm 1.5cm 0.1cm},clip]{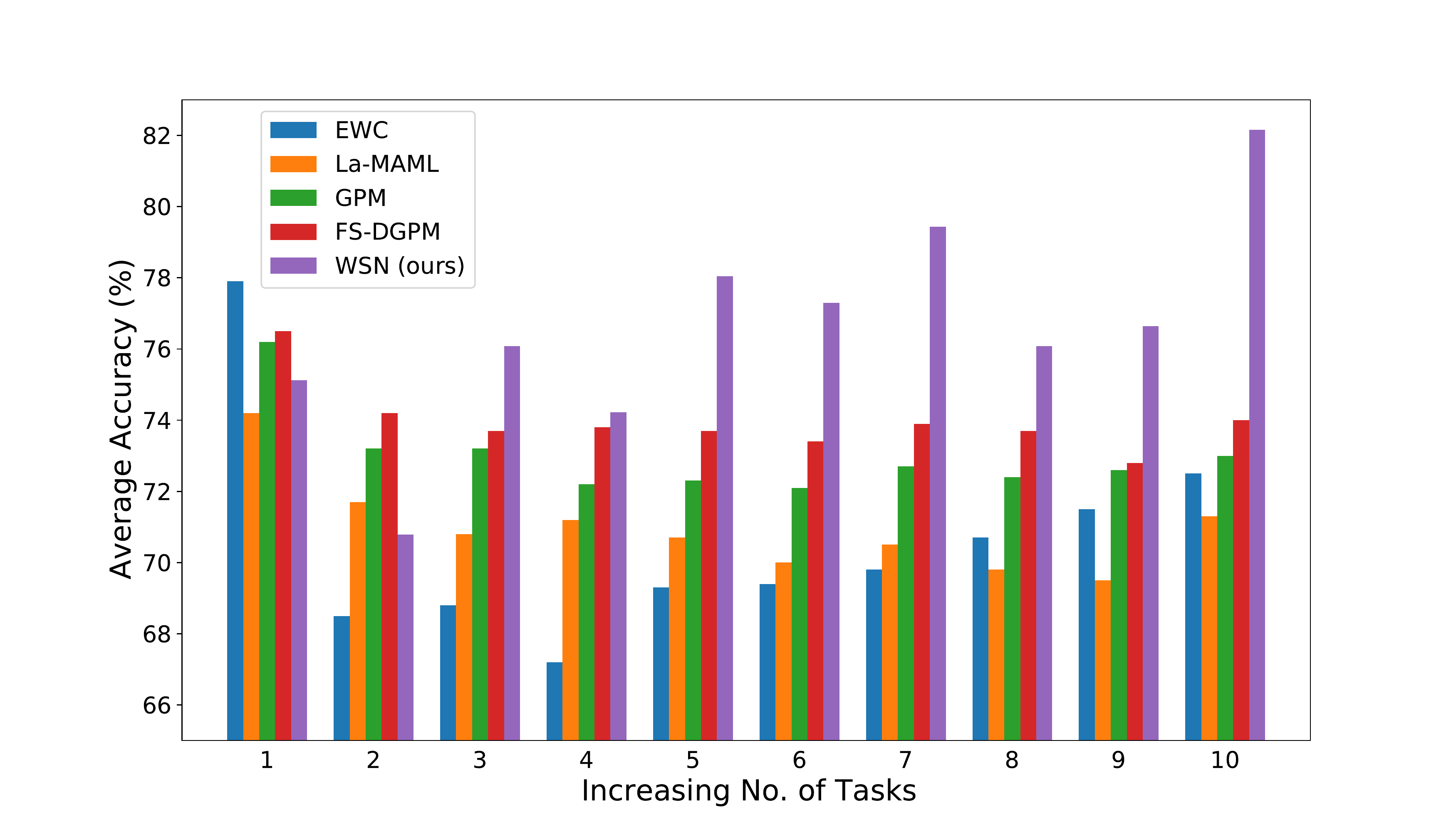}
    
    
    \end{adjustbox}
    \vspace{-0.25in}
    \caption{\textbf{Comparisons on CIFAR100-Split} with \cite{deng2021flattening}: It was difficult to determine the superiority of WSN only by comparing the performances in the early stage, however, it shows the progressively better performances of WSN than others.}
  \label{fig:bar_cifar100_10}
  \vspace{-0.1in}
\end{figure}

We use a multi-head setting to evaluate our WSN algorithm under the more challenging visual classification benchmarks. The WSN's performances are compared with others w.r.t three measurements on three major benchmark datasets as shown in \Cref{tab:main_sota_table}. Our WSN outperformed all state-of-the-art, achieving the best average accuracy of 76.38\%, 61.79\%, and 71.96\%. WSN is also a forget-free model (BWT = ZERO) with the least model capacity in these experiments. Note that we assume the model capacities are compared based on the model size without extra memory, such as samples. We highlight that our method achieves the highest accuracy, the lowest capacity, and backward transfer on all datasets. \Cref{fig:main_tinyimg_plots} shows the process of performance and compressed capacity changing with the number of tasks on the TinyImageNet datasets, where the “Average Progressive Capacity” metric is defined as the average capacity (the proportion of the number of network weights used for any one of the tasks) after five runs of the experiment with different seed values. Furthermore, we consistently showed progressively improved performances of WSN than others on CIFAR-100 Split datasets as shown in \Cref{fig:bar_cifar100_10}. The increasing number of reused weights (see \Cref{fig:main_confs_wsn_packnet}) could explain the progressively improved performances as shown in \Cref{fig:main_conf_hard_soft_wsn_cifar100}. Other forward-transferring results are depicted in Appendix.

\subsection{Forget-Free Performance and Model Capacity}
We prepare results on performance and bit-map compressed capacity on the TinyImageNet dataset as shown in \Cref{fig:main_tinyimg_plots} - “c=0.1” and “c=0.1+7bit-Huffman” refers respectively to using 10\% of network weights and no compression on the binary mask and the latter refers to the same with 7bit-Huffman encoding. In \Cref{fig:main_tinyimg_plots} (a), using initial capacity, $c=0.1$ shows better performances over others. With fixed $c=0.1$, the bit-wise Huffman compression rate delivers positive as the number of tasks increases, as shown in \Cref{fig:main_tinyimg_plots} (b). The most interesting part is that the average compression rate increases as the number of bits to compress increases, and the increasing ratio of reused weights (symbols with a high probability in the Huffman encoding) might affect the compression rate (symbols with a small probability might be rare in the Huffman tree, where the infrequent symbols tend to have long bit codes). We investigated how the compression rate is related to the total model capacity. The more bits the binary mask compression does, the less the model capacity to save is required. This shows that within 40 tasks, N-bit Huffman compressed capacities are sub-linearly increasing as binary map capacities increase linearly. The 7-bit Huffman encoding is enough to compress binary maps without exceeding the model capacity, even though the compression rate $\alpha$ depends on compression methods typically.

\subsection{Catastrophic Forgetting From WSN's Viewpoint} 
We interpreted how reused weights affect the inference performances on the TinyImageNet dataset as shown in \Cref{fig:main_tinyimg_all_acc_plots}. We divide all used weights for solving sequential tasks into specific sets for more precise interpretability. All used weights (a) within a trained dense network are separated as follows:
\textbf{\textit{all used}} represents all activated sets of weights up to task $t - 1$.
\textbf{\textit{per task}} represents an activated set of weights at task $t$.
\textbf{\textit{new per task}} represents a new activated set of weights at task $t$.
\textbf{\textit{reused per task}} represents an intersection set of weights per task and all used weights.
\textbf{\textit{reused for all tasks}} represents an intersection set of weights reused from task $1$ up to task $t-1$.

First, our WSN adaptively reuses weights to solve the sequential tasks. In  \Cref{fig:main_tinyimg_all_acc_plots} (b), initial and progressive task capacity start from $c$ value; the proportion of reused weights per task decreases in the early task learning stage. However, it tends to be progressively saturated to $c=0.1$ since the number of the new activated set of weights decreases, and the proportion of reused weights for all prior tasks tends to decrease. In other words, WSN uses diverse weights to solve the sequential tasks within all used weights rather than depending on the reused weights for all prior tasks as the number of tasks increases.  

Second, our WSN provides a stepping stone for forget-free continual learning. Regarding the benefits of WSN, in \Cref{fig:main_tinyimg_all_acc_plots} (c), we interpret the importance of three types of weights through an ablation study. The additional evaluations were performed by the acquired task binary masks and trained models to investigate the importance of reused weights in each layer, where the "w/" refers to "with reused network weights" and the "w/o" refers to "without reused network weights." Model forgetting occurred from the performances without using weights reused per task severely. The most significant weights were weights reused per task, the subset of all used weights; the importance of the weights reused for all prior tasks decreases as the number of tasks increases since its capacity gets small relatively, as shown in \Cref{fig:main_tinyimg_all_acc_plots} (b). Moreover, in \Cref{fig:main_tinyimg_all_acc_plots} (d), we inspected layer-wise forgetting caused by removing weights reused per task of network layers; the performance sensitivities were quite diverse. In particular, we observed that the most performance drops at the Conv1 layer.

\begin{figure}[ht]
    \centering
    \vspace{-0.1in}
    \setlength{\tabcolsep}{0pt}{%
    \begin{tabular}{cc}
    \includegraphics[width=0.5\columnwidth]{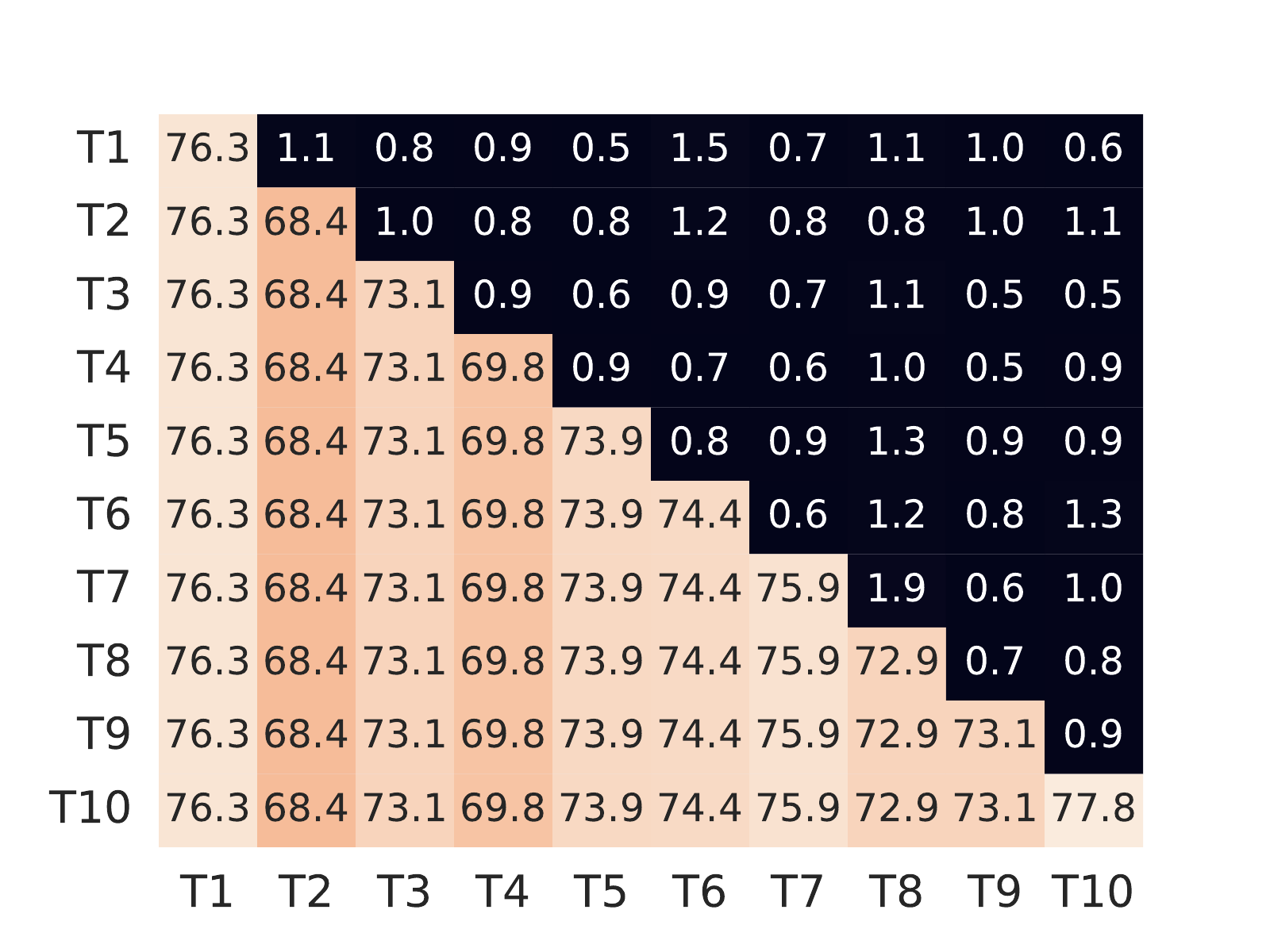} & 
    \includegraphics[width=0.5\columnwidth]{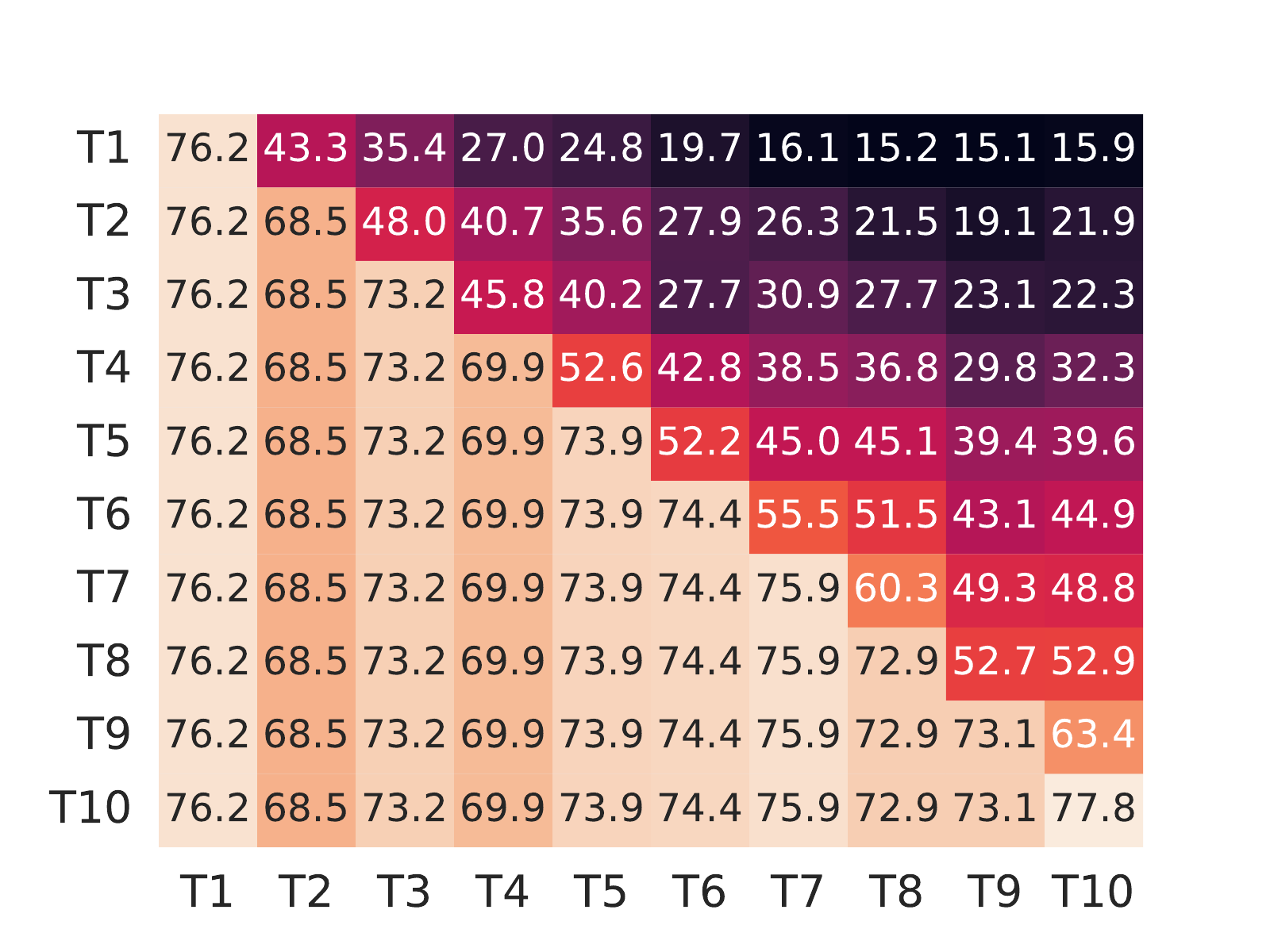} \\
    \small (a) WSN, $c = 5\%$ & \small (b) \textcolor{magenta}{SoftNet}, $c = 5\%$ \\
    \includegraphics[width=0.5\columnwidth]{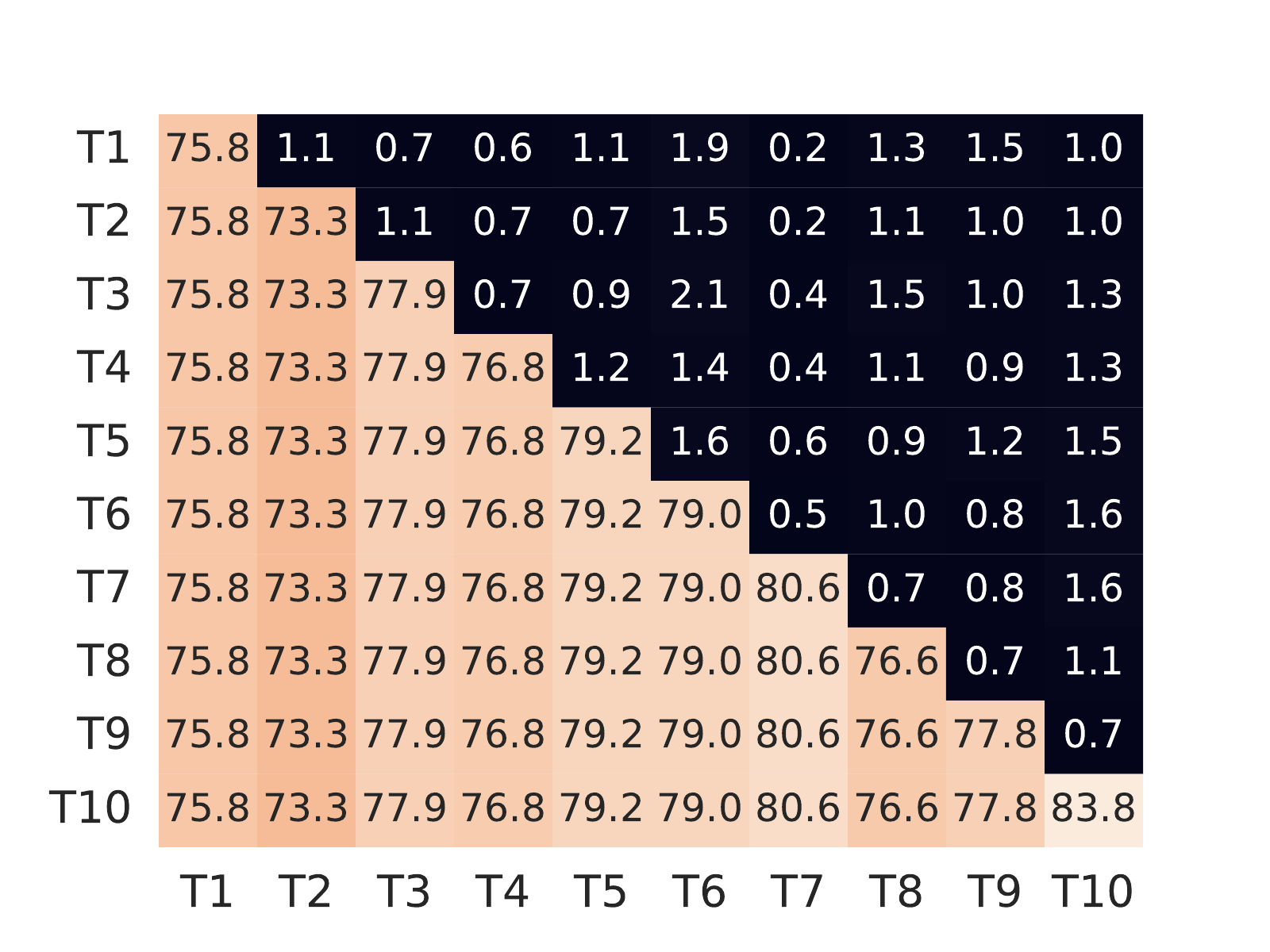} & 
    \includegraphics[width=0.5\columnwidth]{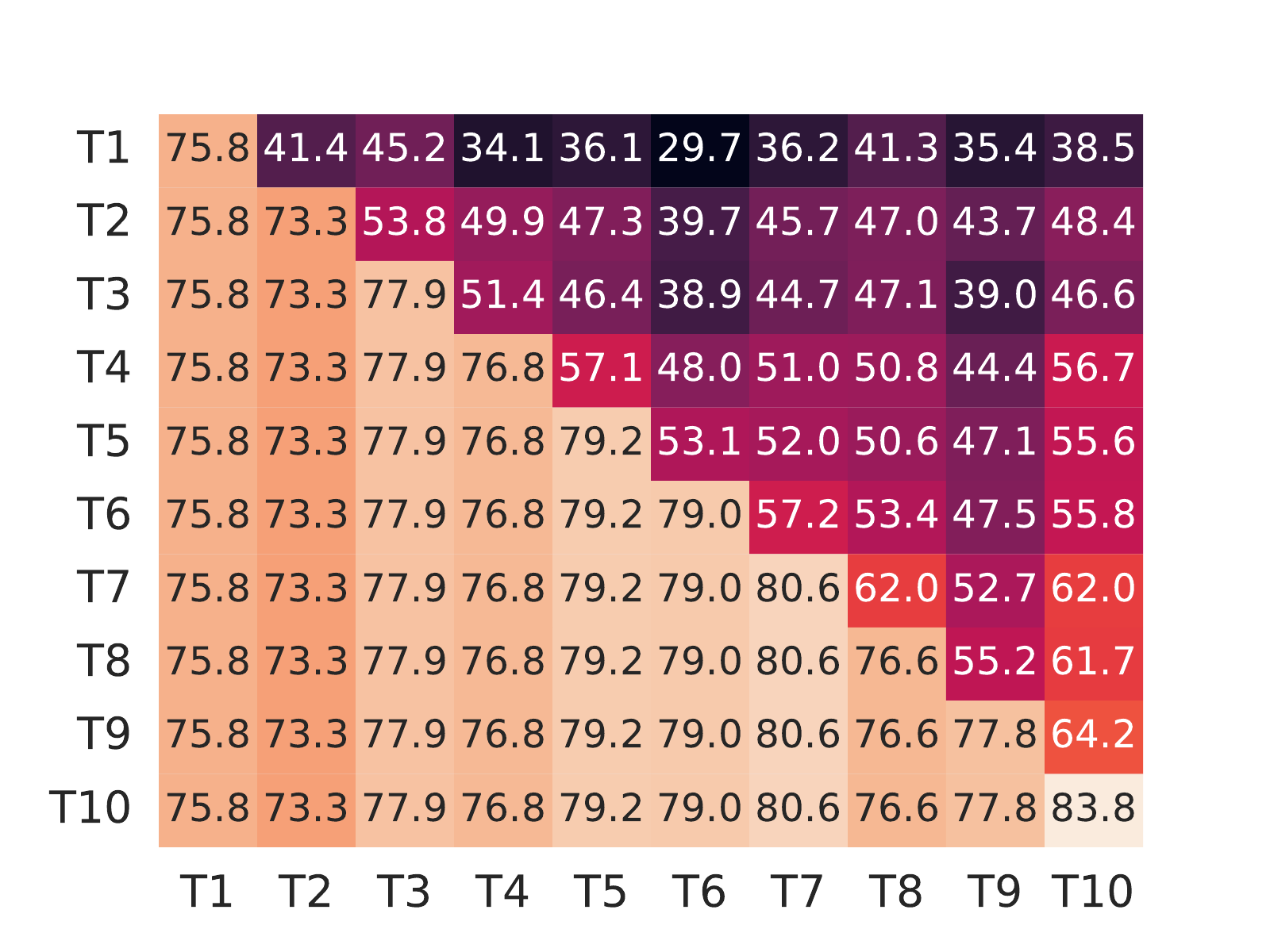} \\
    \small (c) WSN, $c = 70\%$  & \small (d) \textcolor{magenta}{SoftNet}, $c = 70\%$
    \end{tabular}
    }
    \caption{\textbf{Average Forward Transfer Matrix} on CIFAR-100 Split. WSN / \textcolor{magenta}{SoftNet} with $c = 5\%$ and $c = 70\%$, respectively.}
    \label{fig:main_conf_hard_soft_wsn_cifar100}
    \vspace{-0.12in}
\end{figure}

Finally, our WSN reuses weights if the knowledge from previous tasks is already enough to solve the task at hand and employs a few new weights otherwise. Specifically, from task 7, all used weights seem enough to infer all tasks since the weights reused per task catch up with the task performances. For more generalized forget-free continual learning, the model should consider the layer-wise sensitivity of weights reused per task when selecting weights reused for all prior tasks. These analyses might broadly impact other machine learning fields, such as transfer, semi-supervised, and domain adaptation.

\begin{table*}[t]
\begin{center}
\caption{Computational efficiency of WSN compared with PackNet and SupSup. We report model training time in hours.}

\resizebox{0.99\textwidth}{!}{
\begin{tabular}{lccccccccc}
\toprule


\multicolumn{1}{c}{\textbf{Method}}&\multicolumn{3}{c}{\textbf{Permuted MNIST}}&
\multicolumn{3}{c}{\textbf{5 Dataset}}&
\multicolumn{3}{c}{\textbf{Omniglot Rotation}} \\

\midrule
&ACC (\%) & CAP (\%) & Tr. TIME (h) & ACC (\%) & CAP (\%) & Tr. TIME (h) & ACC (\%) & CAP (\%) & Tr. TIME (h) \\
\midrule

PackNet & 96.37~\scriptsize($\pm$ 0.04) & 96.38 \scriptsize($\pm$ 0.00) & 0.49 \scriptsize($\pm$ 0.03) & 92.81~\scriptsize($\pm$ 0.12) & 82.86~\scriptsize($\pm$ 0.00) & 3.38 \scriptsize($\pm$ 0.11) & 30.70~\scriptsize($\pm$ 1.50) & 399.2~\scriptsize($\pm$ 0.00) & 7.30 \scriptsize($\pm$ 0.01)  \\

SupSup & 96.31~\scriptsize($\pm$ 0.09) & 122.89 \scriptsize($\pm$ 0.07) & 0.48 \scriptsize($\pm$ 0.06) & 93.28 \scriptsize($\pm$ 0.21) & 104.27~\scriptsize($\pm$ 0.21) & 3.20 \scriptsize($\pm$ 0.01) & 58.14~\scriptsize($\pm$ 2.42) & 407.12~\scriptsize($\pm$ 0.17) & 6.92 \scriptsize($\pm$ 0.03)  \\

\textbf{WSN (best)} & \textbf{96.41}~\scriptsize($\pm$ \textbf{0.07}) & \textbf{77.73~\scriptsize($\pm$ 0.36)} & \textbf{0.35 \scriptsize($\pm$ 0.02)} & \textbf{93.41}~\scriptsize($\pm$ \textbf{0.13}) & \textbf{86.10}~\scriptsize($\pm$ \textbf{0.57}) & \textbf{3.02 \scriptsize($\pm$ 0.03)} & \textbf{87.28}~\scriptsize($\pm$ \textbf{0.72}) & \textbf{79.85}~\scriptsize($\pm$ \textbf{1.19}) & \textbf{6.33 \scriptsize($\pm$ 0.04)} \\

\bottomrule
\toprule
\multicolumn{1}{c}{\textbf{Method}}&\multicolumn{3}{c}{\textbf{CIFAR-100 Split}}&
\multicolumn{3}{c}{\textbf{CIFAR-100 Superclass}}&
\multicolumn{3}{c}{\textbf{TinyImageNet}} \\

\midrule
&ACC (\%) & CAP (\%) & Tr. TIME (h) & ACC (\%) & CAP (\%) & Tr. TIME (h) & ACC (\%) & CAP (\%) & Tr. TIME (h) \\
\midrule

PackNet & 72.39~\scriptsize($\pm$ 0.37) & 96.38 ~\scriptsize($\pm$ 0.00) & 1.04 \scriptsize($\pm$ 0.19) & 58.78~\scriptsize($\pm$ 0.52) & 126.65 \scriptsize($\pm$ 0.00) & 0.46 \scriptsize($\pm$ 0.01) & 55.46~\scriptsize($\pm$ 1.22) & 188.67 \scriptsize($\pm$ 0.00) & 1.39 \scriptsize($\pm$ 0.03)  \\

SupSup & 75.47~\scriptsize($\pm$ 0.30) & 129.00 \scriptsize($\pm$ 0.03)& 0.79 \scriptsize($\pm$ 0.14) & 61.70~\scriptsize($\pm$ 0.31) & 162.49~\scriptsize($\pm$ 0.00) & 0.37 \scriptsize($\pm$ 0.00) & 59.60~\scriptsize($\pm$ 1.05) & 214.52~\scriptsize($\pm$ 0.89) & 0.92 \scriptsize($\pm$ 0.00)  \\

\textbf{WSN (best)} & \textbf{76.38}~\scriptsize($\pm$ \textbf{0.34}) & \textbf{99.13}~\scriptsize($\pm$ \textbf{0.48}) & \textbf{0.71 \scriptsize($\pm$ 0.09)} & \textbf{61.79}~\scriptsize($\pm$ \textbf{0.23}) & \textbf{80.93}~\scriptsize($\pm$ \textbf{1.58}) & \textbf{0.36 \scriptsize($\pm$ 0.00)} & \textbf{71.96}~\scriptsize($\pm$ \textbf{1.41}) & \textbf{48.65}~\scriptsize($\pm$ \textbf{3.03}) & \textbf{0.89 \scriptsize($\pm$ 0.00)} \\
\bottomrule
\end{tabular}}
\label{tab:comp_efficiency}
\end{center}
\vspace{-0.12in}
\end{table*}

\begin{figure}[ht]
    \centering
    \vspace{-0.1in}
    
    \setlength{\tabcolsep}{0pt}{%
    
    \begin{tabular}{cc}
    \includegraphics[width=0.5\columnwidth]{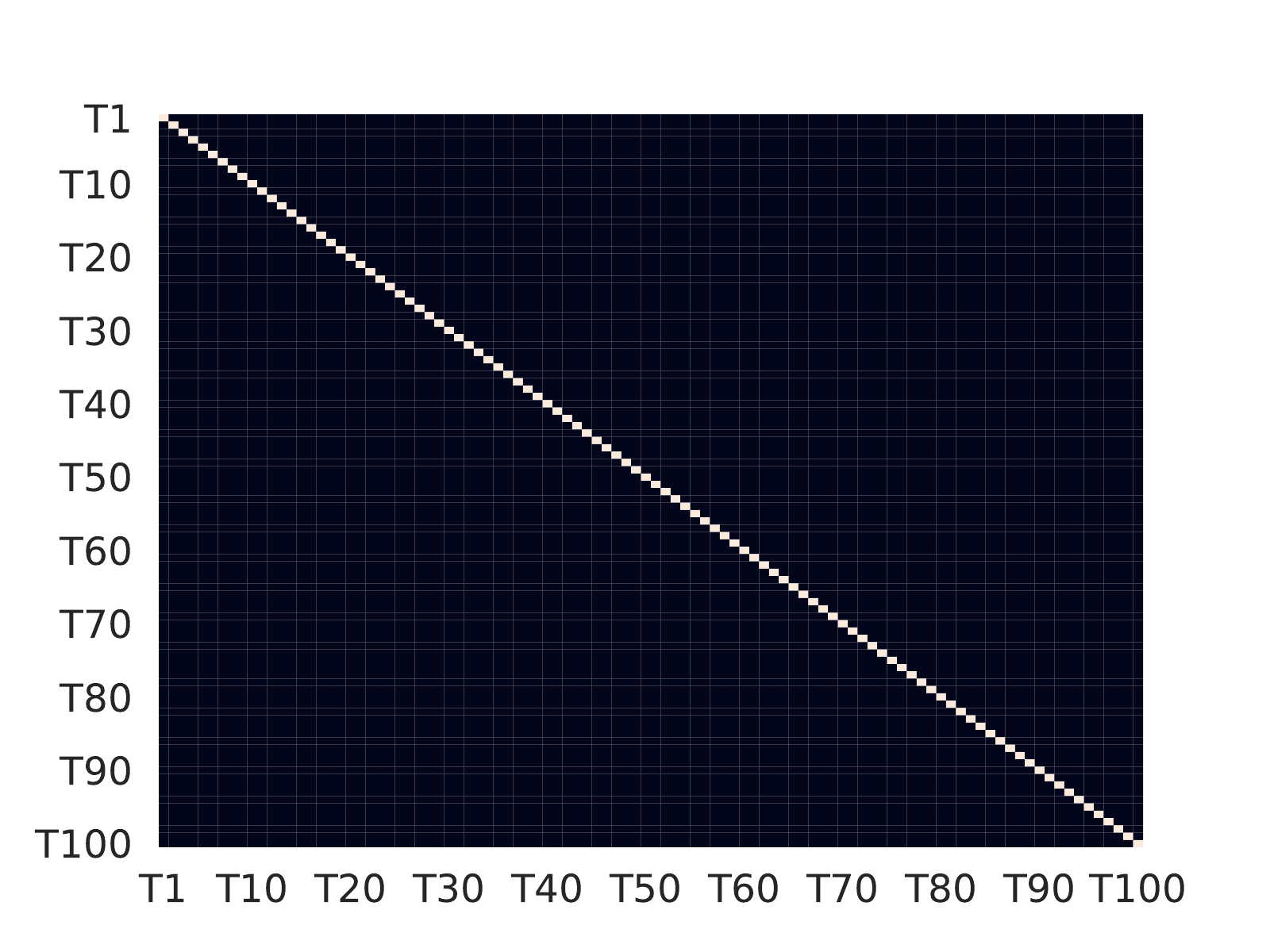} & 
    \includegraphics[width=0.5\columnwidth]{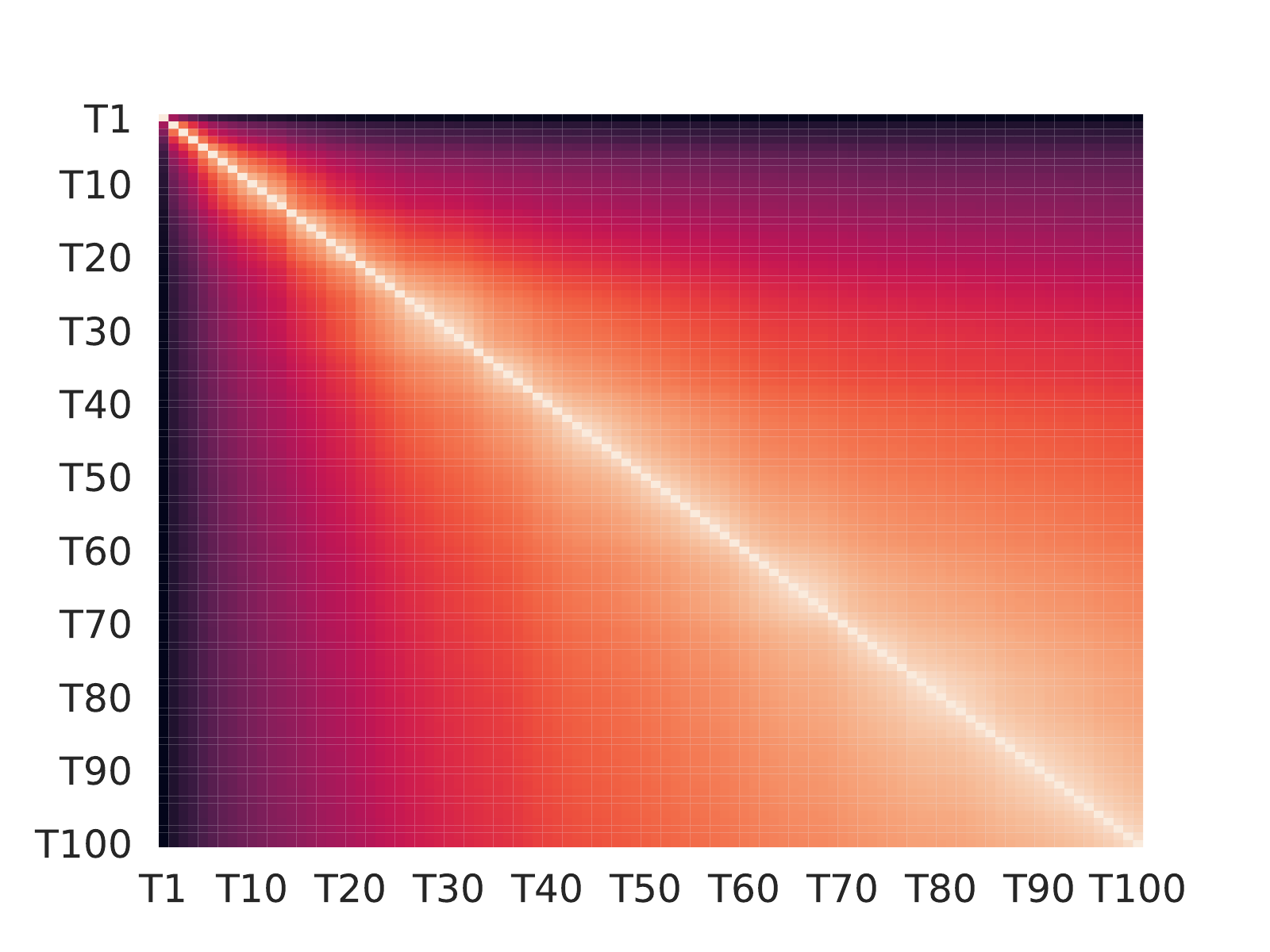} \\
    \small (a) PackNet & \small (b) WSN, $c = 5\%$ \\
    \includegraphics[width=0.5\columnwidth]{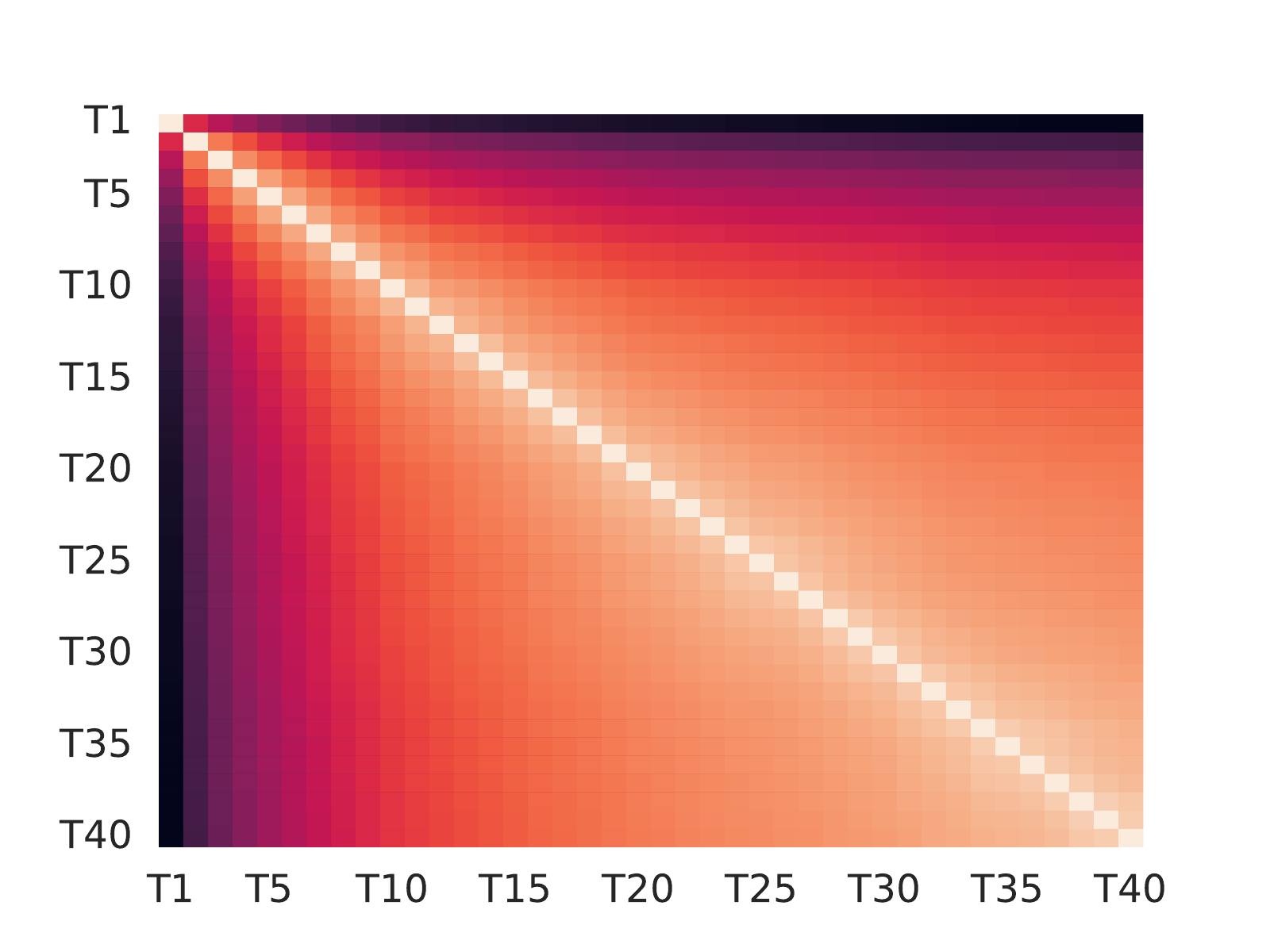} & 
    \includegraphics[width=0.5\columnwidth]{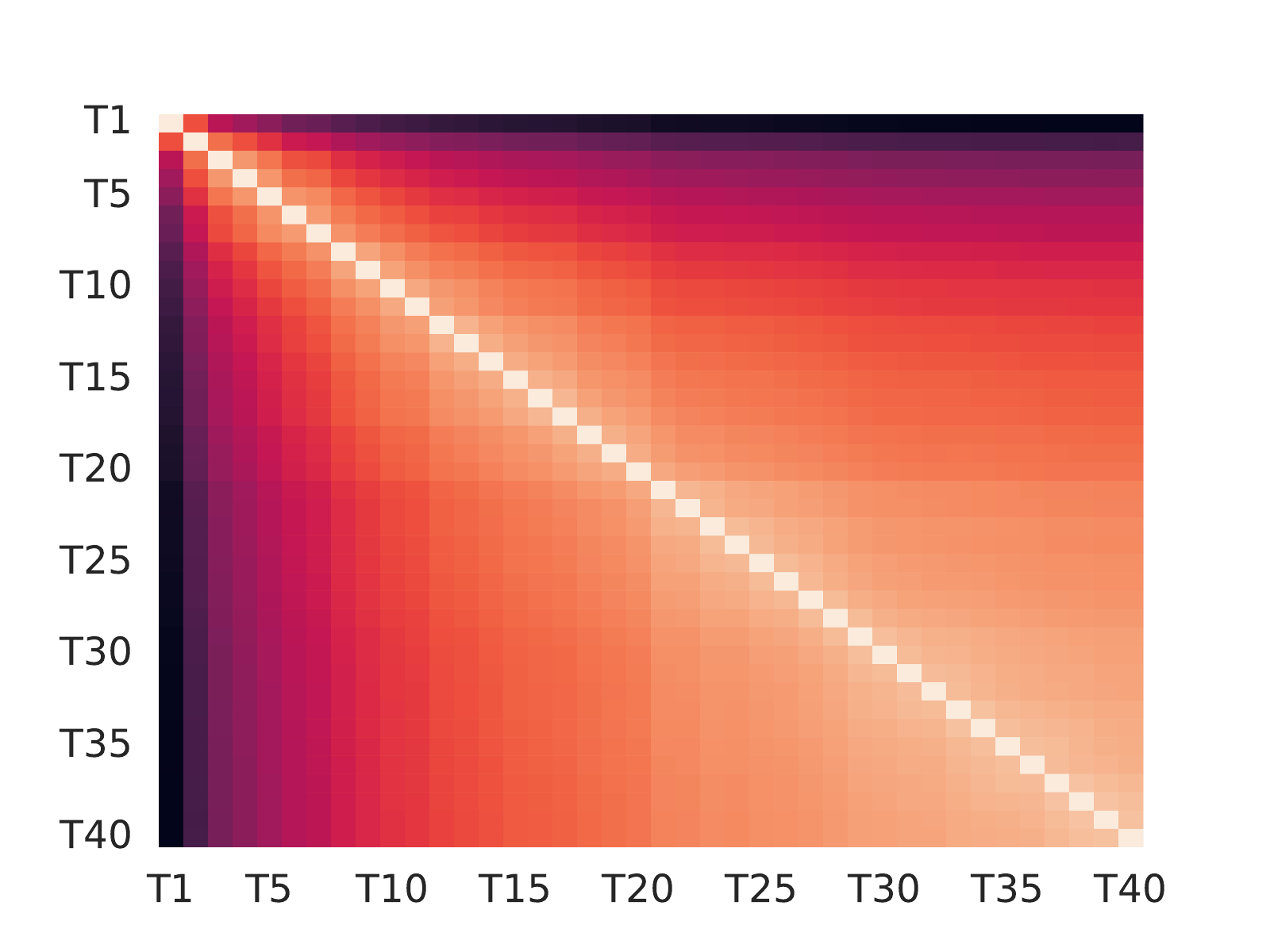} \\
    \small (c) WSN, $c = 10\%$  & \small (d) WSN, $c = 50\%$
    \end{tabular}
    }
    
    \caption{\textbf{Task-wise Binary Map Correlations} on Omniglot Rotation (Top row) and TinyImageNet (Bottom row). WSN reuses the weights, while PackNet does not. WSN with $c = 50\%$ reuses task weights more than the case of $c = 10\%$ as shown in higher correlation results than others.}
    \label{fig:main_confs_wsn_packnet}
    \vspace{-0.12in}
\end{figure}

\subsection{Sparse Binary Maps}
We prepared task-wise binary mask correlations to investigate how WSN reuses weights over sequential tasks. As shown in \Cref{fig:main_confs_wsn_packnet} (a) and (b), WSN tends to progressively transfer weights used for prior tasks to weights for new ones compared with PackNet. \Cref{fig:main_confs_wsn_packnet} (c) and (d) showed that the tendency of reused weights differs according to the $c$. This result might suggest that more sparse reused binary maps lead to generalization than others. 

\section{Results On Few-shot CIL}

\begin{table*}[t]
\begin{center}
\caption{Classification accuracy of ResNet18 on CIFAR-100 for 5-way 5-shot incremental learning. \underbar{Underbar} denotes the comparable results with FSLL~\cite{mazumder2021few}. $\ast$ denotes the results reported from~\cite{shi2021overcoming}.}

\resizebox{0.85\textwidth}{!}{
\begin{tabular}{lcccccccccc}
\toprule
\multicolumn{1}{c}{\multirow{2}{*}{\textbf{Method}}}&\multicolumn{9}{c}{\textbf{sessions}}& \multicolumn{1}{c}{\multirow{2}{*}{\thead{\textbf{The gap} \\ \textbf{with cRT}}}} \\ 
\cline{2-10}
& 1 & 2 & 3 & 4 & 5 & 6 & 7 & 8 & 9 &  \\
\midrule
cRT~\cite{shi2021overcoming} & 65.18 & 63.89 & 60.20 & 57.23 & 53.71 & 50.39 & 48.77 & 47.29 & 45.28 & - \\
\midrule
iCaRL~\cite{rebuffi2017icarl}$^\ast$ & 66.52 & 57.26 & 54.27 & 50.62 & 47.33 & 44.99 & 43.14 & 41.16 & 39.49 & -5.79 \\
Rebalance~\cite{hou2019learning}$^\ast$ & 66.66 & 61.42 & 57.29 & 53.02 & 48.85 & 45.68 & 43.06 & 40.56 & 38.35 & -6.93 \\
FSLL~\cite{mazumder2021few}$^\ast$ & 65.18 & 56.24 & 54.55 & 51.61 & 49.11 & 47.27 & 45.35 & 43.95 & 42.22 & -3.08 \\
iCaRL~\cite{rebuffi2017icarl} & 64.10 & 53.28 & 41.69 & 34.13 & 27.93 & 25.06 & 20.41 & 15.48 & 13.73 & -31.55 \\
Rebalance~\cite{hou2019learning} & 64.10 & 53.05 & 43.96 & 36.97 & 31.61 & 26.73 & 21.23 & 16.78 & 13.54 & -31.74 \\
TOPIC~\cite{cheraghian2021semantic} & 64.10 & 55.88 & 47.07 & 45.16 & 40.11 & 36.38 & 33.96 & 31.55 & 29.37 & -15.91 \\
F2M~\cite{shi2021overcoming} & 64.71 & 62.05 & 59.01 & 55.58 & 52.55 & 49.96 & 48.08 & 46.28 & 44.67 & -0.61 \\
\midrule 
FSLL~\cite{mazumder2021few} & 64.10 & 55.85 & 51.71 & 48.59 & 45.34 & 43.25 & 41.52 & 39.81 & 38.16 & -7.12 \\
HardNet (WSN), $c=50 \%$ & \underbar{64.80} & \underbar{60.77} & \underbar{56.95} & \underbar{53.53} & \underbar{50.40} & \underbar{47.82} & \underbar{45.93} & \underbar{43.95} & \underbar{41.91} & -\underbar{3.37} \\ 

HardNet (WSN), $c=80 \%$   
& 69.65	& 64.60 & 60.59 & 56.93 & 53.60 & 50.80 & 48.69 & 46.69 & 44.63	& -0.65 \\
HardNet (WSN), $c=99 \%$  
&71.95 & 66.83 & 62.75 & 59.09 & 55.92 & 53.03 & 50.78 & 48.52 & 46.31 & +1.03 \\ 
\midrule
\textcolor{cyan}{SoftNet}, ~~$c=50 \%$   
& 69.20 & 64.18 & 60.01 & 56.43 & 53.11 & 50.62 & 48.60 & 46.51 & 44.61 & -0.67 \\ 
\textcolor{cyan}{SoftNet}, ~~$c=80 \%$ 
& 70.38 & 65.04 & 60.94 & 57.26 & 54.13 & 51.58 & 49.52 & 47.36 & 45.16 & -0.12 \\

\textcolor{cyan}{SoftNet}, ~~$c=99 \%$ 
& \textbf{72.62} & \textbf{67.31} & \textbf{63.05} & \textbf{59.39}	& \textbf{56.00} & \textbf{53.23} & \textbf{51.06} & \textbf{48.83} & \textbf{46.63} & +\textbf{1.35}\\ 
\bottomrule
\end{tabular}
}
\label{tab:main_cifar100_5way_5shot}
\end{center}
\end{table*}

\begin{table*}[ht]
\begin{center}
\caption{Classification accuracy of ResNet18 on miniImageNet for 5-way 5-shot incremental learning. \underbar{Underbar} denotes the comparable results with FSLL~\cite{mazumder2021few}. $\ast$ denotes the results reported from \cite{shi2021overcoming}.}

\resizebox{0.85\textwidth}{!}{
\begin{tabular}{lccccccccccc}
\toprule
\multicolumn{1}{c}{\multirow{2}{*}{\textbf{Method}}}&\multicolumn{9}{c}{\textbf{sessions}}& \multicolumn{1}{c}{\multirow{2}{*}{\thead{\textbf{The gap} \\ \textbf{with cRT}}}} \\ 
\cline{2-10}
& 1 & 2 & 3 & 4 & 5 & 6 & 7 & 8 & 9 &  \\
\midrule
cRT \cite{shi2021overcoming} & 67.30 & 64.15 & 60.59 & 57.32 & 54.22 & 51.43 & 48.92 & 46.78 & 44.85 & - \\ 
\midrule
iCaRL \cite{rebuffi2017icarl}$^\ast$ & 67.35 & 59.91 & 55.64 & 52.60 & 49.43 & 46.73 & 44.13 & 42.17 & 40.29 & -4.56 \\ 
Rebalance \cite{hou2019learning}$^\ast$ & 67.91 & 63.11 & 58.75 & 54.83 & 50.68 & 47.11 & 43.88 & 41.19 & 38.72 & -6.13 \\ 
FSLL \cite{mazumder2021few}$^\ast$ & 67.30 & 59.81 & 57.26 & 54.57 & 52.05 & 49.42 & 46.95 & 44.94 & 42.87 & -1.11 \\ 
iCaRL \cite{rebuffi2017icarl} & 61.31 & 46.32 & 42.94 & 37.63 & 30.49 & 24.00 & 20.89 & 18.80 & 17.21 & -27.64 \\ 
Rebalance \cite{hou2019learning} & 61.31 & 47.80 & 39.31 & 31.91 & 25.68 & 21.35 & 18.67 & 17.24 & 14.17 & -30.68 \\ 
TOPIC \cite{cheraghian2021semantic} & 61.31 & 50.09 & 45.17 & 41.16 & 37.48 & 35.52 & 32.19 & 29.46 & 24.42 & -20.43 \\ 
IDLVQ-C \cite{chen2020incremental} & 64.77 & 59.87 & 55.93 & 52.62 & 49.88 & 47.55 & 44.83 & 43.14 & 41.84 & -3.01 \\ 
F2M \cite{shi2021overcoming} & 67.28 & 63.80 & 60.38 & 57.06 & 54.08 & 51.39 & 48.82 & 46.58 & 44.65 & -0.20 \\ 
\midrule 
FSLL \cite{mazumder2021few} & 66.48 & 61.75 & 58.16 & 54.16 & 51.10 & 48.53 & 46.54 & 44.20 & 42.28 & -2.57 \\ 
HardNet (WSN), $c=50\%$  
& \underbar{65.13} & \underbar{60.37} & \underbar{56.12} & \underbar{53.17} & \underbar{50.17} & \underbar{47.74} & \underbar{45.34} & \underbar{43.35} & \underbar{42.13} & -\underbar{2.72} \\ 
HardNet (WSN), $c=80\%$  
& 69.73 & 64.46 & 60.42 & 57.09 & 54.09 & 51.18 & 48.76 & 46.81	& 45.66 & +0.81 \\ 
HardNet (WSN), $c=90\%$  
& 64.68	& 59.80 & 55.70 & 52.82 & 50.01 & 47.30 & 45.17 & 43.34 & 42.09	& -2.76 \\
\midrule 
\textcolor{cyan}{SoftNet}, ~~$c=50\%$  
& 72.83	& 67.23 & 62.82 & 59.41 & 56.44 & 53.55 & 50.92 & 48.99 & 47.60 & +2.75 \\ 
\textcolor{cyan}{SoftNet}, ~~$c=80\%$  
& 76.63	& 70.13	& 65.92 & 62.52 & \textbf{59.49} & \textbf{56.56} & 53.71 & 51.72 & \textbf{50.48} & +\textbf{5.63} \\ 
\textcolor{cyan}{SoftNet}, ~~$c=90\%$  
& 77.00 & \textbf{70.38} & 65.94 & 62.45 & 59.32 & 56.25 & \textbf{53.76} & \textbf{51.75} & 50.39 & +5.54 \\ 
\textcolor{cyan}{SoftNet}, ~~$c=97\%$  
& \textbf{77.17} & 70.32 & \textbf{66.15} & \textbf{62.55} & 59.48 & 56.46 & 53.71 & 51.68 & 50.24 & +5.39 \\ 
\bottomrule
\end{tabular}
}
\label{tab:main_miniImageNet_5way_5shot}
\end{center}
\end{table*}

\subsection{Results and Comparisons}
We compared SoftNet with the architecture-based FSLL and HardNet (WSN) methods. We pick FSLL as an architecture-based baseline since it selects important parameters for acquiring old/new class knowledge. The architecture-based results on CIFAR-100 and miniImageNet are presented in \Cref{tab:main_cifar100_5way_5shot} and \Cref{tab:main_miniImageNet_5way_5shot} respectively. The performances of HardNet show the effectiveness of the subnetworks that go with less model capacity compared to dense networks. To emphasize our point, we found that ResNet18, with approximately $50\%$ parameters, achieves comparable performances with FSLL on CIFAR-100 and miniImageNet. In addition, the performances of ResNet20 with $30\%$ parameters (HardNet) are comparable with those of FSLL on CIFAR-100. 

\begin{figure*}[ht]
    \centering
    \begin{tabular}{cc}
    \includegraphics[height=4.8cm, trim={0.1cm 0.02cm 0.1cm 0.2cm},clip]{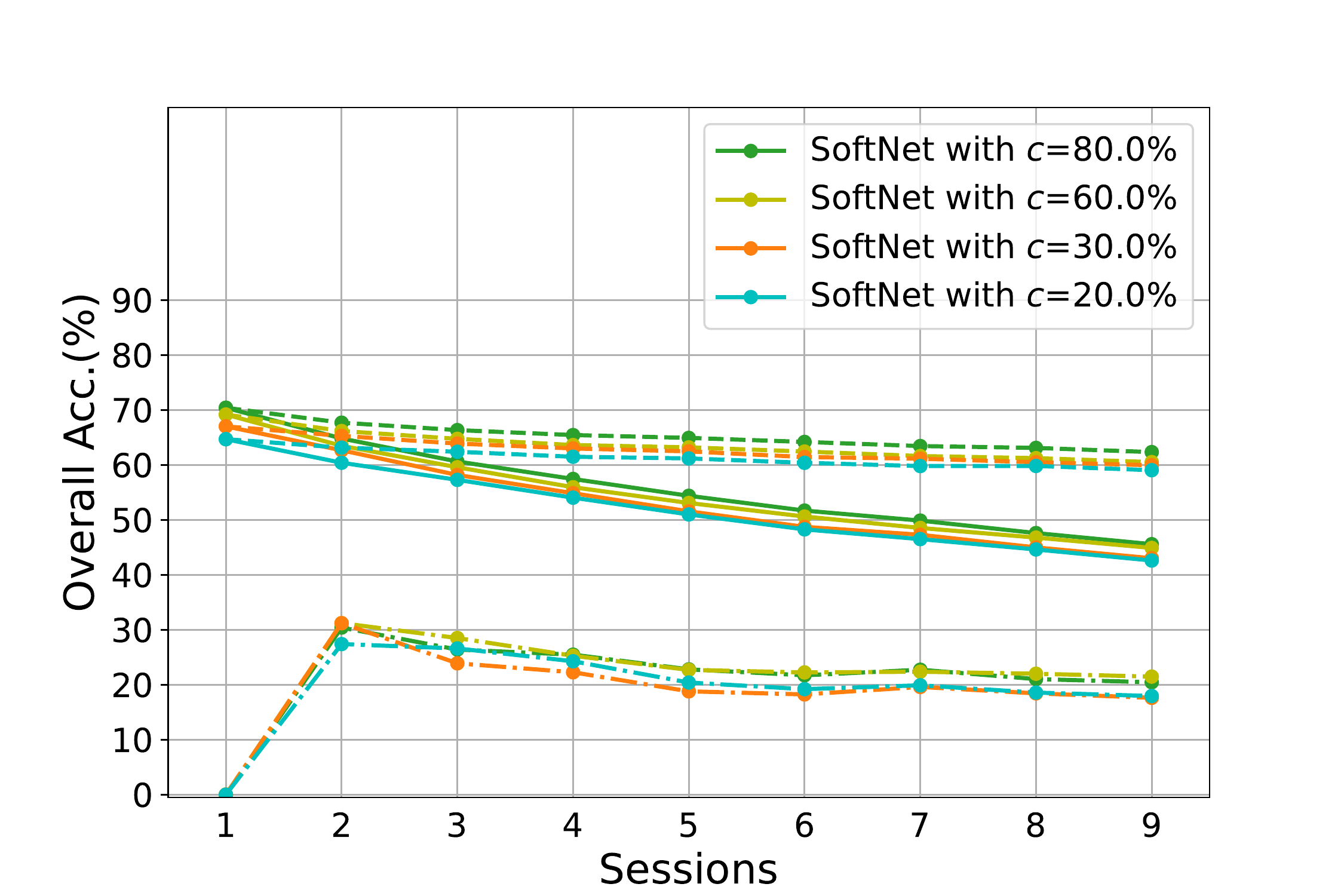} &
    \includegraphics[height=4.8cm, trim={0.1cm 0.02cm 0.1cm 0.2cm},clip]{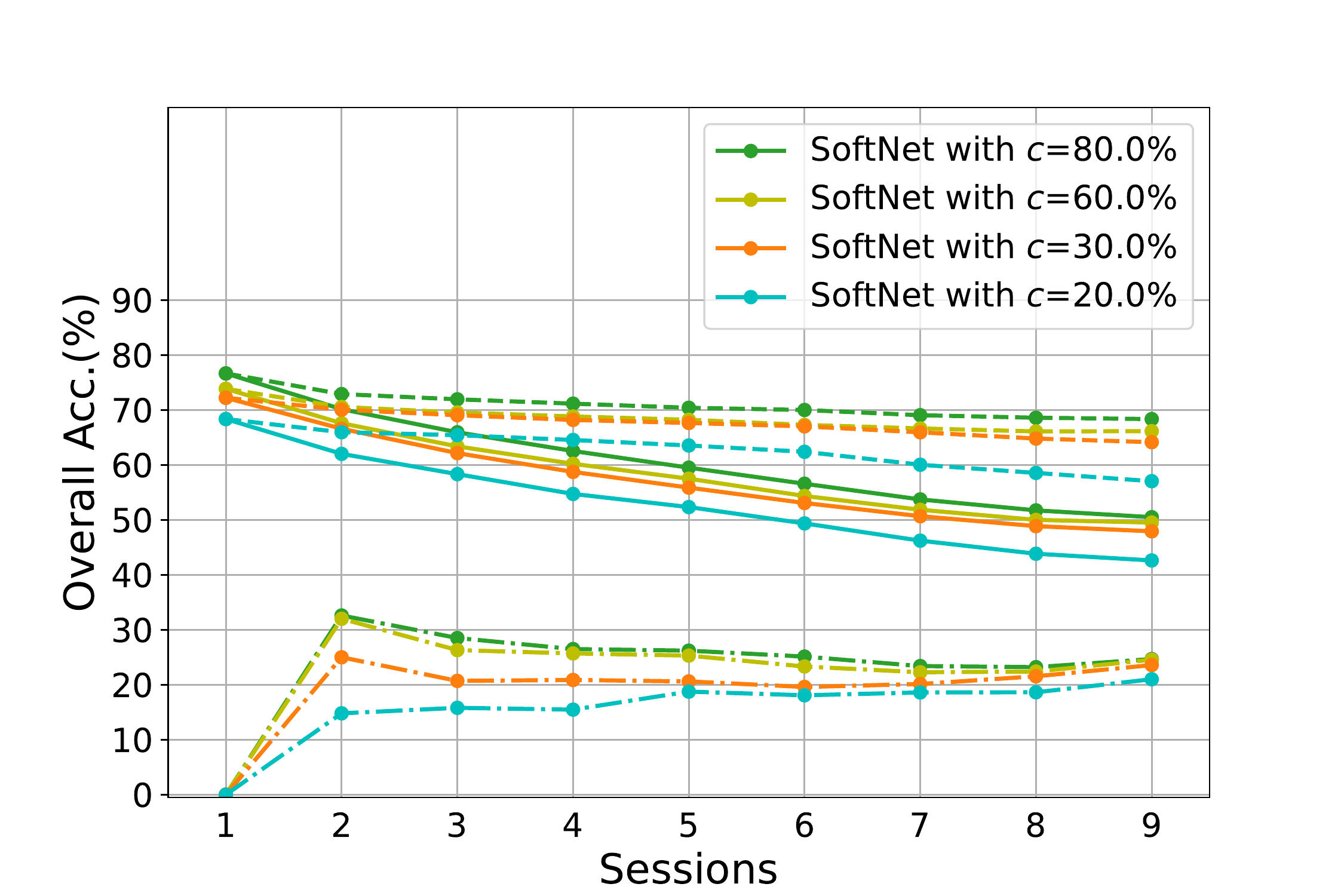} \\
    
    \makecell{\small{(a) CIFAR-100}} &
    \makecell{\small{(b) miniImageNet}} \\
    
    \end{tabular}
    \caption{\small \textbf{Classification accuracy of \textcolor{cyan}{SoftNet} on CIFAR-100 and miniImageNet for 5-way 5-shot FSCIL:} the overall performance depends on capacity $c$ and the softness of subnetwork. Note that solid(\sampleline{}), dashed(\sampleline{dashed}), and dashed-dot(\sampleline{dash pattern=on .7em off .2em on .05em off .2em}) lines denote overall, base, and novel class performances respectively.}
    \label{fig:softnet_cifar100_miniImageNet}
\end{figure*}

Experimental results are prepared to analyze the overall performances of SoftNet according to the sparsity and dataset as shown in \Cref{fig:softnet_cifar100_miniImageNet}. As we increase the number of parameters employed by SoftNet, we achieve performance gain on both benchmark datasets. The performance variance of SoftNet's sparsity seems to depend on datasets because the performance variance on CIFAR-100 is less than that on miniImageNet. In addition, SoftNet retains prior session knowledge successfully in both experiments as described in the dashed line, and the performances of SoftNet ($c=60.0 \%$) on the new class session (8, 9) of CIFAR-100 than those of SoftNet ($c=80.0 \%$) as depicted in the dashed-dot line. From these results, we could expect that the best performances depend on the number of parameters and properties of datasets. 

\begin{figure*}[ht]
    \centering
    \includegraphics[width=0.90\linewidth]{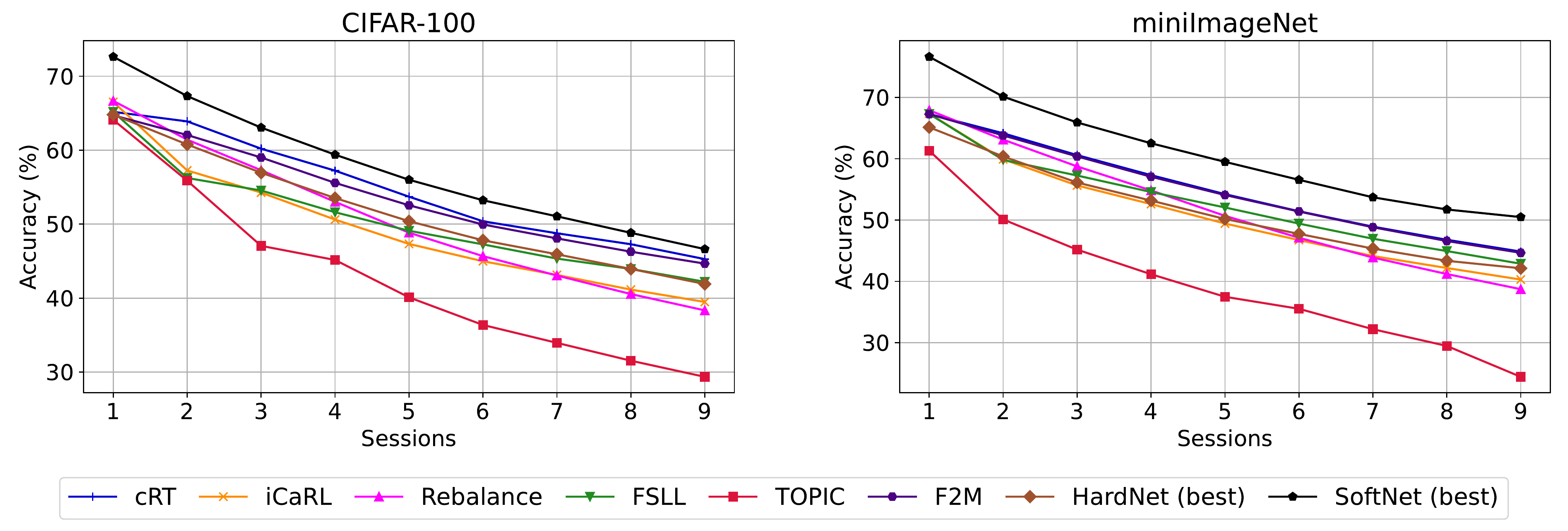}
    \caption{Comparision of subnetworks (HardNet (WSN) and \textcolor{cyan}{SoftNet}) with state-of-the-art methods.}
    \label{fig:overall_classification_acc}
\end{figure*}

Our SoftNet outperforms the state-of-the-art methods and cRT, which is used as the approximate upper bound of FSCIL~\cite{shi2021overcoming} as shown in \Cref{tab:main_cifar100_5way_5shot} and \Cref{tab:main_miniImageNet_5way_5shot}. Moreover, \Cref{fig:overall_classification_acc} represents the outstanding performances of SoftNet on CIFAR-100 and miniImageNet. SoftNet provides a new upper bound on each dataset, outperforming cRT, while HardNet (WSN) provides new baselines among pruning-based methods.

\subsection{Considerations From SoftNet}


Through extensive experiments, we deduce the following conclusions for incorporating our method in the few-shot class incremental learning regarding architectures. 

\begin{figure*}[ht]
    \centering
    \setlength{\tabcolsep}{-6pt}{%
    \begin{tabular}{ccc}
    
    \includegraphics[width=0.36\textwidth]{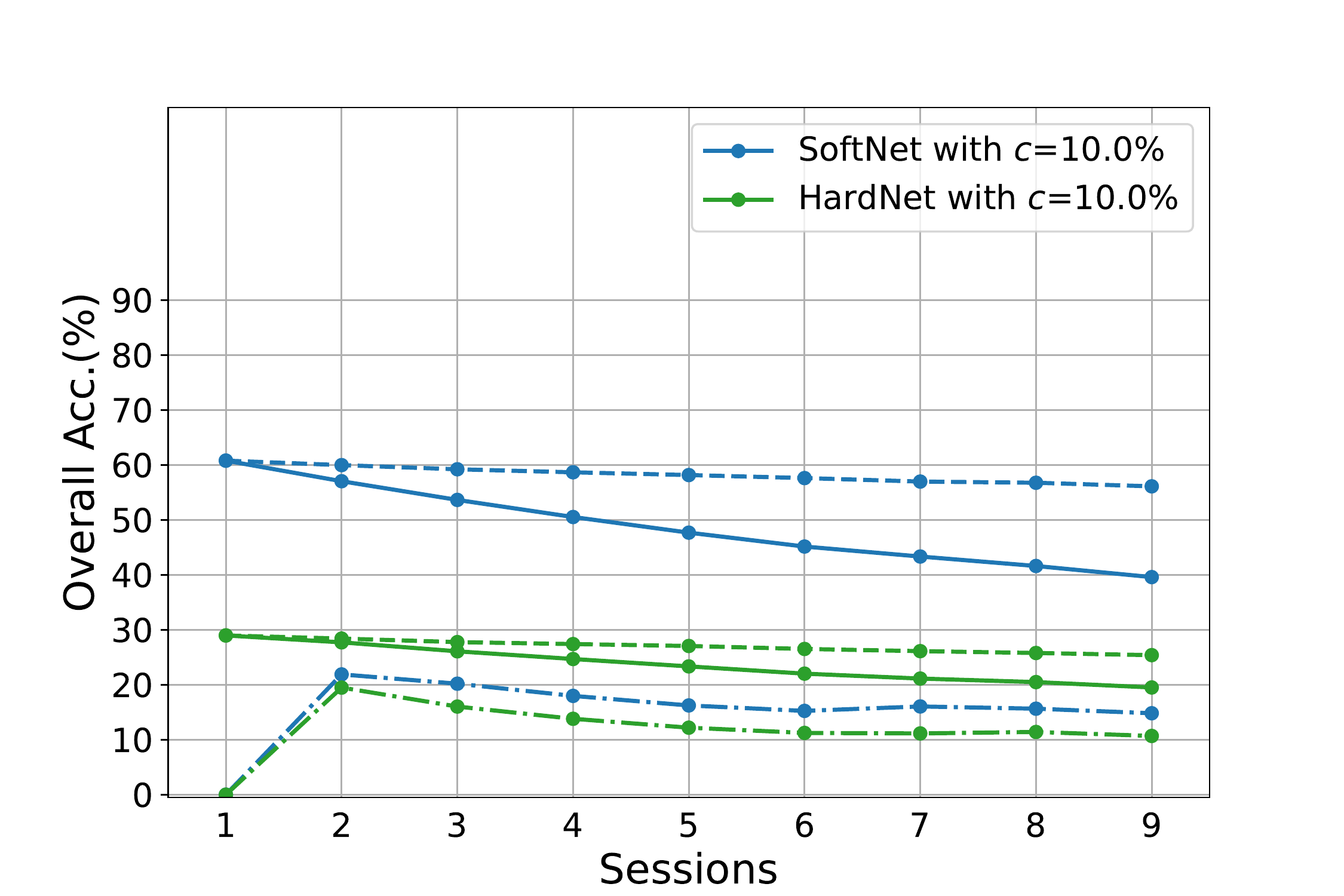} & 
    
    \includegraphics[width=0.36\textwidth]{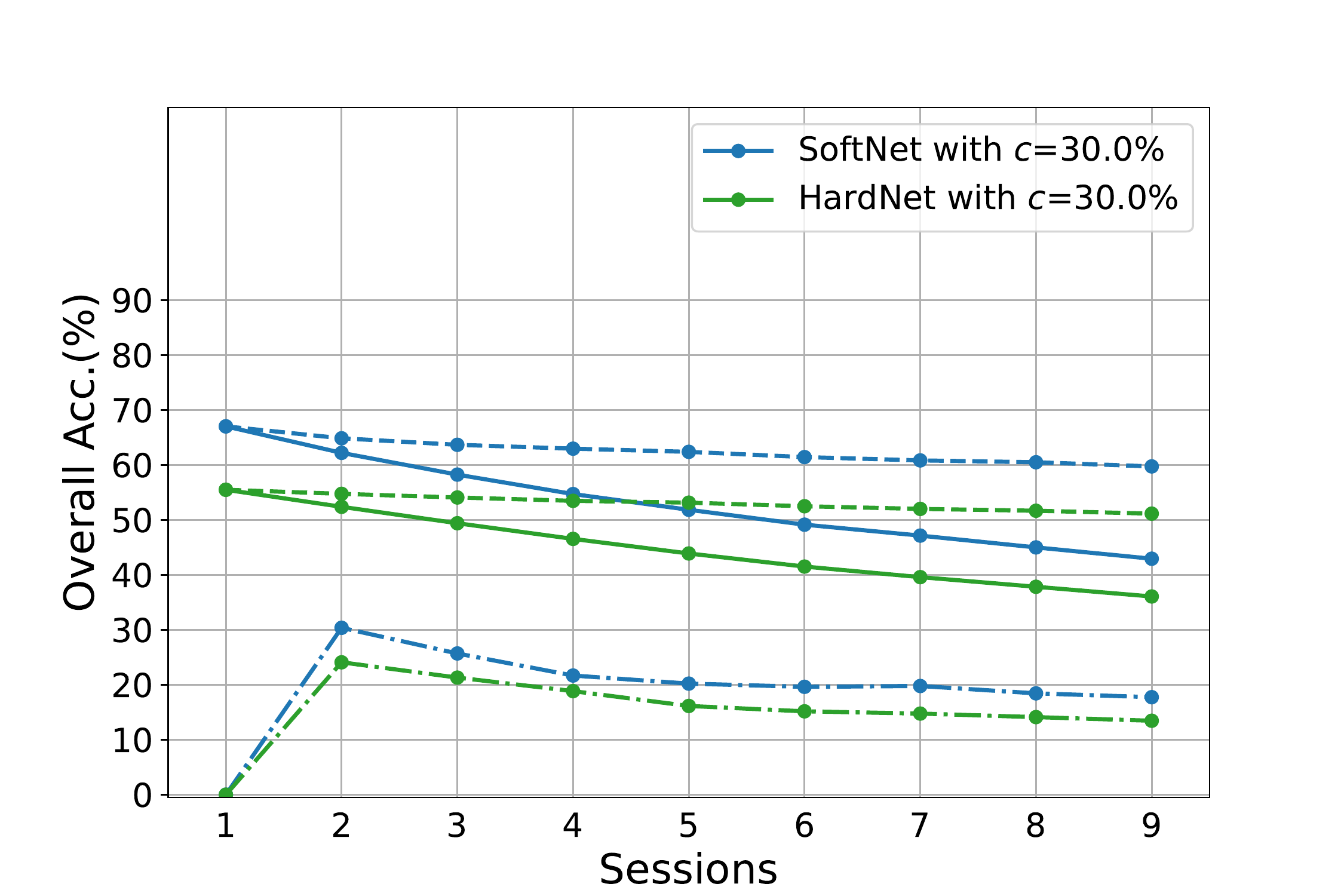} &
        
    \includegraphics[width=0.36\textwidth]{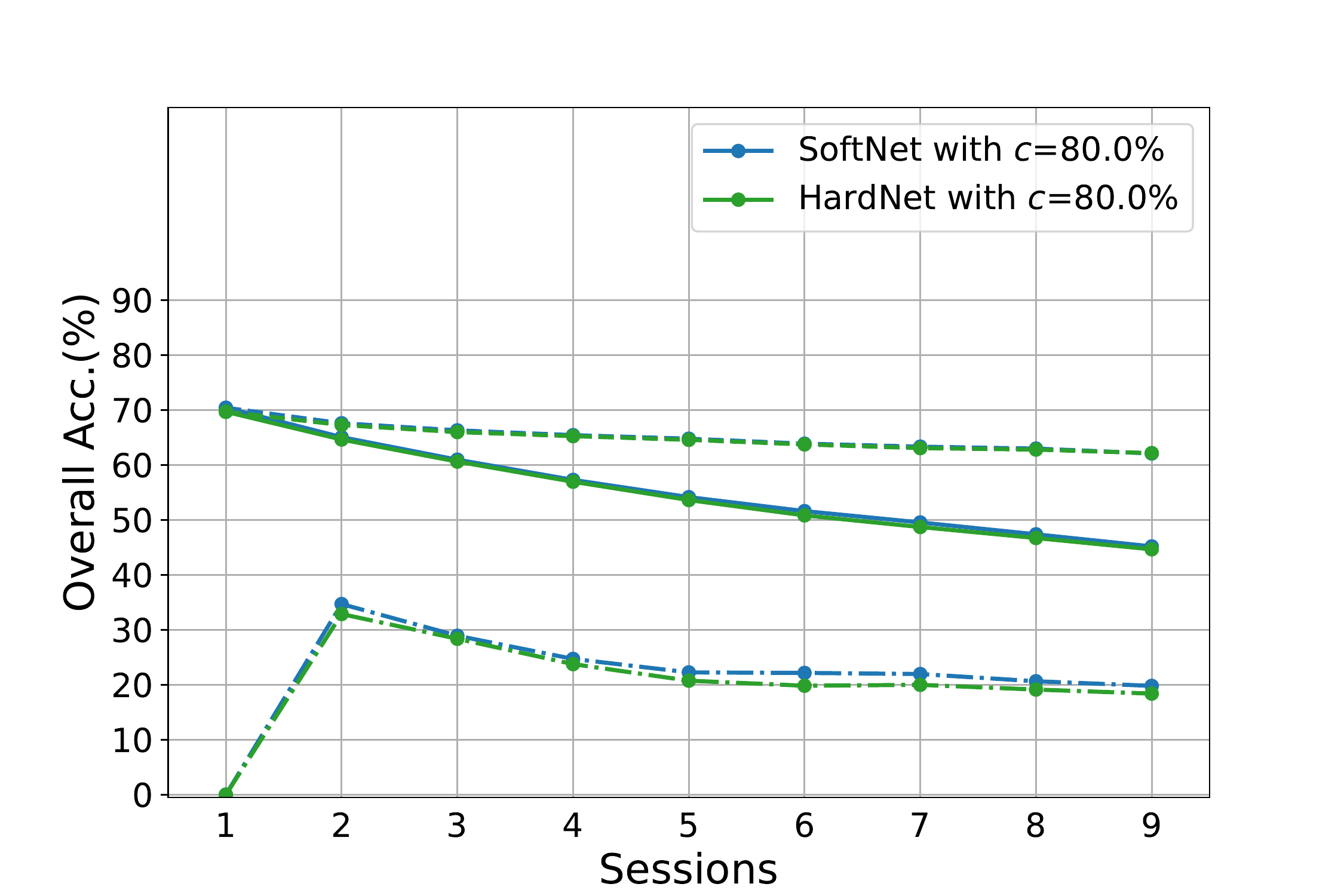} \\ 
    
    \small (a) $c=10\%$ & \small (b) $c=30\%$ & \small (c) $c=80\%$  \\
    \end{tabular}
    }
    \caption{\small \textbf{Performances of HardNet (WSN) v.s. \textcolor{cyan}{SoftNet} on CIFAR-100 for 5-way 5-shot FSCIL:} the overall performance depends on capacity $c$ and the softness of subnetwork. Note that solid(\sampleline{}), dashed(\sampleline{dashed}), and dashed-dot(\sampleline{dash pattern=on .7em off .2em on .05em off .2em}) lines denote overall, base, and novel class performances respectively.}
    \label{fig:main_plot_hardsoft_cifar100}
\end{figure*}
\begin{figure*}[!ht]
    \centering
    \setlength{\tabcolsep}{-6pt}{%
    \begin{tabular}{ccc}
    
    \includegraphics[width=0.36\textwidth]{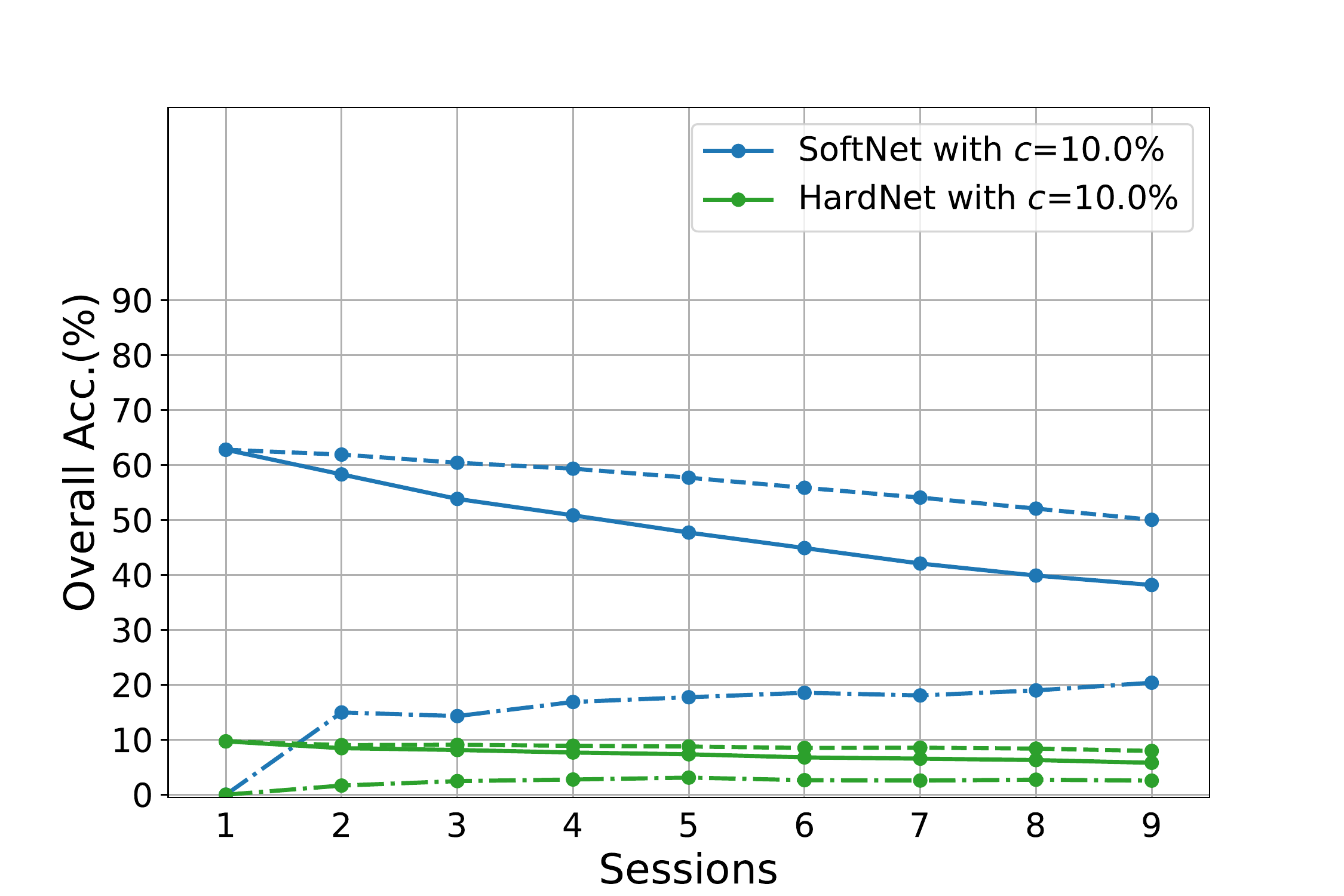} & 
    
    \includegraphics[width=0.36\textwidth]{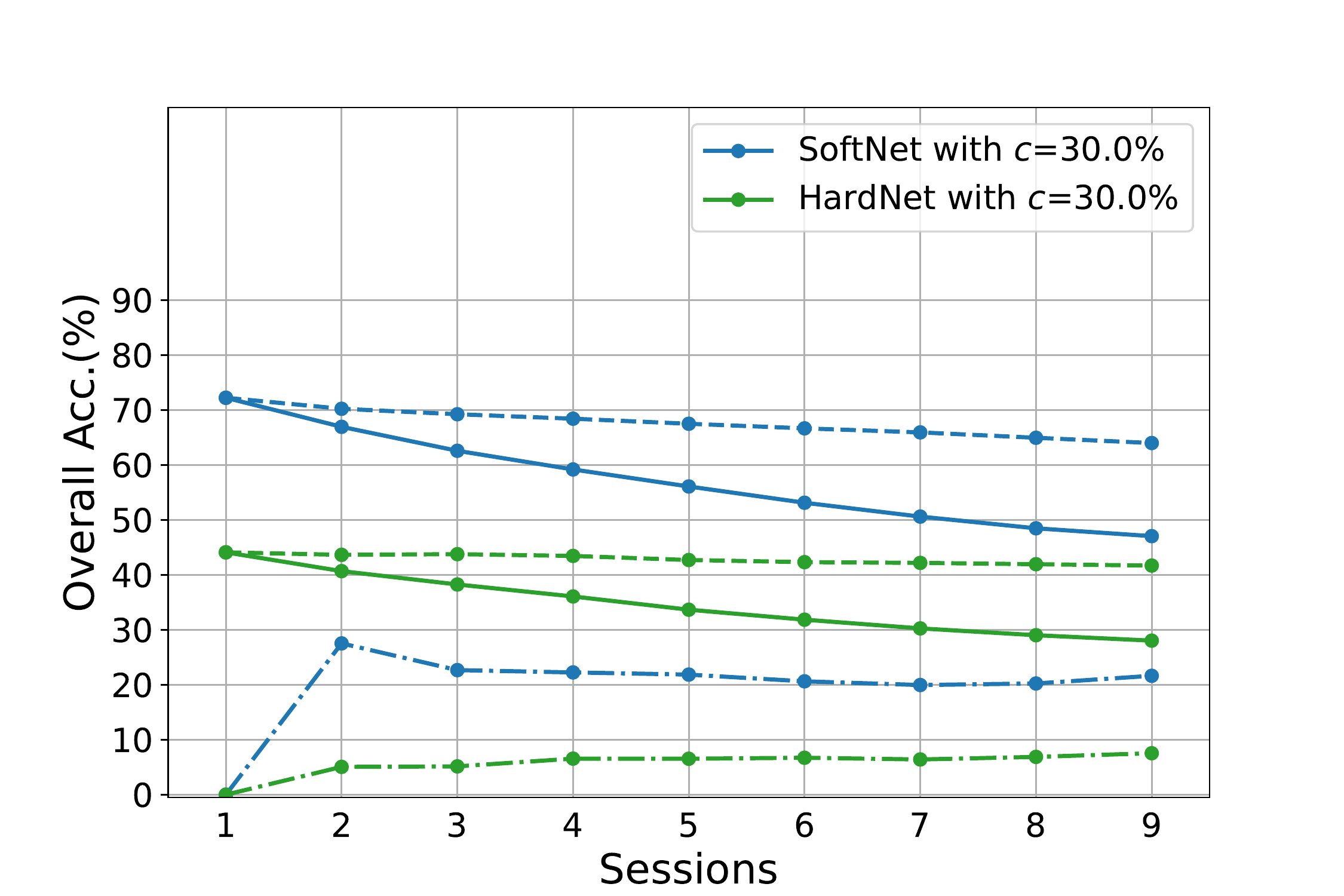} &
    
    \includegraphics[width=0.36\textwidth]{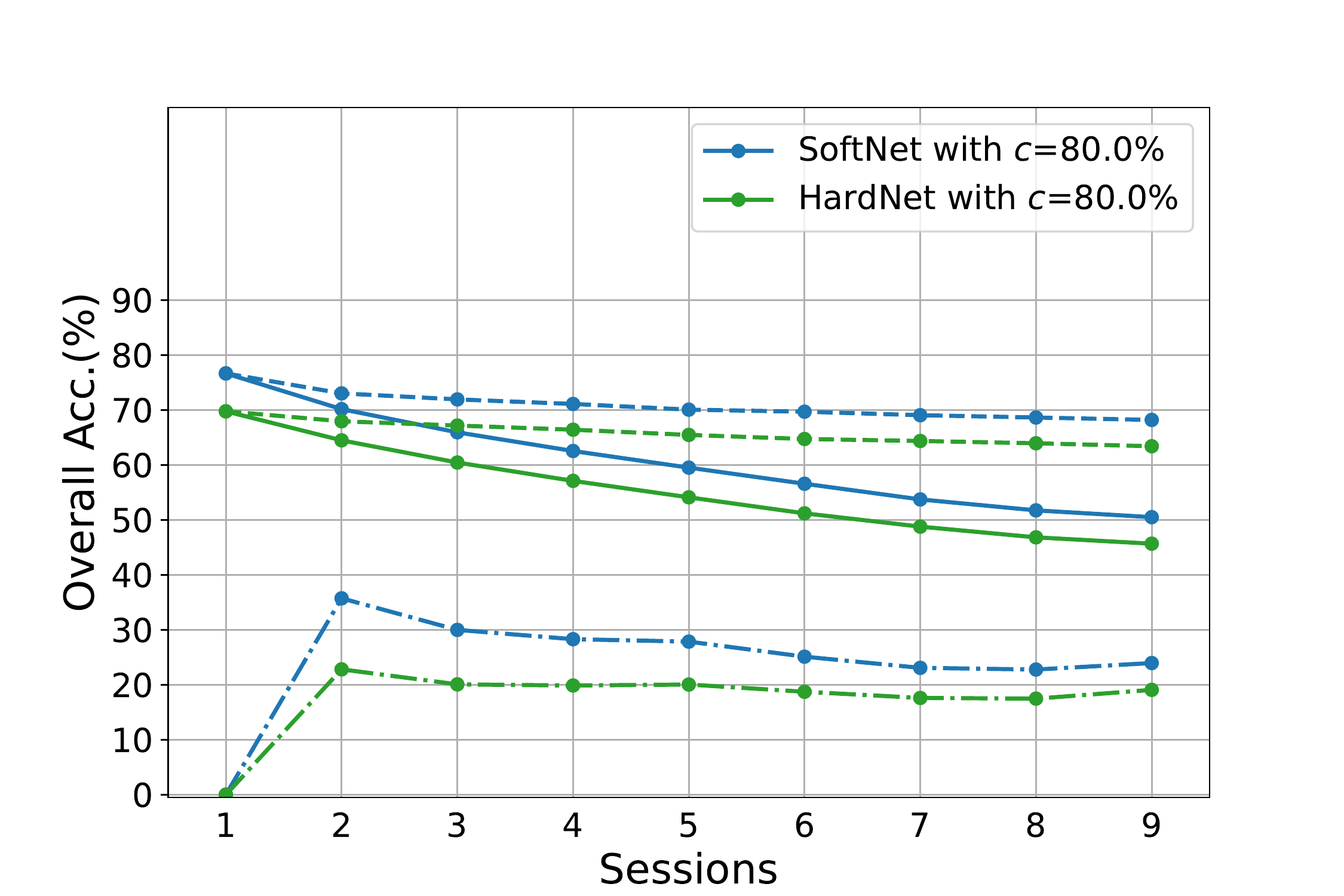} \\ 
    
    \small (a) $c=10\%$ & \small (b) $c=30\%$ & \small (c) $c=80\%$  \\
    \end{tabular}
    }
    \caption{\small \textbf{Performances of HardNet (WSN) v.s. \textcolor{cyan}{SoftNet} on miniImageNet for 5-way 5-shot FSCIL:} the overall performance depends on capacity $c$ and the softness of the subnetwork. Note that solid(\sampleline{}), dashed(\sampleline{dashed}), and dashed-dot(\sampleline{dash pattern=on .7em off .2em on .05em off .2em}) lines denote overall, base, and novel class performances respectively.}
    
    \label{fig:main_plot_hardsoft_miniImageNet}
\end{figure*}

\noindent 
\textbf{Comparisions of HardNet (WSN) and SoftNet}. Furthermore, increasing the number of network parameters leads to better overall performance in both subnetworks types, as shown in \Cref{fig:main_plot_hardsoft_cifar100} and \Cref{fig:main_plot_hardsoft_miniImageNet}. Subnetworks, in the form of HardNet and SoftNet, tend to retain prior (base) session knowledge denoted in dashed (\sampleline{dashed}) line, and HardNet seems to be able to classify new session class samples without continuous updates stated in dashed-dot (\sampleline{dash pattern=on .7em off .2em on .05em off .2em}) line. From this, we could expect how much previous knowledge HardNet learned at the base session to help learn new incoming tasks (Forward Transfer). The overall performances of SoftNet are better than HardNet since SoftNet improves both base/new session knowledge by updating minor subnetworks. Subnetworks have a broader spectrum of performances on miniImageNet (\Cref{fig:main_plot_hardsoft_miniImageNet}) than on CIFAR-100 (\Cref{fig:main_plot_hardsoft_cifar100}). This could be an observation caused by the dataset complexity - i.e., if the miniImagenet dataset is more complex or harder to learn for a subnetwork or a deep model as such subnetworks need more parameters to learn miniImageNet than the CIFAR-100 dataset. 


\begin{figure*}[!ht]
    \centering
    \setlength{\tabcolsep}{-6pt}{%
    \begin{tabular}{ccc}
    
    \includegraphics[width=0.5\textwidth]{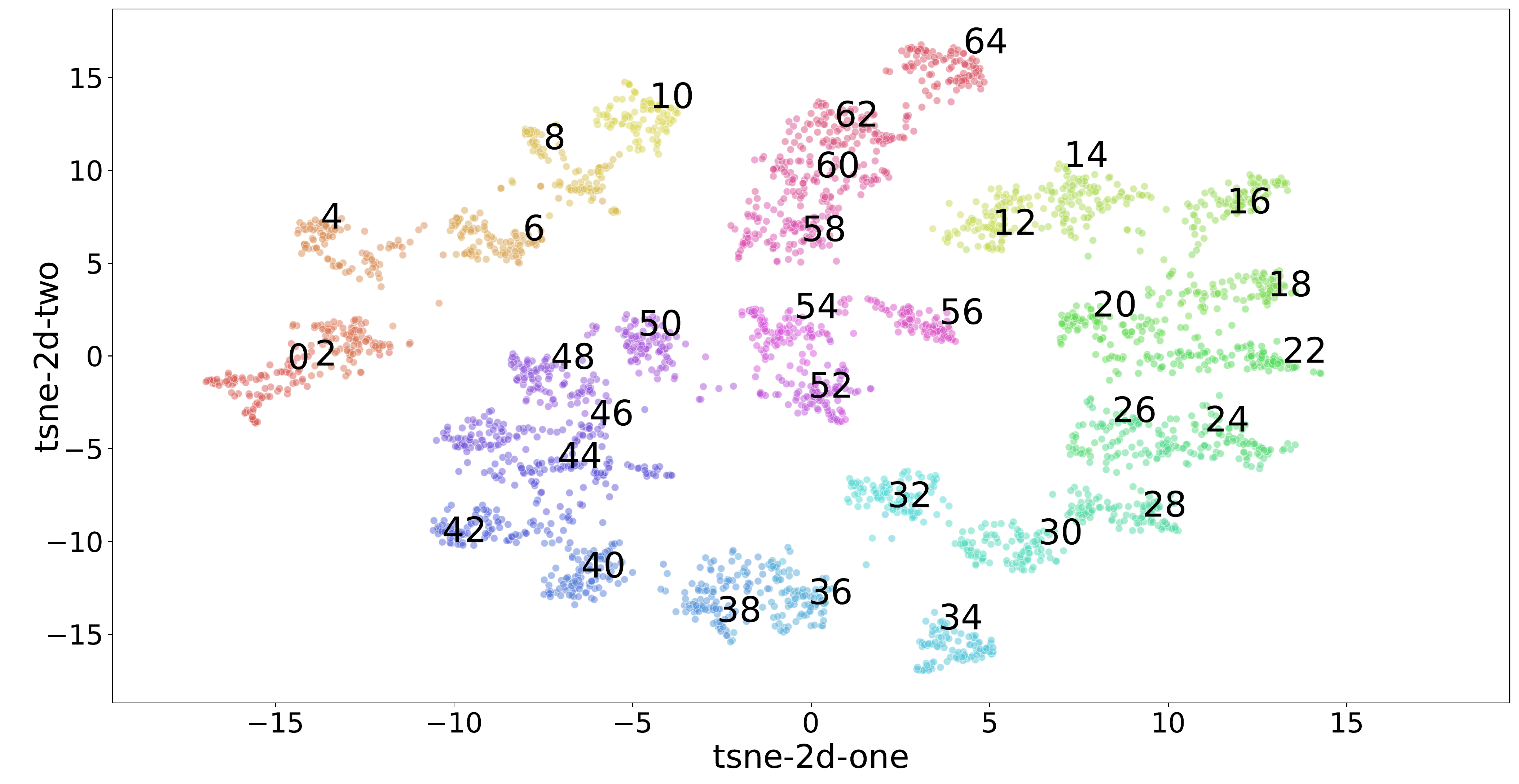} 
    &
    \;\;\;\;\;\;\;\;\;\;
    &
    \includegraphics[width=0.5\textwidth]{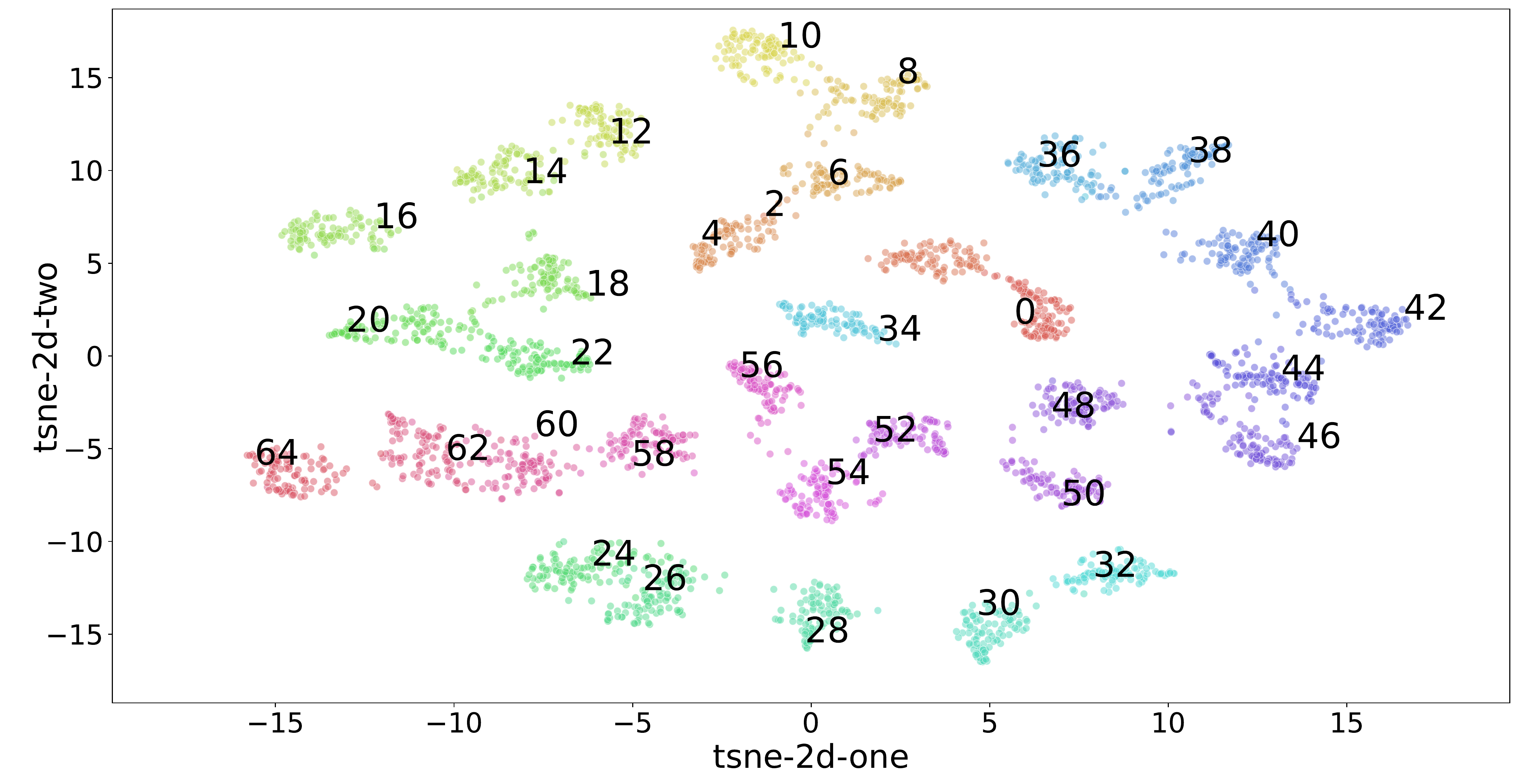} \\ 
    
    \small (a) HardNet (WSN) with $c=30\%$ @Session2 & & \small (b) \textcolor{cyan}{SoftNet} with $c=30\%$ @Session2  \\
    \end{tabular}
    }
    \caption{\small \textbf{t-SNE Plots of HardNet (WSN) v.s. \textcolor{cyan}{SoftNet} on miniImageNet for 5-way 5-shot FSCIL:}  t-SNE plots represent the embeddings of the even-numbered test class samples and compare one another. Note Session1 Class Set:$\{0,\cdots,59\}$ and Session2 Novel Class Set:$\{60,\cdots,64\}$.}
    \label{fig:main_plot_tsne_miniImageNet}
    \vspace{-0.12in}
\end{figure*}

\noindent 
\textbf{Smoothness of SoftNet.} SoftNet has a broader spectrum of performances than HardNet on miniImageNet. $20\%$ of minor subnet might provide a smoother representation than HardNet because the performance of SoftNet was the best approximately at $c=80\%$. We could expect that model parameter smoothness guarantees quite competitive performances from these results. To support the claim, we prepared the loss landscapes of a dense neural network, HardNet, and SoftNet on two Hessian eigenvectors~\cite{yao2020pyhessian} as shown in Fig. \ref{fig:main_plot_loss_lens}. We observed the following points through simple experiments. From these results, we can expect how much knowledge the specified subnetworks can retain and acquire on each dataset. The loss landscapes of Subnetworks (HardNet and SoftNet) were flatter than those of dense neural networks. The minor subnet of SoftNet helped find a flat global minimum despite random scaling weights in the training process. 

Moreover, we compared the embeddings using t-SNE plots as shown in \Cref{fig:main_plot_tsne_miniImageNet}. In t-SNE's 2D embedding spaces, the overall discriminative of SoftNet is better than that of HardNet in terms of base class set and novel class set. This $70\%$ of minor subnet affects SoftNet positively in base session training and offers good initialized weights in novel session training.

\noindent 
\textbf{Preciseness.} Regarding fine-grained and small-sized CUB200-2011 FSCIL settings as shown in Appendix Table, HardNet (WSN) also shows comparable results with the baselines, and SoftNet outperforms others as denoted in \Cref{tab:main_cub200_10way_5shot}. In this FSCIL setting, we acquired the best performances of SoftNet through the specific parameter selections. As of now, our SoftNet achieves state-of-the-art results on the three datasets.

\subsection{Additional Comparisons with SOTA}
\textbf{Comparisons with SOTA}. We compare SoftNet with the following state-of-art-methods on TOPIC class split~\cite{tao2020few} of three benchmark datasets - CIFAR100 (Appendix, Table. 7), miniImageNet (Appendix, Table. 8), and CUB-200-2011 (Appendix, Table. 9). We summarize the current FSCIL methods such as \textbf{CEC}~\cite{zhang2021few}, \textbf{LIMIT}~\cite{zhou2022few}, \textbf{MetaFSCIL}~\cite{chi2022metafscil}, \textbf{C-FSCIL}~\cite{hersche2022constrained}, \textbf{Subspace Reg.}~\cite{akyurek2021subspace}, \textbf{Entropy-Reg}~\cite{liu2022few}, and \textbf{ALICE}~\cite{peng2022few}. Leveraged by regularized backbone ResNet, SoftNet outperformed all existing current works on CIFAR100,  miniImageNet. On CUB-200-201, the performances of SoftNet were comparable with those of ALICE and LIMIT, considering that ALICE used class/data augmentations and LIMIT added an extra multi-head attention layer.

\section{Conclusion}
Inspired by \emph{Regularized Lottery Ticket Hypothesis (RLTH)}, which states that competitive smooth (non-binary) subnetworks exist within a dense network in continual learning tasks, we investigated the performances of the proposed two architecture-based continual learning methods referred to as Winning SubNetworks (WSN) which sequentially learns and selects an optimal binary-subnetwork (WSN) and an optimal non-binary Soft-Subnetwork (SoftNet) for each task, respectively. Specifically, WSN and SoftNet jointly learned the regularized model weights and task-adaptive non-binary masks of subnetworks associated with each task whilst attempting to select a small set of weights to be activated (winning ticket) by reusing weights of the prior subnetworks. The proposed WSN and SoftNet were inherently immune to catastrophic forgetting as each selected subnetwork model does not infringe upon other subnetworks in Task Incremental Learning (TIL). In TIL, binary masks spawned per winning ticket were encoded into one N-bit binary digit mask, then compressed using Huffman coding for a sub-linear increase in network capacity to the number of tasks. Surprisingly, we observed that in the inference step, SoftNet generated by injecting small noises to the backgrounds of acquired WSN (holding the foregrounds of WSN) provides excellent forward transfer power for future tasks in TIL. Softnet showed its effectiveness over WSN in regularizing parameters to tackle the overfitting, to a few examples in Few-shot Class Incremental Learning (FSCIL). 


\bibliographystyle{plain}
\bibliography{reference}

\begin{thebibliography}{10}

\bibitem{akyurek2021subspace}
Afra~Feyza Aky{\"u}rek, Ekin Aky{\"u}rek, Derry Wijaya, and Jacob Andreas.
\newblock Subspace regularizers for few-shot class incremental learning.
\newblock {\em arXiv preprint arXiv:2110.07059}, 2021.

\bibitem{aljundi2019online}
Rahaf Aljundi, Eugene Belilovsky, Tinne Tuytelaars, Laurent Charlin, Massimo
  Caccia, Min Lin, and Lucas Page-Caccia.
\newblock Online continual learning with maximal interfered retrieval.
\newblock In {\em Advances in Neural Information Processing Systems (NeurIPS)},
  2019.

\bibitem{Bengio2013}
Yoshua Bengio, Nicholas L{\'{e}}onard, and Aaron~C. Courville.
\newblock Estimating or propagating gradients through stochastic neurons for
  conditional computation.
\newblock {\em CoRR}, 2013.

\bibitem{bottou2018optimization}
L{\'e}on Bottou, Frank~E Curtis, and Jorge Nocedal.
\newblock Optimization methods for large-scale machine learning.
\newblock {\em Siam Review}, 60(2):223--311, 2018.

\bibitem{boyd2004convex}
Stephen Boyd, Stephen~P Boyd, and Lieven Vandenberghe.
\newblock {\em Convex optimization}.
\newblock Cambridge university press, 2004.

\bibitem{Yaroslav2011notMNIST}
Yaroslav Bulatov.
\newblock notmnist dataset.
\newblock 2011.

\bibitem{chaudhry2020continual}
Arslan Chaudhry, Naeemullah Khan, Puneet~K Dokania, and Philip~HS Torr.
\newblock Continual learning in low-rank orthogonal subspaces.
\newblock In {\em Advances in Neural Information Processing Systems (NeurIPS)},
  2020.

\bibitem{chaudhry2018efficient}
Arslan Chaudhry, Marc'Aurelio Ranzato, Marcus Rohrbach, and Mohamed Elhoseiny.
\newblock Efficient lifelong learning with a-gem.
\newblock In {\em Proceedings of the International Conference on Learning
  Representations (ICLR)}, 2019.

\bibitem{chaudhry2019continual}
Arslan Chaudhry, Marcus Rohrbach, Mohamed Elhoseiny, Thalaiyasingam Ajanthan,
  Puneet~K Dokania, Philip~HS Torr, and M~Ranzato.
\newblock Continual learning with tiny episodic memories.
\newblock {\em arXiv preprint arXiv:1902.10486}, 2019.

\bibitem{chen2020incremental}
Kuilin Chen and Chi-Guhn Lee.
\newblock Incremental few-shot learning via vector quantization in deep
  embedded space.
\newblock In {\em International Conference on Learning Representations}, 2020.

\bibitem{Chen2021lifelonglottery}
Tianlong Chen, Zhenyu Zhang, Sijia Liu, Shiyu Chang, and Zhangyang Wang.
\newblock Long live the lottery: The existence of winning tickets in lifelong
  learning.
\newblock In {\em Proceedings of the International Conference on Learning
  Representations (ICLR)}, 2021.

\bibitem{cheraghian2021semantic}
Ali Cheraghian, Shafin Rahman, Pengfei Fang, Soumava~Kumar Roy, Lars Petersson,
  and Mehrtash Harandi.
\newblock Semantic-aware knowledge distillation for few-shot class-incremental
  learning.
\newblock In {\em Proceedings of the IEEE/CVF Conference on Computer Vision and
  Pattern Recognition}, pages 2534--2543, 2021.

\bibitem{chi2022metafscil}
Zhixiang Chi, Li~Gu, Huan Liu, Yang Wang, Yuanhao Yu, and Jin Tang.
\newblock Metafscil: A meta-learning approach for few-shot class incremental
  learning.
\newblock In {\em Proceedings of the IEEE/CVF Conference on Computer Vision and
  Pattern Recognition}, pages 14166--14175, 2022.

\bibitem{chijiwa2022metaticket}
Daiki Chijiwa, Shin'ya Yamaguchi, Atsutoshi Kumagai, and Yasutoshi Ida.
\newblock Meta-ticket: Finding optimal subnetworks for few-shot learning within
  randomly initialized neural networks.
\newblock In {\em Advances in Neural Information Processing Systems}, 2022.

\bibitem{dasarathy1980nosing}
Belur~V Dasarathy.
\newblock Nosing around the neighborhood: A new system structure and
  classification rule for recognition in partially exposed environments.
\newblock {\em IEEE Transactions on Pattern Analysis and Machine Intelligence},
  (1):67--71, 1980.

\bibitem{deng2021flattening}
Danruo Deng, Guangyong Chen, Jianye Hao, Qiong Wang, and Pheng-Ann Heng.
\newblock Flattening sharpness for dynamic gradient projection memory benefits
  continual learning.
\newblock In {\em Advances in Neural Information Processing Systems (NeurIPS)},
  2021.

\bibitem{Denil2013}
Misha Denil, Babak Shakibi, Laurent Dinh, Marc~Aurelio Ranzato, and Nando
  de~Freitas.
\newblock Predicting parameters in deep learning.
\newblock In {\em Advances in Neural Information Processing Systems (NeurIPS)},
  2013.

\bibitem{frankle2018lottery}
Jonathan Frankle and Michael Carbin.
\newblock The lottery ticket hypothesis: Finding sparse, trainable neural
  networks.
\newblock In {\em Proceedings of the International Conference on Learning
  Representations (ICLR)}, 2019.

\bibitem{golkar2019continual}
Siavash Golkar, Michael Kagan, and Kyunghyun Cho.
\newblock Continual learning via neural pruning.
\newblock {\em arXiv preprint arXiv:1903.04476}, 2019.

\bibitem{gupta2020maml}
Gunshi Gupta, Karmesh Yadav, and Liam Paull.
\newblock La-maml: Look-ahead meta learning for continual learning.
\newblock In {\em Advances in Neural Information Processing Systems (NeurIPS)},
  2020.

\bibitem{Han2016learning_both_weights_struct}
Song Han, Jeff Pool, John Tran, and William Dally.
\newblock Learning both weights and connections for efficient neural network.
\newblock In {\em Proceedings of the International Conference on Learning
  Representations (ICLR)}, 2016.

\bibitem{hart1968condensed}
Peter Hart.
\newblock The condensed nearest neighbor rule (corresp.).
\newblock {\em IEEE transactions on information theory}, 14(3):515--516, 1968.

\bibitem{hassabis2017neuroscience}
Demis Hassabis, Dharshan Kumaran, Christopher Summerfield, and Matthew
  Botvinick.
\newblock Neuroscience-inspired artificial intelligence.
\newblock {\em Neuron}, 95(2):245--258, 2017.

\bibitem{he2016deep}
Kaiming He, Xiangyu Zhang, Shaoqing Ren, and Jian Sun.
\newblock Deep residual learning for image recognition.
\newblock In {\em Proceedings of the IEEE conference on computer vision and
  pattern recognition}, pages 770--778, 2016.

\bibitem{he2018overcoming}
Xu~He and Herbert Jaeger.
\newblock Overcoming catastrophic interference using conceptor-aided
  backpropagation.
\newblock In {\em International Conference on Learning Representations}, 2018.

\bibitem{he2019task}
Xu~He, Jakub Sygnowski, Alexandre Galashov, Andrei~A Rusu, Yee~Whye Teh, and
  Razvan Pascanu.
\newblock Task agnostic continual learning via meta learning.
\newblock {\em arXiv preprint arXiv:1906.05201}, 2019.

\bibitem{hersche2022constrained}
Michael Hersche, Geethan Karunaratne, Giovanni Cherubini, Luca Benini, Abu
  Sebastian, and Abbas Rahimi.
\newblock Constrained few-shot class-incremental learning.
\newblock In {\em Proceedings of the IEEE/CVF Conference on Computer Vision and
  Pattern Recognition}, pages 9057--9067, 2022.

\bibitem{Hinton2012}
Geoffrey Hinton.
\newblock Neural networks for machine learning, 2012.

\bibitem{hinton1993keeping}
Geoffrey~E Hinton and Drew Van~Camp.
\newblock Keeping the neural networks simple by minimizing the description
  length of the weights.
\newblock In {\em Proceedings of the sixth annual conference on Computational
  learning theory}, pages 5--13, 1993.

\bibitem{hochreiter1994simplifying}
Sepp Hochreiter and J{\"u}rgen Schmidhuber.
\newblock Simplifying neural nets by discovering flat minima.
\newblock {\em Advances in neural information processing systems}, 7, 1994.

\bibitem{hou2019learning}
Saihui Hou, Xinyu Pan, Chen~Change Loy, Zilei Wang, and Dahua Lin.
\newblock Learning a unified classifier incrementally via rebalancing.
\newblock In {\em Proceedings of the IEEE/CVF Conference on Computer Vision and
  Pattern Recognition}, pages 831--839, 2019.

\bibitem{huffman1952method}
David~A Huffman.
\newblock A method for the construction of minimum-redundancy codes.
\newblock {\em Proceedings of the IRE}, 40(9):1098--1101, 1952.

\bibitem{jiang2019fantastic}
Yiding Jiang, Behnam Neyshabur, Hossein Mobahi, Dilip Krishnan, and Samy
  Bengio.
\newblock Fantastic generalization measures and where to find them.
\newblock {\em arXiv preprint arXiv:1912.02178}, 2019.

\bibitem{Jung2020}
Sangwon Jung, Hongjoon Ahn, Sungmin Cha, and Taesup Moon.
\newblock Continual learning with node-importance based adaptive group sparse
  regularization.
\newblock In {\em Advances in Neural Information Processing Systems (NeurIPS)},
  2020.

\bibitem{kang2019decoupling}
Bingyi Kang, Saining Xie, Marcus Rohrbach, Zhicheng Yan, Albert Gordo, Jiashi
  Feng, and Yannis Kalantidis.
\newblock Decoupling representation and classifier for long-tailed recognition.
\newblock {\em arXiv preprint arXiv:1910.09217}, 2019.

\bibitem{kang2022forget}
Haeyong Kang, Rusty John~Lloyd Mina, Sultan Rizky~Hikmawan Madjid, Jaehong
  Yoon, Mark Hasegawa-Johnson, Sung~Ju Hwang, and Chang~D Yoo.
\newblock Forget-free continual learning with winning subnetworks.
\newblock In {\em International Conference on Machine Learning}, pages
  10734--10750. PMLR, 2022.

\bibitem{kang2022soft}
Haeyong Kang, Jaehong Yoon, Sultan Rizky~Hikmawan Madjid, Sung~Ju Hwang, and
  Chang~D Yoo.
\newblock On the soft-subnetwork for few-shot class incremental learning.
\newblock {\em arXiv preprint arXiv:2209.07529}, 2022.

\bibitem{Kirkpatrick2017}
James Kirkpatrick, Razvan Pascanu, Neil Rabinowitz, Joel Veness, Guillaume
  Desjardins, Andrei~A Rusu, Kieran Milan, John Quan, Tiago Ramalho, Agnieszka
  Grabska-Barwinska, Demis Hassabis, Claudia Clopath, Dharshan Kumaran, and
  Raia Hadsell.
\newblock Overcoming catastrophic forgetting in neural networks.
\newblock 2017.

\bibitem{Krizhevsky2009}
Alex Krizhevsky, Geoffrey Hinton, et~al.
\newblock Learning multiple layers of features from tiny images.
\newblock 2009.

\bibitem{krizhevsky2012imagenet}
Alex Krizhevsky, Ilya Sutskever, and Geoffrey~E Hinton.
\newblock Imagenet classification with deep convolutional neural networks.
\newblock {\em Advances in neural information processing systems},
  25:1097--1105, 2012.

\bibitem{KumarA2012icml}
Abhishek Kumar and Hal Daume~III.
\newblock Learning task grouping and overlap in multi-task learning.
\newblock In {\em Proceedings of the International Conference on Machine
  Learning (ICML)}, 2012.

\bibitem{LeCun1998}
Yann LeCun.
\newblock The mnist database of handwritten digits.
\newblock 1998.

\bibitem{Li2016pruning_convnets}
Hao Li, Asim Kadav, Igor Durdanovic, Hanan Samet, and Hans~Peter Graf.
\newblock Pruning filters for efficient convnets.
\newblock {\em arXiv preprint arXiv:1608.08710}, 2016.

\bibitem{li2019learn}
Xilai Li, Yingbo Zhou, Tianfu Wu, Richard Socher, and Caiming Xiong.
\newblock Learn to grow: A continual structure learning framework for
  overcoming catastrophic forgetting.
\newblock In {\em Proceedings of the International Conference on Machine
  Learning (ICML)}, 2019.

\bibitem{LiZ2016eccv}
Zhizhong Li and Derek Hoiem.
\newblock Learning without forgetting.
\newblock In {\em Proceedings of the European Conference on Computer Vision
  (ECCV)}, 2016.

\bibitem{Liu2018Hier}
Hanxiao Liu, Karen Simonyan, Oriol Vinyals, Chrisantha Fernando, and Koray
  Kavukcuoglu.
\newblock Hierarchical representations for efficient architecture search, 2017.

\bibitem{liu2022few}
Huan Liu, Li~Gu, Zhixiang Chi, Yang Wang, Yuanhao Yu, Jun Chen, and Jin Tang.
\newblock Few-shot class-incremental learning via entropy-regularized data-free
  replay.
\newblock {\em arXiv preprint arXiv:2207.11213}, 2022.

\bibitem{Lopez-Paz2017}
David Lopez-Paz and Marc'Aurelio Ranzato.
\newblock Gradient episodic memory for continual learning.
\newblock In {\em Advances in Neural Information Processing Systems (NeurIPS)},
  2017.

\bibitem{mallya2018piggyback}
Arun Mallya, Dillon Davis, and Svetlana Lazebnik.
\newblock Piggyback: Adapting a single network to multiple tasks by learning to
  mask weights.
\newblock In {\em Proceedings of the European Conference on Computer Vision
  (ECCV)}, 2018.

\bibitem{mallya2018packnet}
Arun Mallya and Svetlana Lazebnik.
\newblock Packnet: Adding multiple tasks to a single network by iterative
  pruning.
\newblock In {\em Proceedings of the IEEE conference on Computer Vision and
  Pattern Recognition}, pages 7765--7773, 2018.

\bibitem{masse2018alleviating}
Nicolas~Y Masse, Gregory~D Grant, and David~J Freedman.
\newblock Alleviating catastrophic forgetting using context-dependent gating
  and synaptic stabilization.
\newblock {\em Proceedings of the National Academy of Sciences},
  115(44):E10467--E10475, 2018.

\bibitem{mazumder2021few}
Pratik Mazumder, Pravendra Singh, and Piyush Rai.
\newblock Few-shot lifelong learning.
\newblock {\em arXiv preprint arXiv:2103.00991}, 2021.

\bibitem{McCloskey1989}
Michael McCloskey and Neal~J Cohen.
\newblock Catastrophic interference in connectionist networks: The sequential
  learning problem.
\newblock In {\em Psychology of learning and motivation}, volume~24, pages
  109--165. Elsevier, 1989.

\bibitem{mensink2013distance}
Thomas Mensink, Jakob Verbeek, Florent Perronnin, and Gabriela Csurka.
\newblock Distance-based image classification: Generalizing to new classes at
  near-zero cost.
\newblock {\em IEEE transactions on pattern analysis and machine intelligence},
  35(11):2624--2637, 2013.

\bibitem{mirzadeh2020linear}
Seyed~Iman Mirzadeh, Mehrdad Farajtabar, Dilan Gorur, Razvan Pascanu, and
  Hassan Ghasemzadeh.
\newblock Linear mode connectivity in multitask and continual learning.
\newblock In {\em Proceedings of the International Conference on Learning
  Representations (ICLR)}, 2021.

\bibitem{Netzer2011SVHN}
Yuval Netzer, Tao Wang, Adam Coates, Alessandro Bissacco, Bo~Wu, and Andrew~Y.
  Ng.
\newblock Reading digits in natural images with unsupervised feature learning.
\newblock In {\em NIPS Workshop on Deep Learning and Unsupervised Feature
  Learning 2011}, 2011.

\bibitem{peng2022few}
Can Peng, Kun Zhao, Tianren Wang, Meng Li, and Brian~C Lovell.
\newblock Few-shot class-incremental learning from an open-set perspective.
\newblock In {\em European Conference on Computer Vision}, pages 382--397.
  Springer, 2022.

\bibitem{Ramanujan2020}
Vivek Ramanujan, Mitchell Wortsman, Aniruddha Kembhavi, Ali Farhadi, and
  Mohammad Rastegari.
\newblock What's hidden in a randomly weighted neural network?
\newblock In {\em Proceedings of the IEEE International Conference on Computer
  Vision and Pattern Recognition (CVPR)}, 2020.

\bibitem{rebuffi2017icarl}
Sylvestre-Alvise Rebuffi, Alexander Kolesnikov, Georg Sperl, and Christoph~H
  Lampert.
\newblock icarl: Incremental classifier and representation learning.
\newblock In {\em Proceedings of the IEEE conference on Computer Vision and
  Pattern Recognition}, pages 2001--2010, 2017.

\bibitem{riemer2018learning}
Matthew Riemer, Ignacio Cases, Robert Ajemian, Miao Liu, Irina Rish, Yuhai Tu,
  and Gerald Tesauro.
\newblock Learning to learn without forgetting by maximizing transfer and
  minimizing interference.
\newblock {\em arXiv preprint arXiv:1810.11910}, 2018.

\bibitem{rusu2016progressive}
Andrei~A Rusu, Neil~C Rabinowitz, Guillaume Desjardins, Hubert Soyer, James
  Kirkpatrick, Koray Kavukcuoglu, Razvan Pascanu, and Raia Hadsell.
\newblock Progressive neural networks.
\newblock {\em arXiv preprint arXiv:1606.04671}, 2016.

\bibitem{Saha2021}
Gobinda Saha, Isha Garg, and Kaushik Roy.
\newblock Gradient projection memory for continual learning.
\newblock In {\em Proceedings of the International Conference on Learning
  Representations (ICLR)}, 2021.

\bibitem{serra2018overcoming}
Joan Serra, Didac Suris, Marius Miron, and Alexandros Karatzoglou.
\newblock Overcoming catastrophic forgetting with hard attention to the task.
\newblock In {\em International Conference on Machine Learning}, pages
  4548--4557. PMLR, 2018.

\bibitem{Serra2018}
Joan Serrà, Didac Suris, Marius Miron, and Alexandros Karatzoglou.
\newblock Overcoming catastrophic forgetting with hard attention to the task.
\newblock In {\em Proceedings of the International Conference on Machine
  Learning (ICML)}, 2018.

\bibitem{shi2021overcoming}
Guangyuan Shi, Jiaxin Chen, Wenlong Zhang, Li-Ming Zhan, and Xiao-Ming Wu.
\newblock Overcoming catastrophic forgetting in incremental few-shot learning
  by finding flat minima.
\newblock {\em Advances in Neural Information Processing Systems}, 34, 2021.

\bibitem{ShinH2017nips}
Hanul Shin, Jung~Kwon Lee, Jaehon Kim, and Jiwon Kim.
\newblock Continual learning with deep generative replay.
\newblock In {\em Advances in Neural Information Processing Systems (NeurIPS)},
  2017.

\bibitem{Stanford}
Stanford.
\newblock Available online at http://cs231n.stanford.edu/tiny-imagenet-200.zip.
\newblock {\em CS 231N}, 2021.

\bibitem{tao2020few}
Xiaoyu Tao, Xiaopeng Hong, Xinyuan Chang, Songlin Dong, Xing Wei, and Yihong
  Gong.
\newblock Few-shot class-incremental learning.
\newblock In {\em Proceedings of the IEEE/CVF Conference on Computer Vision and
  Pattern Recognition}, pages 12183--12192, 2020.

\bibitem{ThrunS1995}
Sebastian Thrun.
\newblock {\em A Lifelong Learning Perspective for Mobile Robot Control}.
\newblock Elsevier, 1995.

\bibitem{titsias2019functional}
Michalis~K Titsias, Jonathan Schwarz, Alexander G de~G Matthews, Razvan
  Pascanu, and Yee~Whye Teh.
\newblock Functional regularisation for continual learning with gaussian
  processes.
\newblock In {\em Proceedings of the International Conference on Learning
  Representations (ICLR)}, 2020.

\bibitem{wortsman2019discovering}
Mitchell Wortsman, Ali Farhadi, and Mohammad Rastegari.
\newblock Discovering neural wirings.
\newblock {\em Advances in Neural Information Processing Systems}, 32, 2019.

\bibitem{wortsman2020supermasks}
Mitchell Wortsman, Vivek Ramanujan, Rosanne Liu, Aniruddha Kembhavi, Mohammad
  Rastegari, Jason Yosinski, and Ali Farhadi.
\newblock Supermasks in superposition.
\newblock In {\em Advances in Neural Information Processing Systems (NeurIPS)},
  2020.

\bibitem{wu2019large}
Yue Wu, Yinpeng Chen, Lijuan Wang, Yuancheng Ye, Zicheng Liu, Yandong Guo, and
  Yun Fu.
\newblock Large scale incremental learning.
\newblock In {\em Proceedings of the IEEE/CVF Conference on Computer Vision and
  Pattern Recognition}, pages 374--382, 2019.

\bibitem{Xiao2017Fashion}
Han Xiao, Kashif Rasul, and Roland Vollgraf.
\newblock Fashion-mnist: a novel image dataset for benchmarking machine
  learning algorithms.
\newblock {\em arXiv}, 2017.

\bibitem{xu2018reinforced}
Ju~Xu and Zhanxing Zhu.
\newblock Reinforced continual learning.
\newblock In {\em Advances in Neural Information Processing Systems (NeurIPS)},
  2018.

\bibitem{yao2020pyhessian}
Zhewei Yao, Amir Gholami, Kurt Keutzer, and Michael~W Mahoney.
\newblock Pyhessian: Neural networks through the lens of the hessian.
\newblock In {\em 2020 IEEE international conference on big data (Big data)},
  pages 581--590. IEEE, 2020.

\bibitem{ye2020good}
Mao Ye, Chengyue Gong, Lizhen Nie, Denny Zhou, Adam Klivans, and Qiang Liu.
\newblock Good subnetworks provably exist: Pruning via greedy forward
  selection.
\newblock In {\em International Conference on Machine Learning}, pages
  10820--10830. PMLR, 2020.

\bibitem{Yoon2020}
Jaehong Yoon, Saehoon Kim, Eunho Yang, and Sung~Ju Hwang.
\newblock Scalable and order-robust continual learning with additive parameter
  decomposition.
\newblock In {\em Proceedings of the International Conference on Learning
  Representations (ICLR)}, 2020.

\bibitem{yoon2022online}
Jaehong Yoon, Divyam Madaan, Eunho Yang, and Sung~Ju Hwang.
\newblock Online coreset selection for rehearsal-based continual learning.
\newblock In {\em Proceedings of the International Conference on Learning
  Representations (ICLR)}, 2022.

\bibitem{YoonJ2018iclr}
Jaehong Yoon, Eunho Yang, Jeongtae Lee, and Sung~Ju Hwang.
\newblock Lifelong learning with dynamically expandable networks.
\newblock In {\em Proceedings of the International Conference on Learning
  Representations (ICLR)}, 2018.

\bibitem{you2019drawing}
Haoran You, Chaojian Li, Pengfei Xu, Yonggan Fu, Yue Wang, Xiaohan Chen,
  Richard~G Baraniuk, Zhangyang Wang, and Yingyan Lin.
\newblock Drawing early-bird tickets: Towards more efficient training of deep
  networks.
\newblock {\em arXiv preprint arXiv:1909.11957}, 2019.

\bibitem{zenke2017continual}
Friedemann Zenke, Ben Poole, and Surya Ganguli.
\newblock Continual learning through synaptic intelligence.
\newblock In {\em International Conference on Machine Learning}, pages
  3987--3995. PMLR, 2017.

\bibitem{zhang2021few}
Chi Zhang, Nan Song, Guosheng Lin, Yun Zheng, Pan Pan, and Yinghui Xu.
\newblock Few-shot incremental learning with continually evolved classifiers.
\newblock In {\em Proceedings of the IEEE/CVF Conference on Computer Vision and
  Pattern Recognition}, pages 12455--12464, 2021.

\bibitem{zhou2022few}
Da-Wei Zhou, Han-Jia Ye, Liang Ma, Di~Xie, Shiliang Pu, and De-Chuan Zhan.
\newblock Few-shot class-incremental learning by sampling multi-phase tasks.
\newblock {\em IEEE Transactions on Pattern Analysis and Machine Intelligence},
  2022.

\bibitem{zhou2019deconstructing}
Hattie Zhou, Janice Lan, Rosanne Liu, and Jason Yosinski.
\newblock Deconstructing lottery tickets: Zeros, signs, and the supermask.
\newblock In {\em Advances in Neural Information Processing Systems (NeurIPS)},
  2019.

\bibitem{zhu2021self}
Kai Zhu, Yang Cao, Wei Zhai, Jie Cheng, and Zheng-Jun Zha.
\newblock Self-promoted prototype refinement for few-shot class-incremental
  learning.
\newblock In {\em Proceedings of the IEEE/CVF Conference on Computer Vision and
  Pattern Recognition}, pages 6801--6810, 2021.

\end{thebibliography}

\vspace{-0.16in}

\begin{IEEEbiography}[{\includegraphics[width=1in,height=1.25in,clip,keepaspectratio]{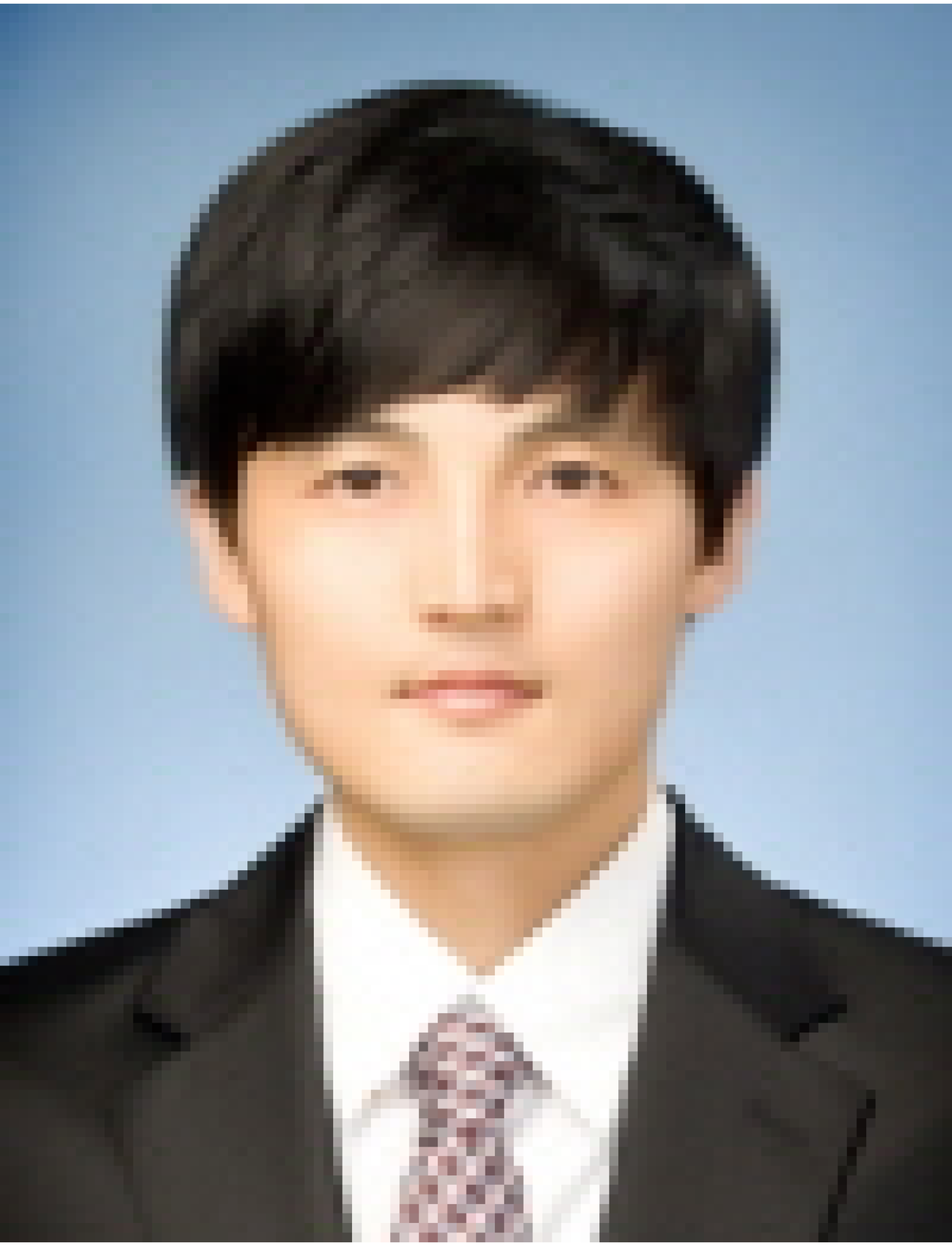}}]{Haeyong Kang}
(S'05) received the M.S. degree in Systems and Information Engineering from University of Tsukuba in 2007. From April 2007 to October 2010, he worked as an associate research engineer at LG Electronics. With working experiences at Korea Institute of Science and Technology (KIST) and the University of Tokyo, He is currently pursuing the Ph.D at School of Electrical Engineering, KAIST. His current research interests include unbiased machine learning and continual learning.
\end{IEEEbiography}


\begin{IEEEbiography}[{\includegraphics[width=1in,height=1.25in,clip,keepaspectratio]{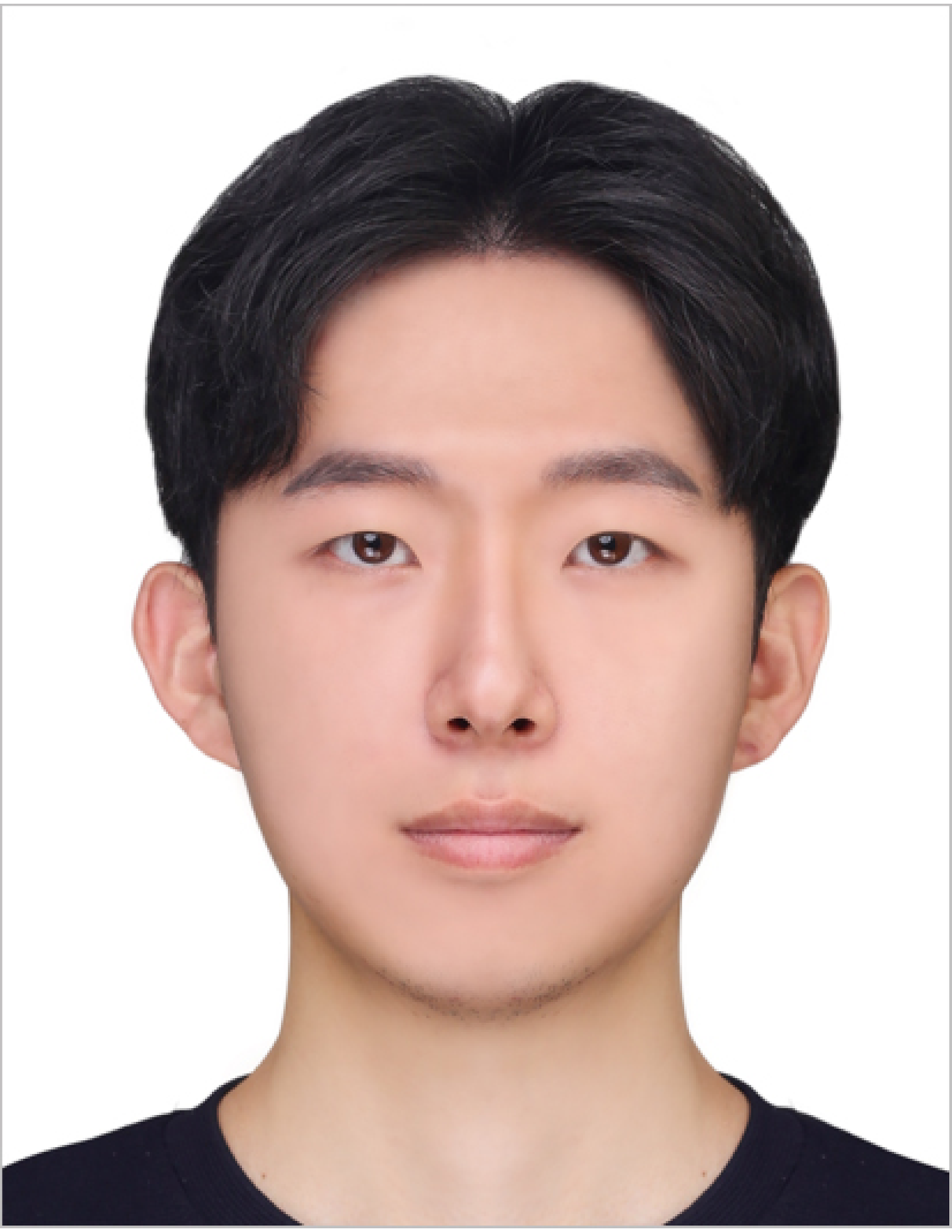}}]{Jaehong Yoon}
He received the B.S. and M.S. degrees in Computer Science from Ulsan National Institute of Science and Technology (UNIST), and received the Ph.D. degree in the School of Computing from Korea Advanced Institute of Science and Technology (KAIST). He is currently working as a postdoctoral research fellow at KAIST. His current research interests include efficient deep learning, on-device learning, and learning with real-world data.
\end{IEEEbiography}


\begin{IEEEbiography}[{\includegraphics[width=1in,height=1.25in,clip,keepaspectratio]{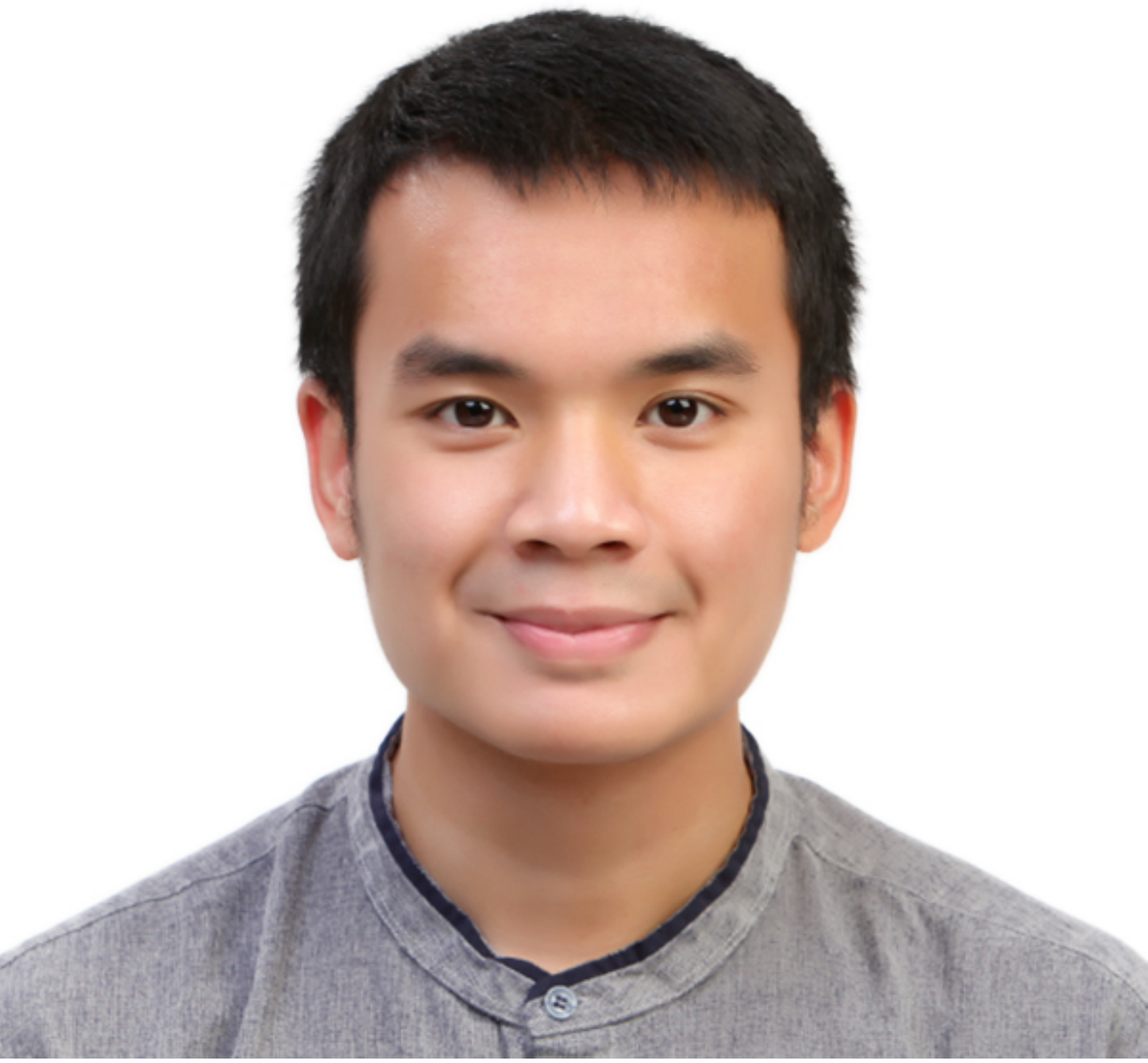}}]{Sultan Rizky Madjid}
received a B.S. degree in Electrical Engineering with a double major in Mechanical Engineering from KAIST in 2021 and an M.S. degree in Electrical Engineering from KAIST in 2023. His research interests include model compression, sparse representations in deep learning, and continual learning.
\end{IEEEbiography}


\begin{IEEEbiography}[{\includegraphics[width=1in,height=1.25in,clip,keepaspectratio]{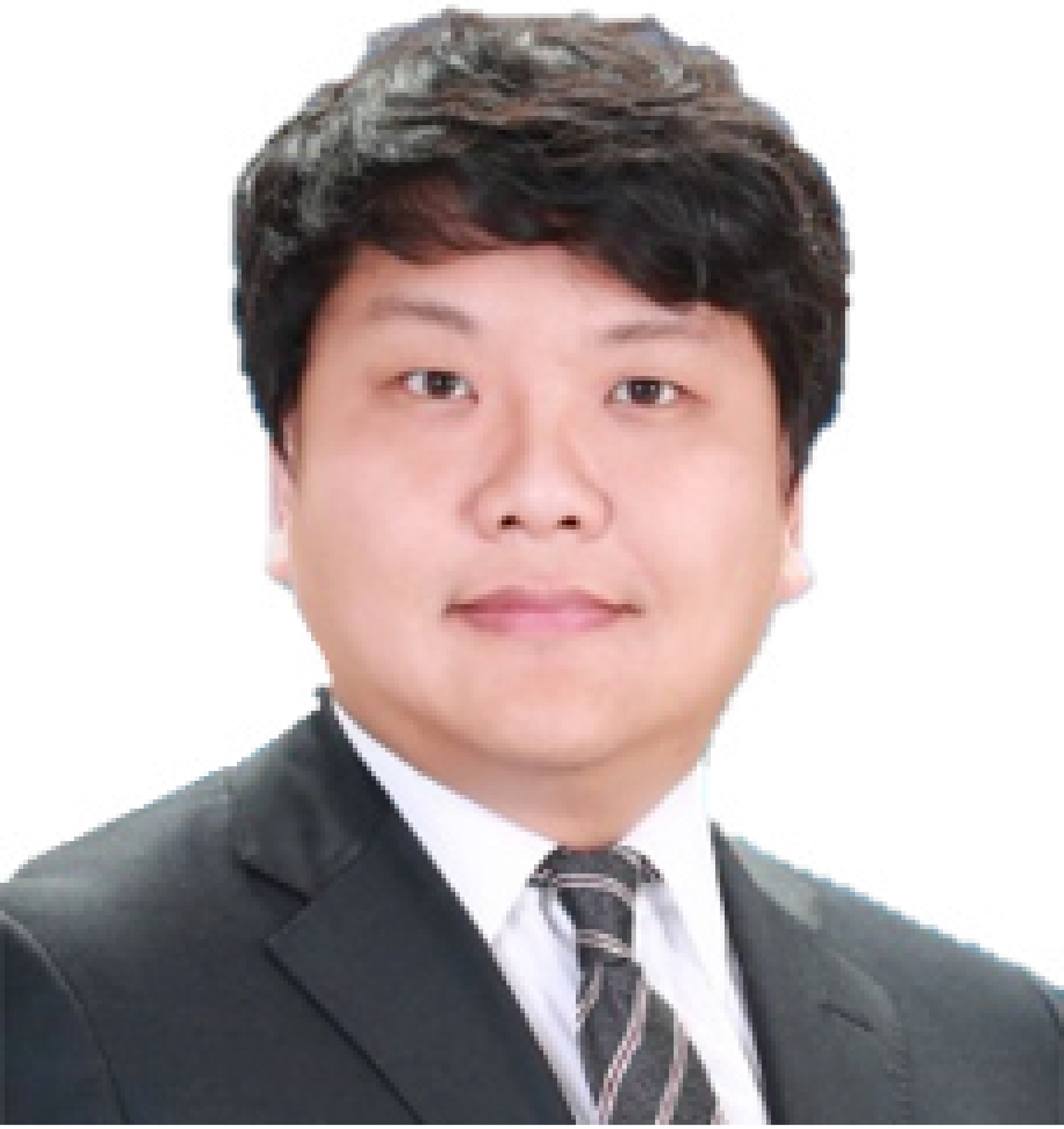}}]{Sung Ju Hwang}
He received the B.S. degree in Computer Science and Engineering from Seoul National University. He received the M.S. and Ph.D. degrees in Computer Science from The University of Texas at Austin. From September 2013 to August 2014, he was a postdoctoral research associate at Disney Research. From September 2013 to December 2017, he was an assistant professor in the School of Electric and Computer Engineering at UNIST. Since 2017, he has been on the faculty at the Korea Advanced Institute of Science and Technology (KAIST), where he is currently a KAIST Endowed Chair Professor in the Kim Jaechul School of Artificial Intelligence and School of Computing at KAIST. 
\end{IEEEbiography}


\begin{IEEEbiography}[{\includegraphics[width=1in,height=1.25in,clip,keepaspectratio]{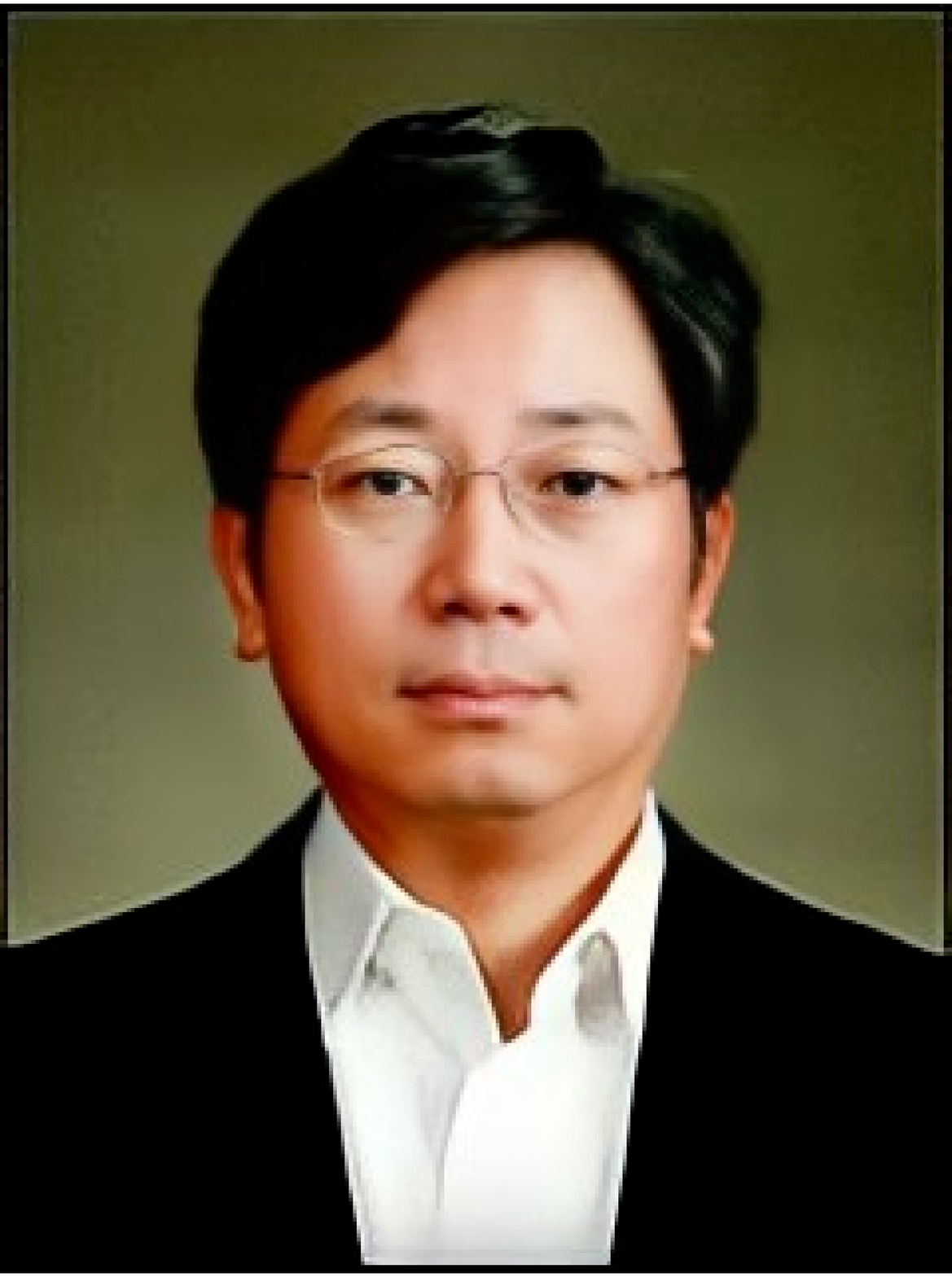}}]{Chang D. Yoo}
(Senior Member, IEEE) He received the B.S. degree in Engineering and Applied Science from the California Institute of Technology, the M.S. degree in Electrical Engineering from Cornell University, and the Ph.D. degree in Electrical Engineering from the Massachusetts Institute of Technology. From January 1997 to March 1999, he was Senior Researcher at Korea Telecom (KT). Since 1999, he has been on the faculty at the Korea Advanced Institute of Science and Technology (KAIST), where he is currently a Full Professor with tenure in the School of Electrical Engineering and an Adjunct Professor in the Department of Computer Science. He also served as Dean of the Office of Special Projects and Dean of the Office of International Relations.
\end{IEEEbiography}

\appendices
\section{Experimental Details of WSN for TIL}\label{app_sec:exper_detail}

We followed similar experimental setups (architectures and hyper-parameters) described in \cite{Saha2021} for baseline comparisons and explained in \cite{deng2021flattening} for SOTA comparisons.

\subsection{Architecture Details}
All the networks for our experiments are implemented in a multi-head setting. 
\noindent 
\textbf{Two-layered MLP:} In conducting the PMNIST experiments, we are following the exact setup as denoted by \cite{Saha2021} fully-connected network with two hidden layers of 100~\cite{Lopez-Paz2017}.

\noindent
\textbf{Reduced ResNet18:} In conducting the 5-Dataset experiments, we use a smaller version of ResNet18 with three times fewer feature maps across all layers as denoted by \cite{Lopez-Paz2017}.

\noindent
\textbf{Modified LeNet:} In conducting the Omniglot Rotation and CIFAR-100 Superclass experiments, we use a large variant of LeNet as the base network with 64-128-2500-1500 neurons based on \cite{Yoon2020}.

\noindent
\textbf{Modified AlexNet:} In conducting the split CIFAR-100 dataset, we use a modified version of AlexNet similar to \cite{Serra2018, Saha2021}.

\noindent
\textbf{4 Conv layers and 3 Fully connected layers:} For TinyImageNet, we use the same network architecture as \cite{gupta2020maml, deng2021flattening}.

\subsection{List of Hyperparameters}

\begin{table}[h!]
\vspace{-0.1in}
\small
\centering
\caption{\small  List of hyperparameters for the baselines and our WSN. Here, 'lr' and 'optim' represents (initial) learning rate and optimizer used for training. We represent PMNIST as 'perm', 5-Datasets as '5data', Omniglot Rotation as 'omniglot', CIFAR-100 Split as 'cifar100-split', and CIFAR-100 Superclass as 'cifar100-sc'.}

\resizebox{\columnwidth}{!}{%
\begin{tabular}{ll}
\toprule
\textbf{Methods} & \textbf{Hyperparameters} \\
\midrule
EWC & lr : 0.03 (perm) \\
& optim : sgd \\
& regularization coefficient : 1000 (perm) \\
\midrule
GPM & lr : 0.01(perm, omniglot), 0.1 (5data) \\
& optim : sgd \\
& $n_s$ : 300 (perm), 100 (5data), 125 (omniglot) \\
\midrule
PackNet & lr = 0.001 (perm, 5data, omniglot) \\
& optim : adam \\
& $c$ : 0.1 (perm), 0.2 (5data), 0.02 (omniglot) \\
\midrule
WSN (ours) & lr = 0.001 \\
& (perm, 5data, omniglot, \\
& cifar100-split, cifar100-sc, tinyimagenet) \\
& optim : adam \\
\bottomrule
\end{tabular}
}
\label{tab:main_hyper_table}
\end{table}

\noindent 
\Cref{tab:main_hyper_table} details the hyperparameter setup for the baselines and our approach.
$n_s$ in GPM denotes the number of random training examples sampled from the replay buffer to construct the representation matrix for each architecture layer. 

\begin{figure}[ht]
    \centering

    \setlength{\tabcolsep}{0pt}{%
    \begin{tabular}{cc}
    \includegraphics[width=0.5\columnwidth]{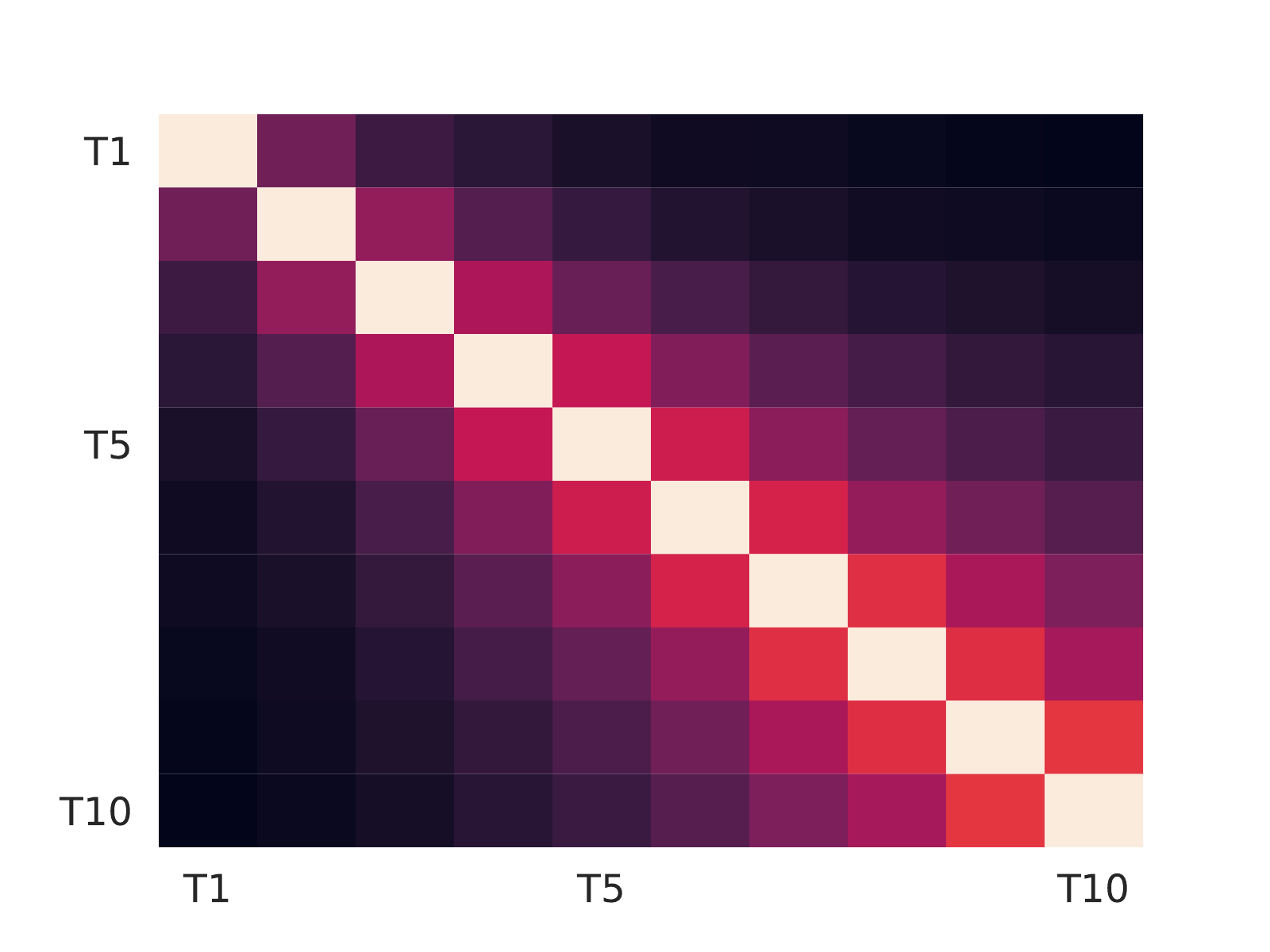} & 
    \includegraphics[width=0.5\columnwidth]{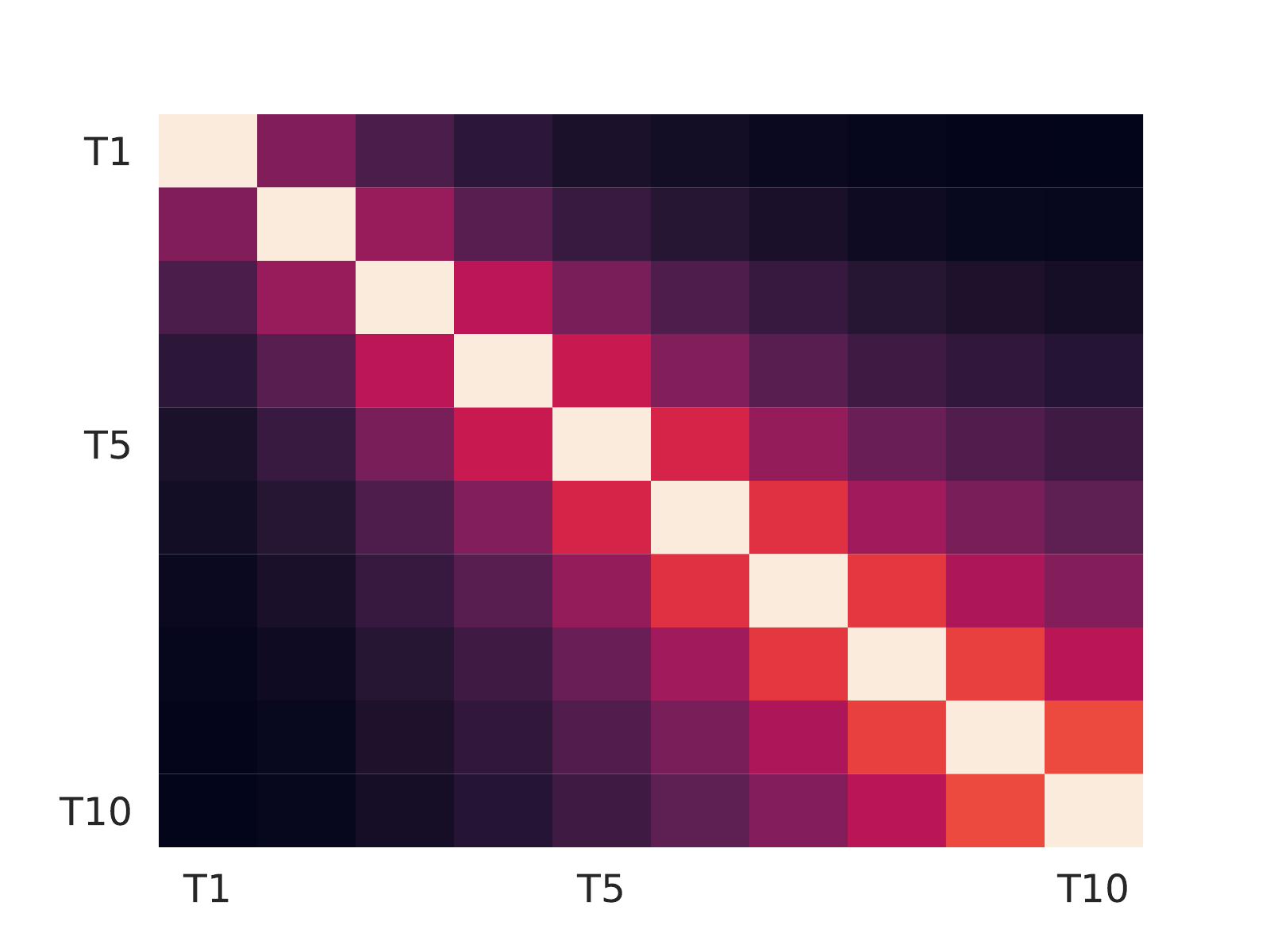} \\
    \small (a) $c=5\%$ & \small (b) $c=10\%$ \\
    \includegraphics[width=0.5\columnwidth]{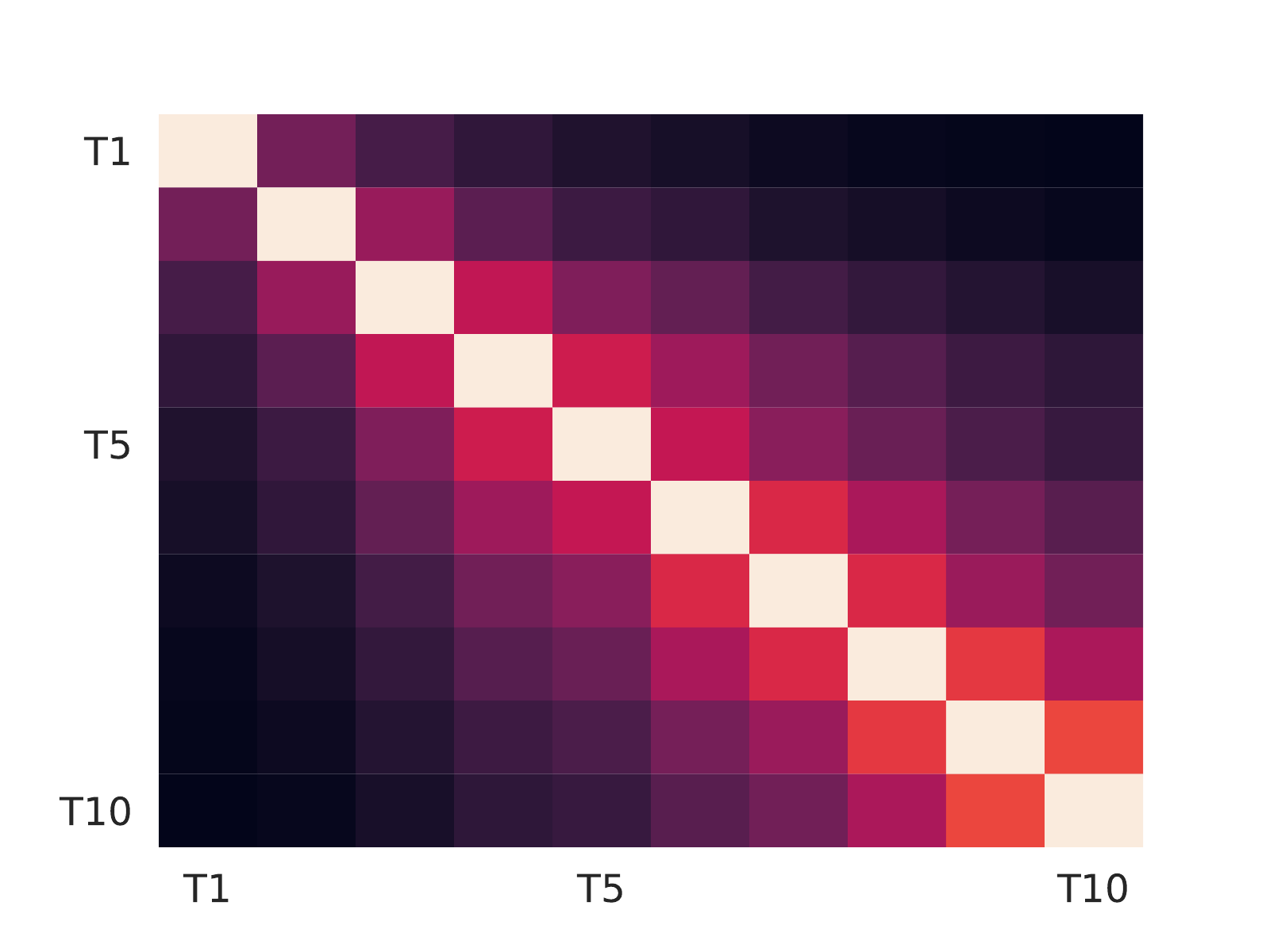} & 
    \includegraphics[width=0.5\columnwidth]{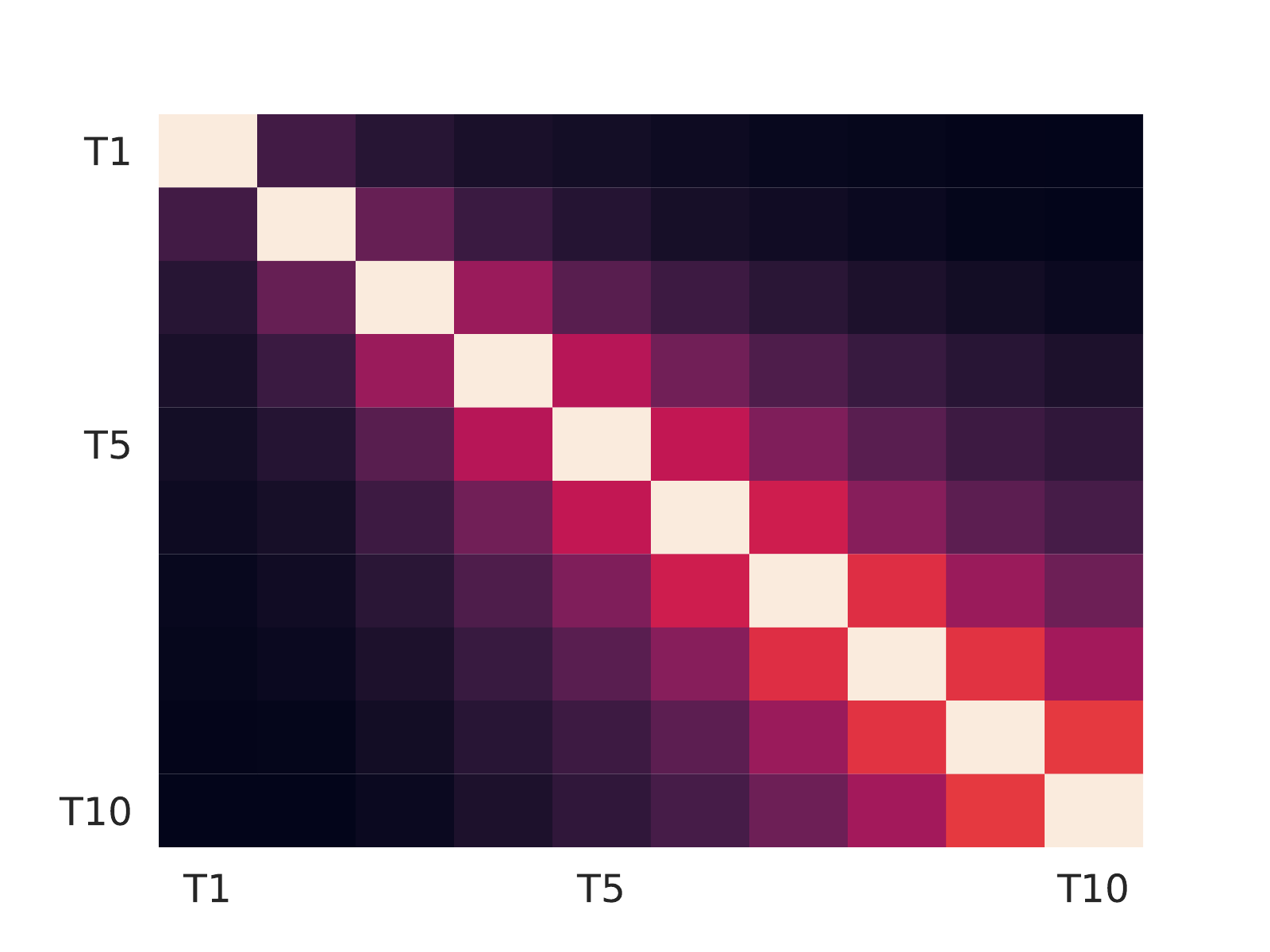} \\
    \small (c) $c=30\%$ & \small (d) $c=50\%$
    \end{tabular}
    }
    \caption{\footnotesize \textbf{Average Binary Map Correlation} on Sequence of PMNIST Experiments: the binary maps get overlapped with prior ones as the number of tasks increases.}
    \label{fig:main_pmnist_conf}
\end{figure}
\begin{figure}[ht]
    \centering
    \setlength{\tabcolsep}{0pt}{%
    \begin{tabular}{cc}
    \includegraphics[width=0.5\columnwidth]{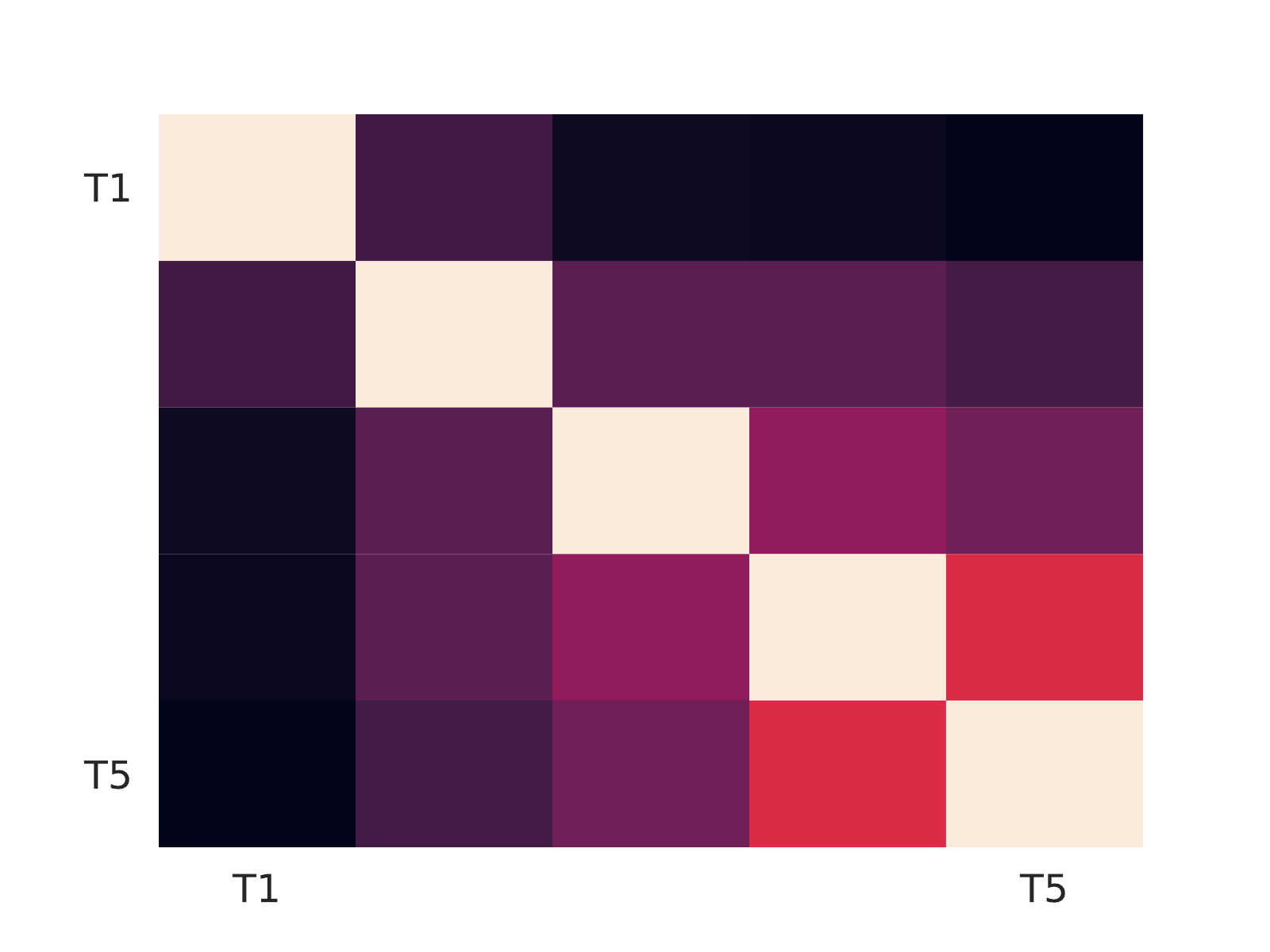} & 
    \includegraphics[width=0.5\columnwidth]{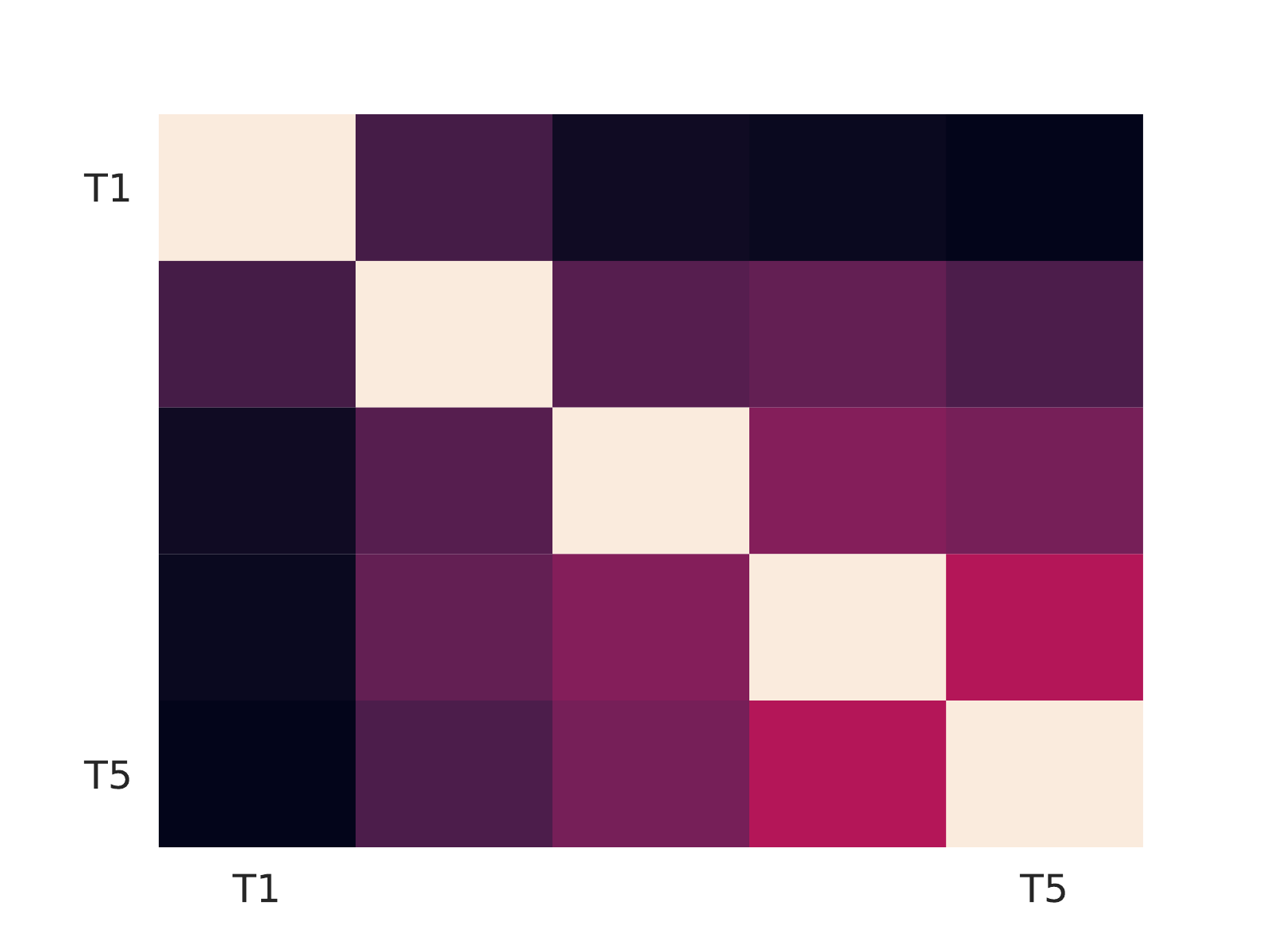} \\
    \small (a) $c=5\%$ & \small (b) $c=10\%$ \\
    \includegraphics[width=0.5\columnwidth]{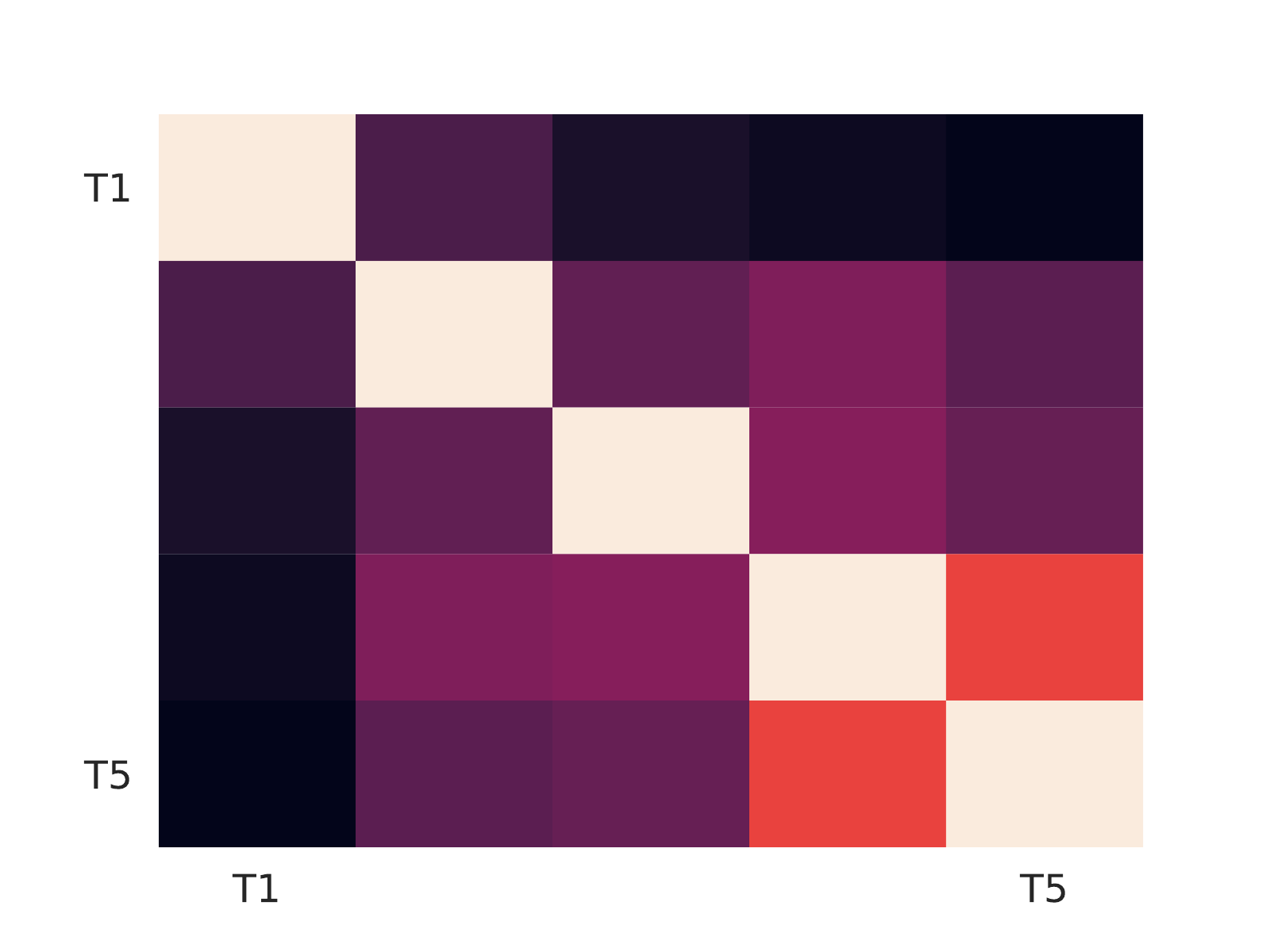} & 
    \includegraphics[width=0.5\columnwidth]{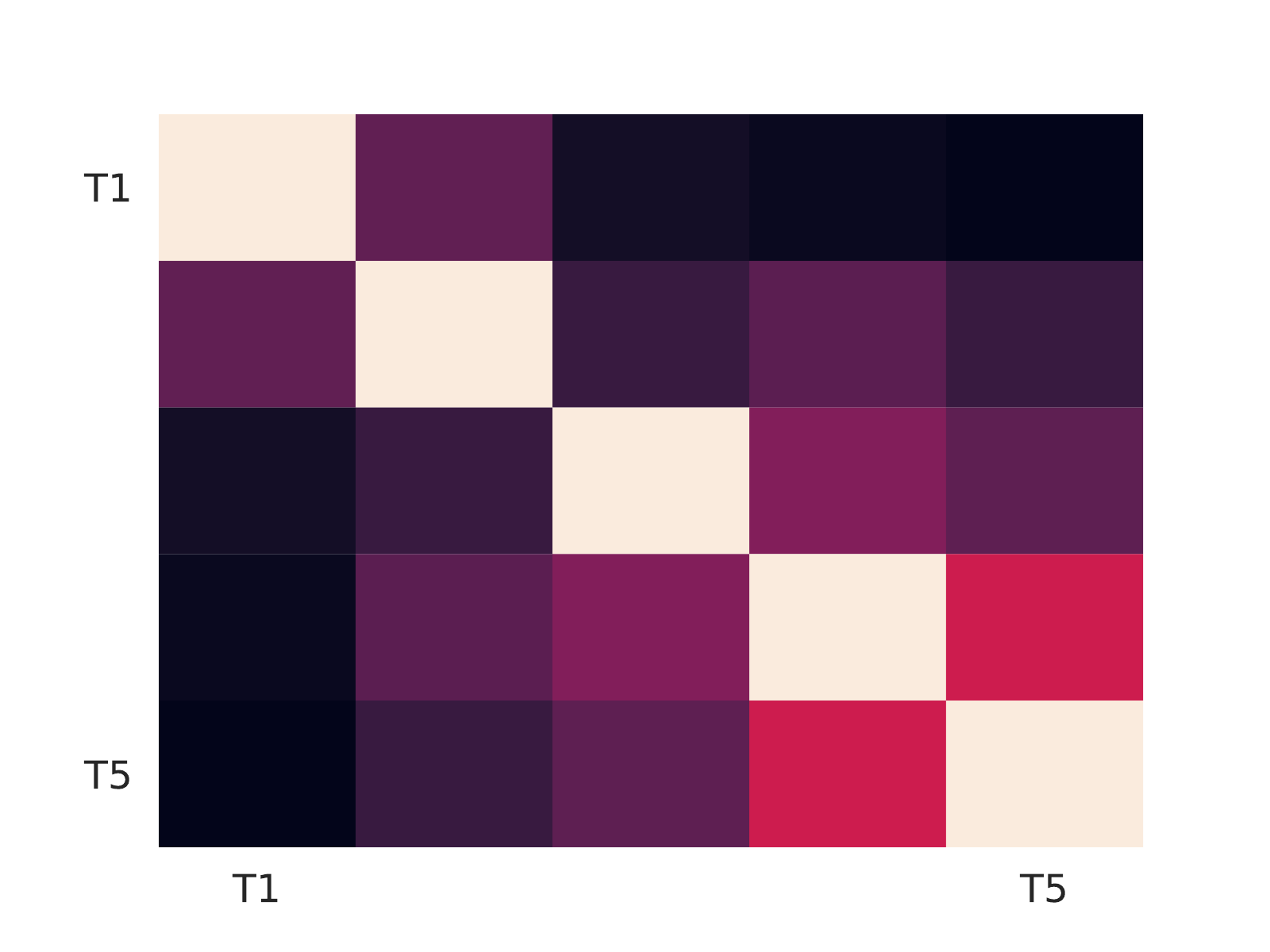} \\
    \small (c) $c=30\%$ & \small (d) $c=50\%$
    \end{tabular}
    }

    \caption{\footnotesize \textbf{Average Binary Map Correlation} on Sequence of 5-Dataset Experiments: the binary maps get overlapped with prior ones as the number of tasks increases.}
    \label{fig:main_5_data_conf}
\end{figure}

\begin{figure}[ht]
    \centering
    \setlength{\tabcolsep}{0pt}{%
    \begin{tabular}{cc}
    \includegraphics[width=0.5\columnwidth]{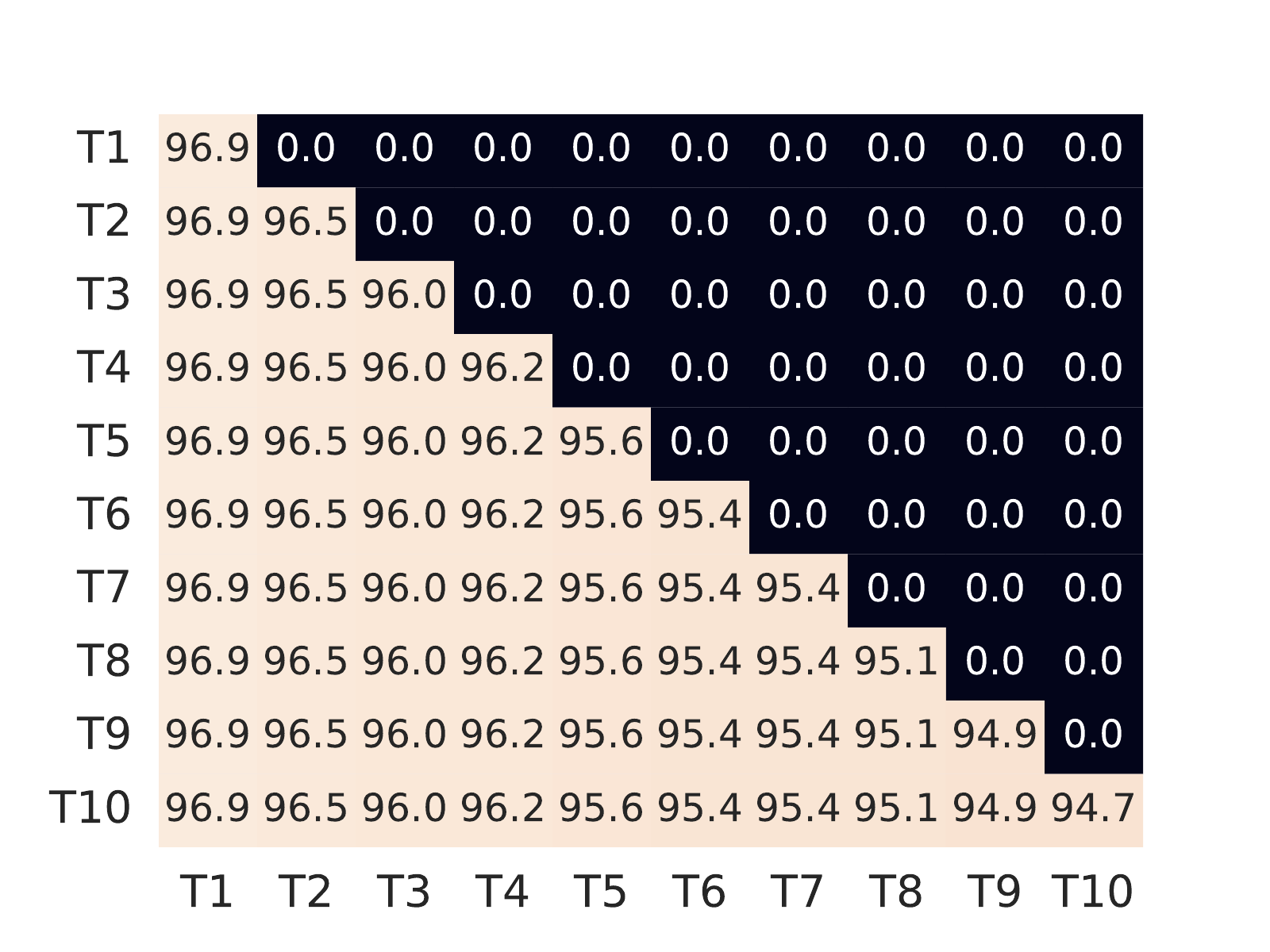} & 
    \includegraphics[width=0.5\columnwidth]{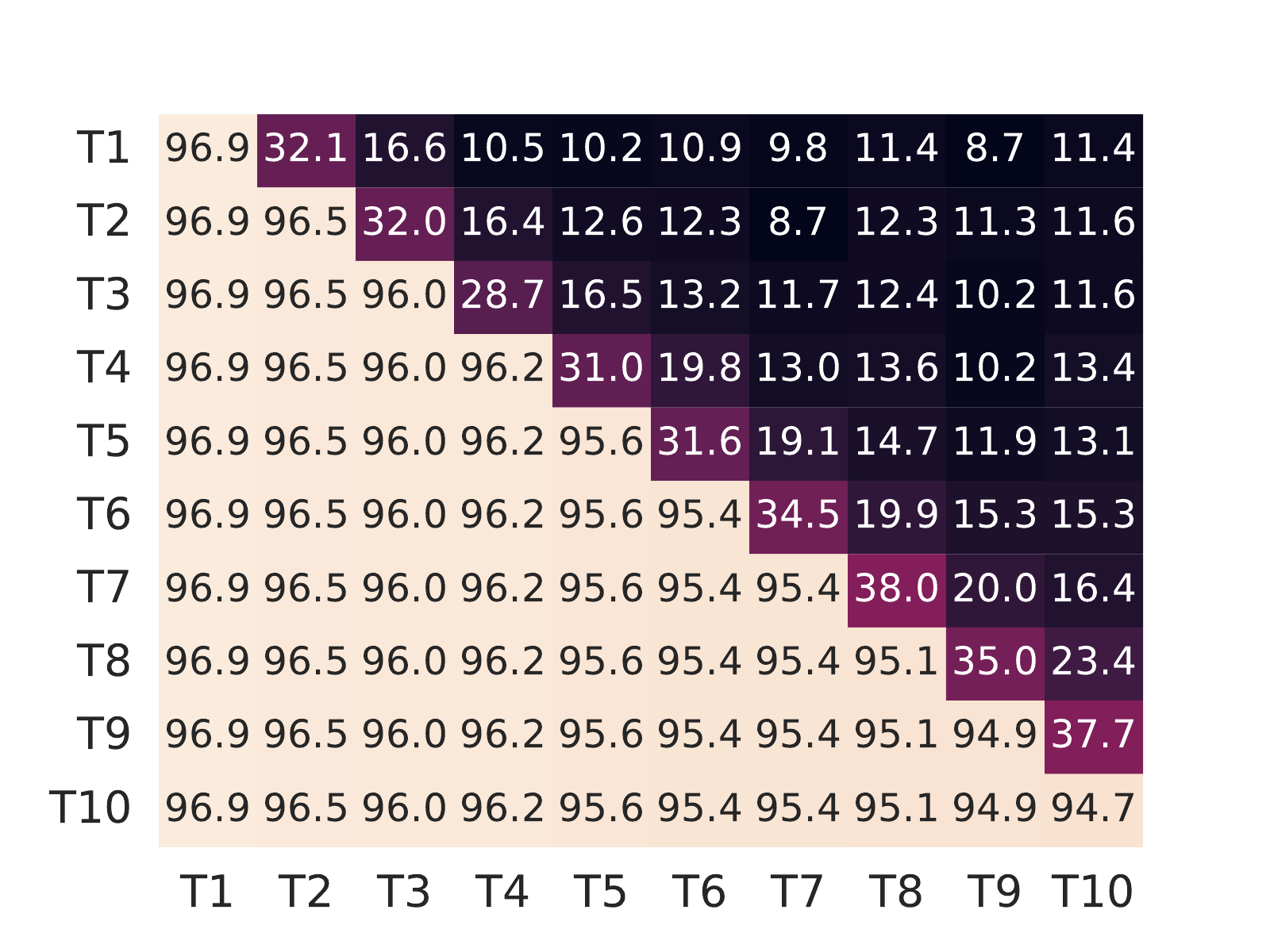} \\
    \small (a) WSN, $c = 5\%$ & \small (b) \textcolor{magenta}{SoftNet}, $c = 5\%$ \\
    \includegraphics[width=0.5\columnwidth]{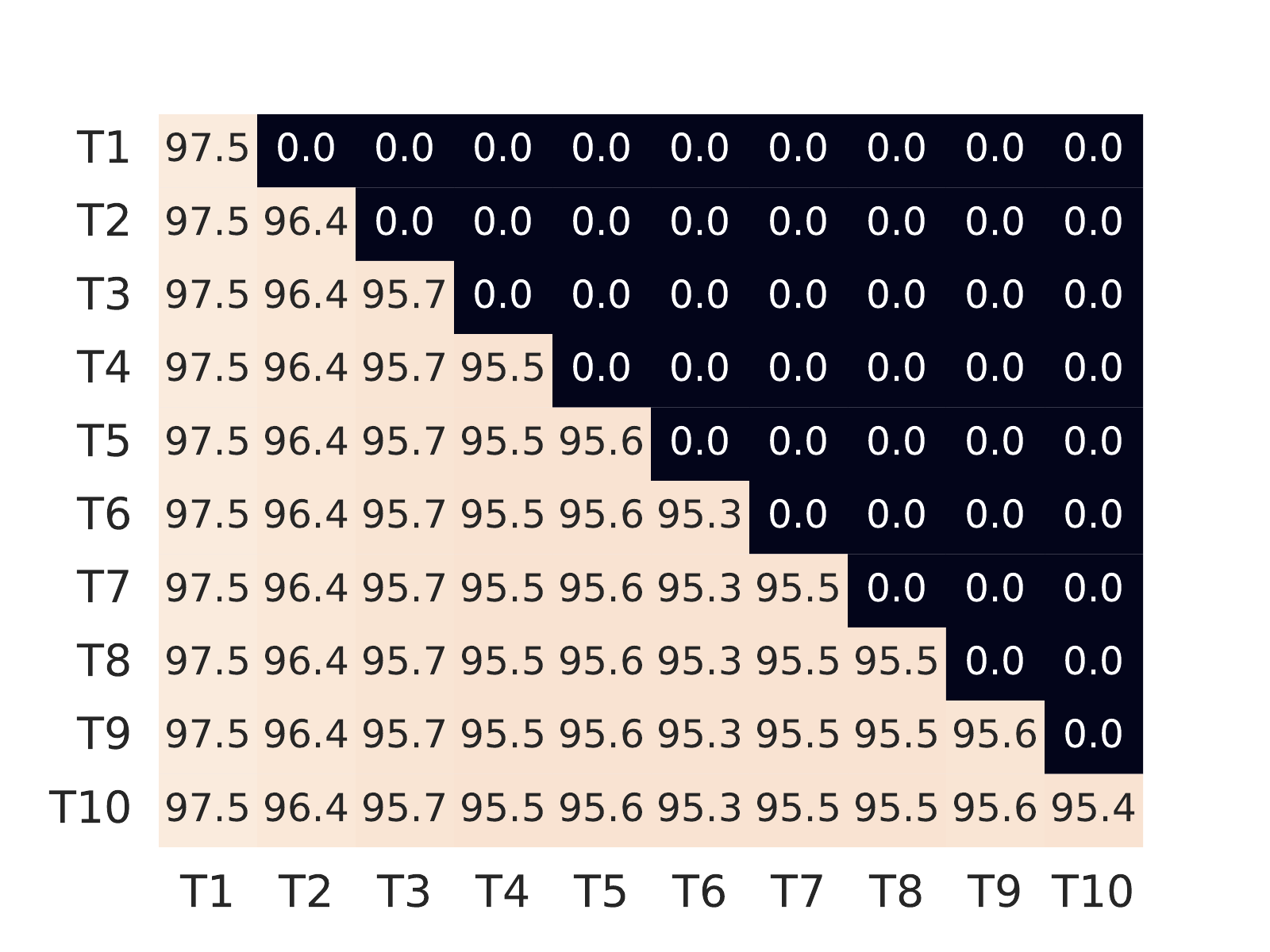} & 
    \includegraphics[width=0.5\columnwidth]{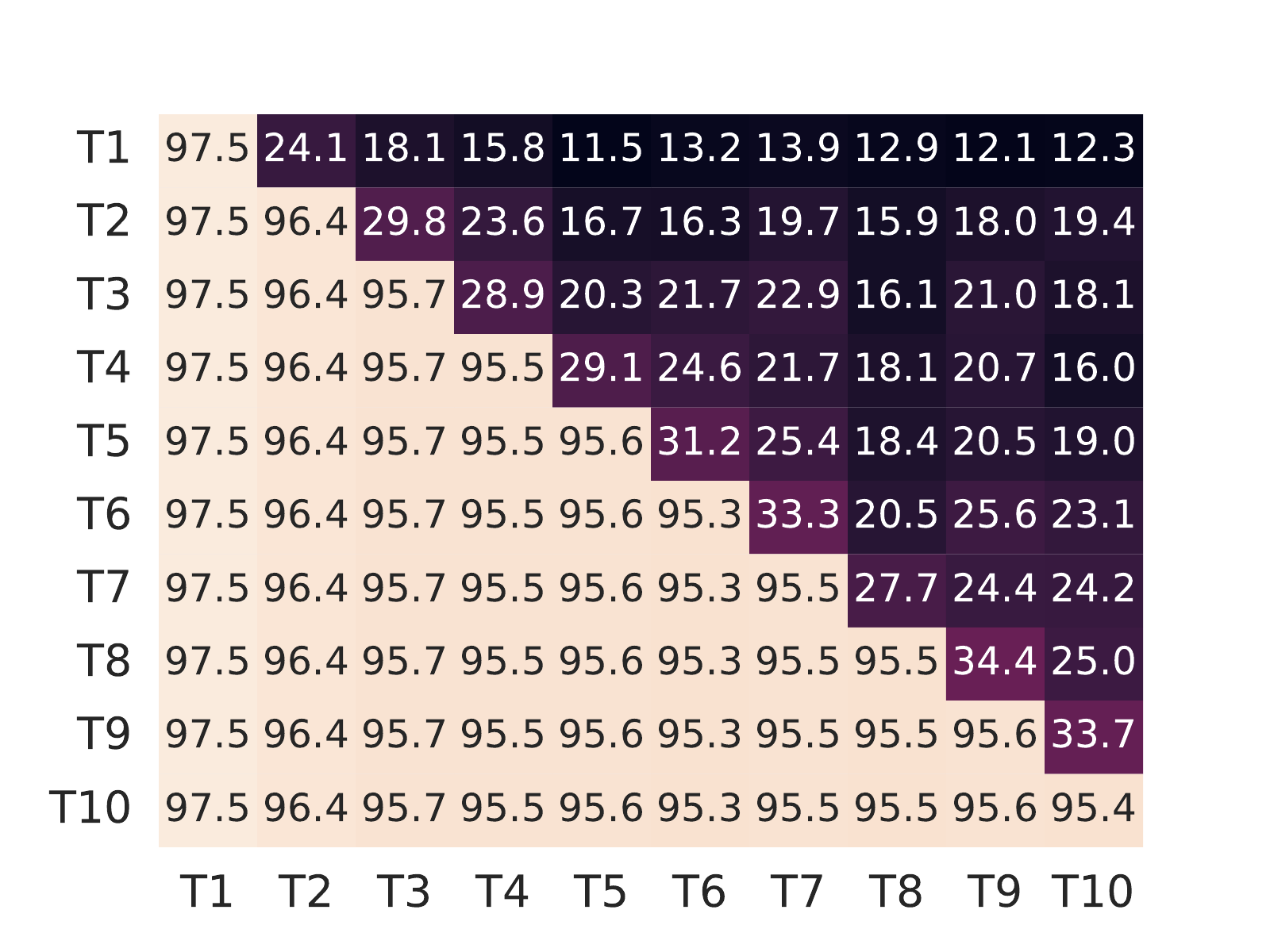} \\
    \small (c) WSN, $c = 70\%$  & \small (d) \textcolor{magenta}{SoftNet}, $c = 70\%$
    \end{tabular}}
    \caption{\textbf{Average Forward Transfer Matrix} on Permuted MNIST.
WSN / SoftNet with c = 5\% and c = 70\%, respectively}
    \label{fig:main_conf_hard_soft_wsn_pmnist}
\end{figure}
\begin{figure}[ht]
    \centering
    \vspace{-0.1in}
    \setlength{\tabcolsep}{0pt}{%
    \begin{tabular}{cc}
    \includegraphics[width=0.5\columnwidth]{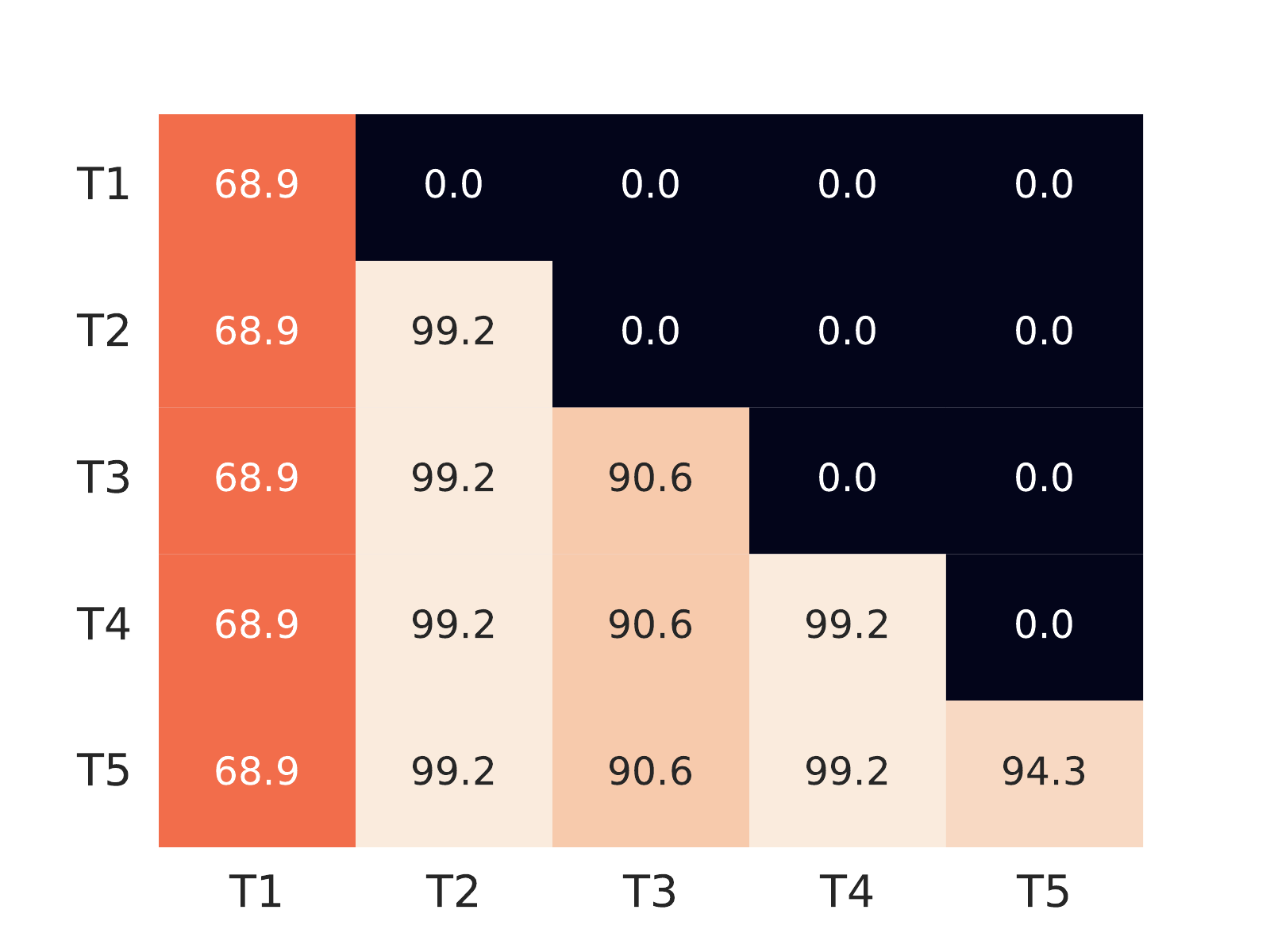} & 
    \includegraphics[width=0.5\columnwidth]{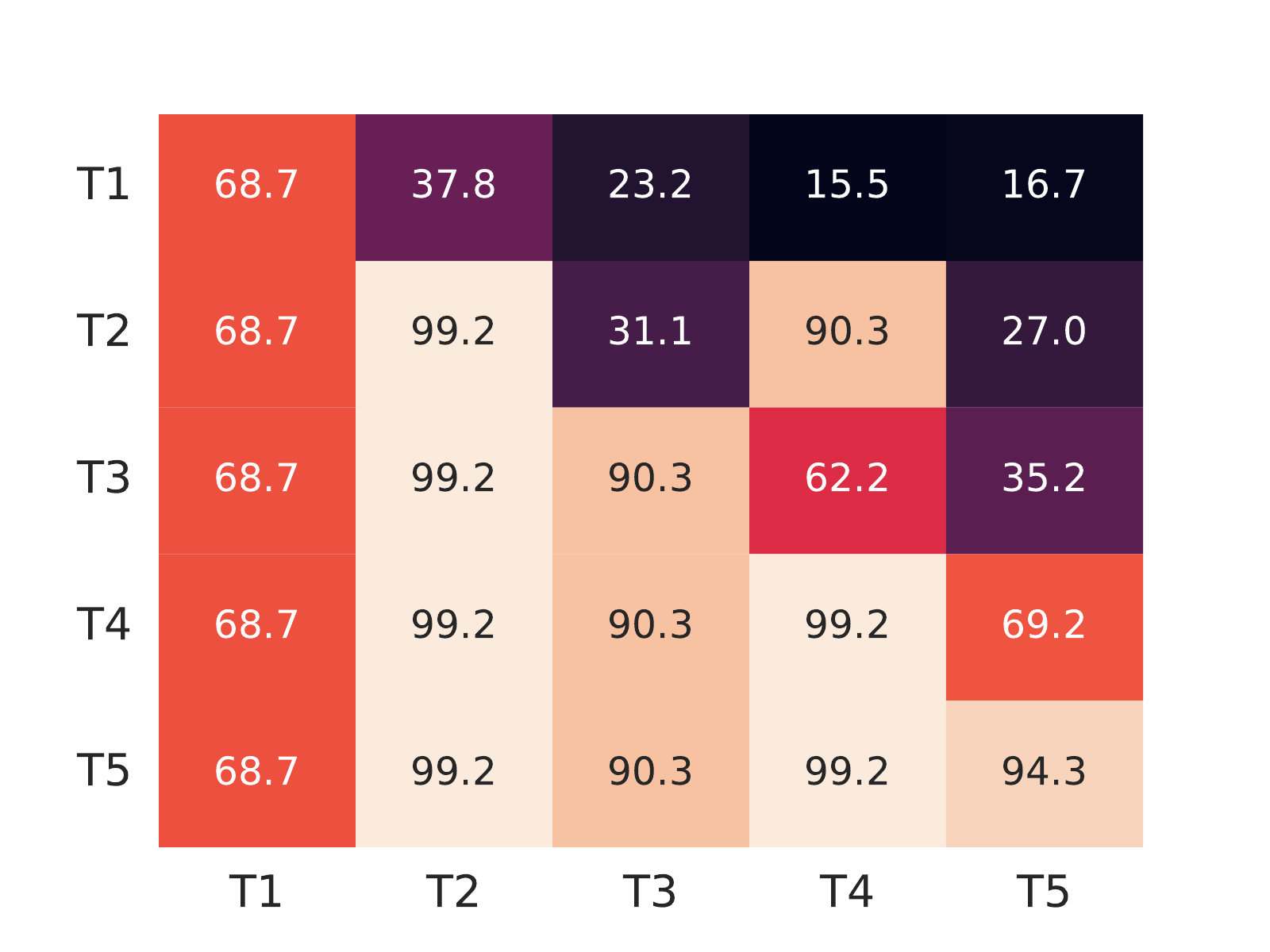} \\
    \small (a) WSN, $c = 5\%$ & \small (b) \textcolor{magenta}{SoftNet}, $c = 5\%$ \\
    \includegraphics[width=0.5\columnwidth]{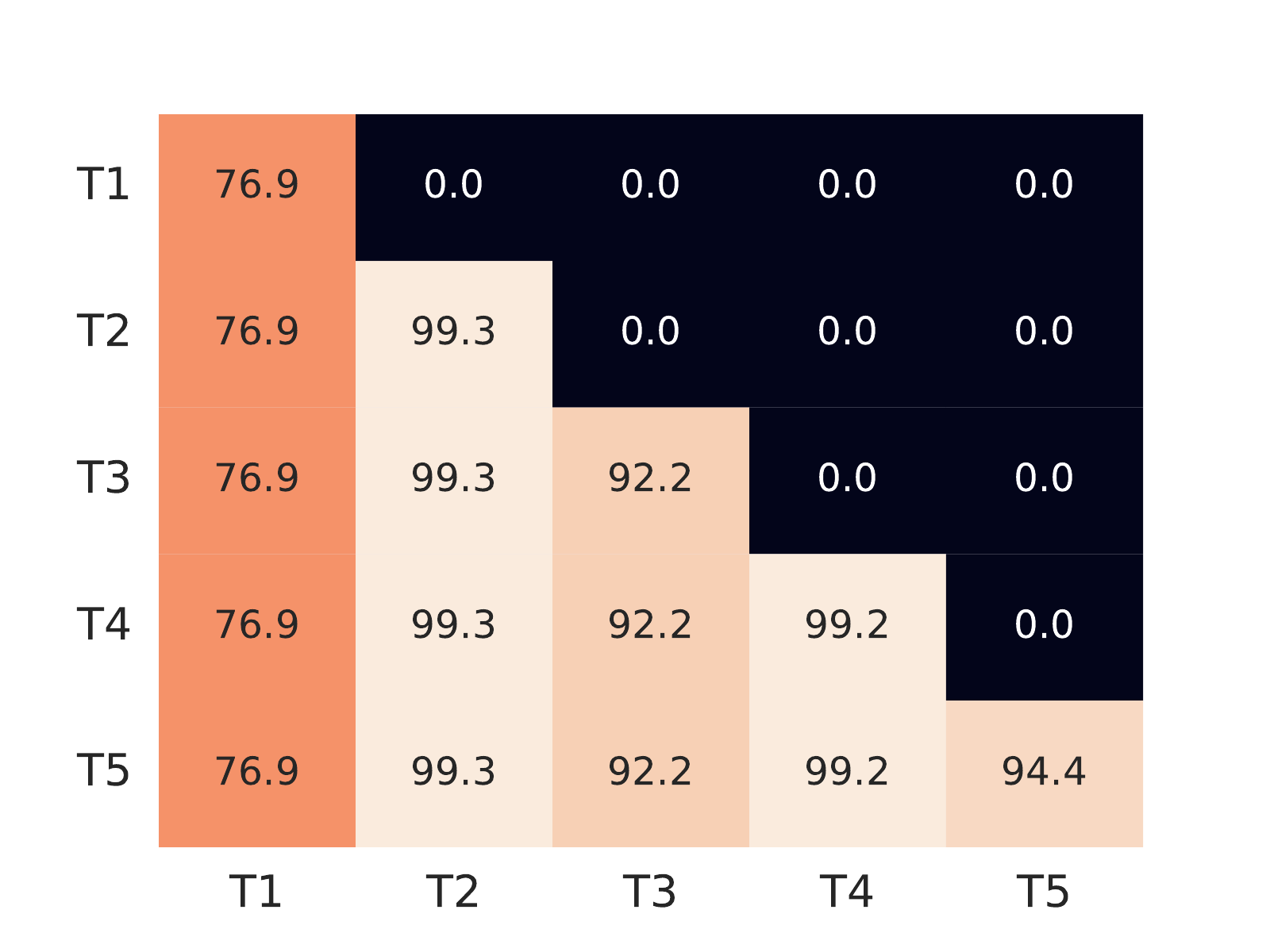} & 
    \includegraphics[width=0.5\columnwidth]{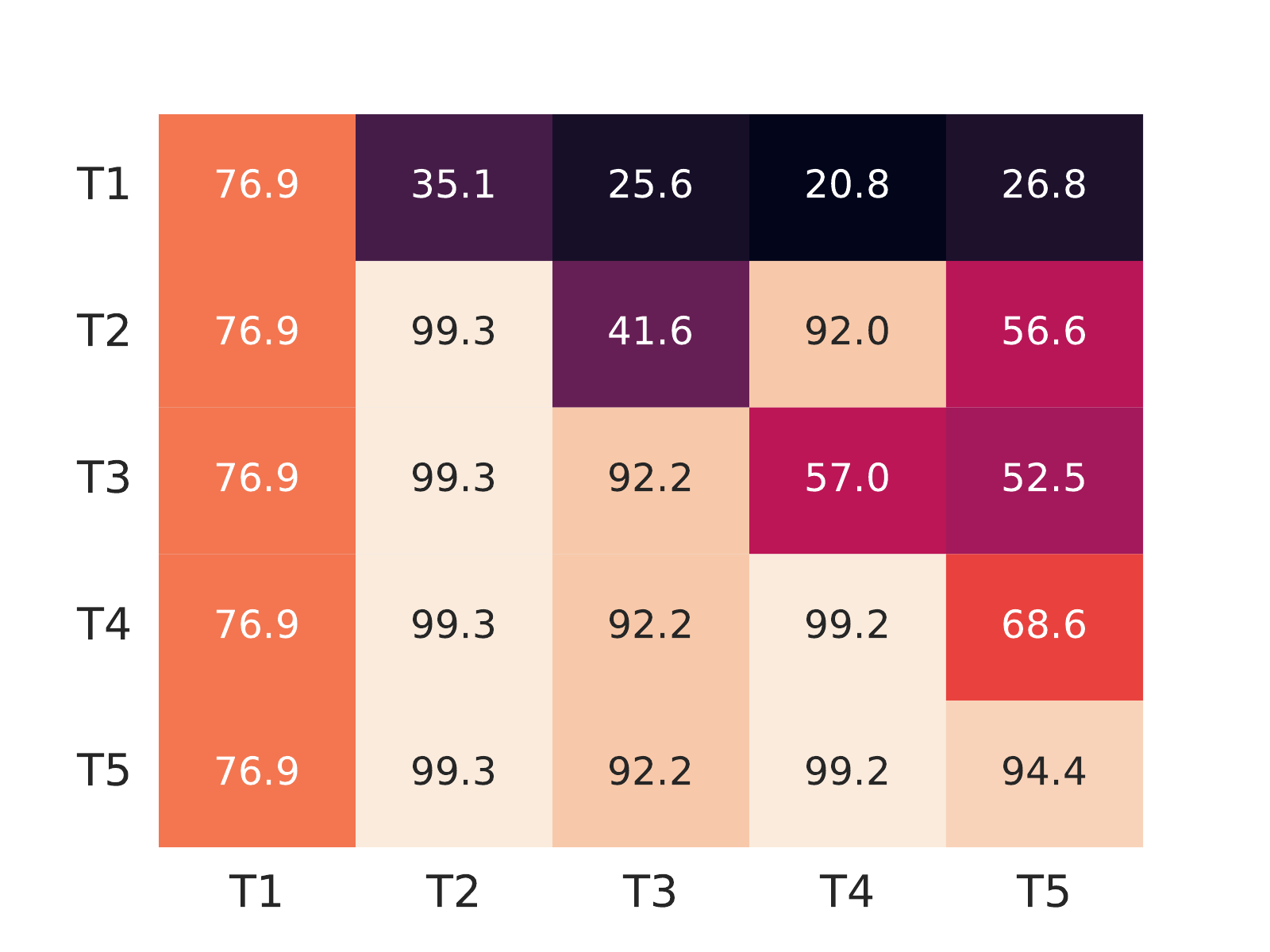} \\
    \small (c) WSN, $c = 70\%$  & \small (d) \textcolor{magenta}{SoftNet}, $c = 70\%$
    \end{tabular}}
    \caption{\textbf{Average Forward Transfer Matrix} on 5-Dataset.
WSN / SoftNet with c = 5\% and c = 70\%, respectively}
    \label{fig:main_conf_hard_soft_wsn_five}
    \vspace{-0.12in}
\end{figure}

\section{Experimental Details of SoftNet for FSCIL}\label{app_sec:exp_detail}


\subsection{Datasets} The following datasets are utilized for comparisons with current state-of-the-art:

\noindent 
\textbf{CIFAR-100} In CIFAR-100, each class contains $500$ images for training and $100$ images for testing. Each image has a size of $32\times 32$. Here, we follow an identical FSCIL procedure as in \cite{shi2021overcoming}, dividing the dataset into a base session with 60 base classes and eight novel sessions with a 5-way 5-shot problem on each session.

\noindent
\textbf{miniImageNet} miniImageNet consists of RGB images from 100 different classes, where each class contains $500$ training images and $100$ test images of size $84 \times 84$. Initially proposed for few-shot learning problems, miniImageNet is part of a much larger ImageNet dataset. Compared with CIFAR-100, the miniImageNet dataset is more complex and suitable for prototyping. The setup of miniImageNet is similar to that of CIFAR-100. To proceed with our evaluation, we follow the procedure described in \cite{shi2021overcoming}, incorporating 60 base classes and eight novel sessions through 5-way 5-shot problems. 

\noindent
\textbf{CUB-200-2011} CUB-200-2011 contains $200$ fine-grained bird species with $11,788$ images with varying images for each class. To proceed with experiments, we split the dataset into $6,000$ training images, and $6,000$ test images as in \cite{tao2020few}. During training, We randomly crop each image to size $224 \times 224$. We fix the first 100 classes as base classes, utilizing all samples in these respective classes to train the model. On the other hand, we treat the remaining 100 classes as novel categories split into ten novel sessions with a 10-way 5-shot problem in each session.

\subsection{Comparisons with SOTA}
We compare SoftNet with the following state-of-art-methods on TOPIC class split~\cite{tao2020few} of three benchmark datasets - CIFAR100 (\Cref{tab:main_cifar100_5way_5shot_resnet18_split}), miniImageNet (\Cref{tab:miniImageNet_5way_5shot_baseline_split}), and CUB-200-2011 (\Cref{tab:main_cub200_10way_5shot}).

\begin{table*}[ht]
\begin{center}
\caption{Classification accuracy of ResNet18 on CIFAR-100 for 5-way 5-shot incremental learning with the same class split as in TOPIC~\cite{cheraghian2021semantic}. $^\ast$ denotes the results reported from \cite{shi2021overcoming}. $^\dagger$ represents our reproduced results.}

\resizebox{0.8\textwidth}{!}{
\begin{tabular}{lcccccccccc}
\toprule
\multicolumn{1}{c}{\multirow{2}{*}{\textbf{Method}}}&\multicolumn{9}{c}{\textbf{sessions}}& \multicolumn{1}{c}{\multirow{2}{*}{\makecell{\textbf{The gap} \\ \textbf{with cRT}}}} \\ 
\cline{2-10}
& 1 & 2 & 3 & 4 & 5 & 6 & 7 & 8 & 9 &  \\
\midrule
cRT \cite{shi2021overcoming}$^\ast$ & 72.28 & 69.58 & 65.16 & 61.41 & 58.83 & 55.87 & 53.28 & 51.38 & 49.51 \\
\midrule
TOPIC \cite{cheraghian2021semantic} & 64.10 & 55.88 & 47.07 & 45.16 & 40.11 & 36.38 & 33.96 & 31.55 & 29.37 & -20.14 \\
CEC \cite{zhang2021few} & 73.07 & 68.88 & 65.26 & 61.19 & 58.09 & 55.57 & 53.22 & 51.34 & 49.14 & -0.37  \\
F2M \cite{shi2021overcoming} & 71.45 & 68.10 & 64.43 & 60.80 & 57.76 & 55.26 & 53.53 & 51.57 & 49.35 & -0.16 \\ 
LIMIT \cite{zhou2022few} & 73.81 & 72.09 & 67.87 & 63.89 & 60.70 & 57.77 & 55.67 & 53.52 & 51.23 & +1.72\\
MetaFSCIL \cite{chi2022metafscil} & 74.50 & 70.10 & 66.84 & 62.77 & 59.48 & 56.52 &  54.36 & 52.56 & 49.97 & +0.46 \\
ALICE \cite{peng2022few} & 79.00 & 70.50 & 67.10 & 63.40 & 61.20 & 59.20 & 58.10 & 56.30 & 54.10 & +4.59 \\
Entropy-Reg \cite{liu2022few} & 74.40 & 70.20 & 66.54 & 62.51 & 59.71 & 56.58 & 54.52 & 52.39 & 50.14 & +0.63 \\
C-FSCIL \cite{hersche2022constrained} & 77.50 & 72.45 & 67.94 & 63.80 & 60.24 & 57.34 & 54.61 & 52.41 & 50.23 & +0.72 \\
\midrule
FSLL \cite{mazumder2021few} & 64.10 & 55.85 & 51.71 & 48.59 & 45.34 & 43.25 & 41.52 & 39.81 & 38.16 & -11.35 \\

HardNet (WSN), $c=50\%$   
& 78.35	& 74.12 & 70.13 & 65.88 & 62.74 & 59.56 & 57.98 & 56.31 & 54.32 & +4.81 \\



HardNet (WSN), $c=80\%$   
& 79.27	& 75.38 & 71.11 & 66.68 & 63.32 & 60.06 & 58.16 & 56.40 & 54.31 & +4.80 \\





\midrule






\textcolor{cyan}{SoftNet}, ~~$c=50\%$   
& 79.88  & 75.54  & 71.64  & 67.47  & 64.45  & 61.09  & 59.07  & 57.29  & \textbf{55.33}  & \textbf{+5.82} \\

\textcolor{cyan}{SoftNet}, ~~$c=80\%$   
& \textbf{80.33} & \textbf{76.23}  & \textbf{72.19}  &  \textbf{67.83}  &  \textbf{64.64}  &  \textbf{61.39} & \textbf{59.32}  & \textbf{57.37} & 54.94  & +5.43 \\





\bottomrule
\end{tabular}
}
\label{tab:main_cifar100_5way_5shot_resnet18_split}
\end{center}
\end{table*}

\begin{table*}[ht]
\begin{center}
\caption{Classification accuracy of ResNet18 on miniImageNet for 5-way 5-shot incremental learning with the same class split as in TOPIC~\cite{cheraghian2021semantic}. $^\ast$ denotes results reported from \cite{shi2021overcoming}.}

\resizebox{0.8\textwidth}{!}{
\begin{tabular}{lcccccccccc}
\toprule
\multicolumn{1}{c}{\multirow{2}{*}{\textbf{Method}}}&\multicolumn{9}{c}{\textbf{sessions}}& \multicolumn{1}{c}{\multirow{2}{*}{\makecell{\textbf{The gap} \\ \textbf{with cRT}}}} \\ 
\cline{2-10}
& 1 & 2 & 3 & 4 & 5 & 6 & 7 & 8 & 9 &  \\
\midrule
cRT \cite{shi2021overcoming}$^\ast$ & 72.08 & 68.15 & 63.06 & 61.12 & 56.57 & 54.47 & 51.81 & 49.86 & 48.31 & -\\
\midrule
TOPIC \cite{cheraghian2021semantic} & 61.31 & 50.09 & 45.17 & 41.16 & 37.48 & 35.52 & 32.19 & 29.46 & 24.42 & -23.89 \\
IDLVQ-C \cite{chen2020incremental} & 64.77 & 59.87 & 55.93 & 52.62 & 49.88 & 47.55 & 44.83 & 43.14 & 41.84 & -6.47 \\
CEC \cite{zhang2021few} &  72.00 & 66.83 & 62.97 & 59.43 & 56.70 & 53.73 & 51.19 & 49.24 & 47.63 & -0.68 \\
F2M \cite{shi2021overcoming} & 72.05 & 67.47 & 63.16 & 59.70 & 56.71 & 53.77 & 51.11 & 49.21 & 47.84 & -0.43 \\ 
LIMIT \cite{zhou2022few} & 73.81 & 72.09 & 67.87 & 63.89 & 60.70 & 57.77 & 55.67 & 53.52 & 51.23 & +2.92  \\
MetaFSCIL \cite{chi2022metafscil} & 72.04 & 67.94 & 63.77 & 60.29 & 57.58 & 55.16 & 52.90 & 50.79 & 49.19 & +0.88 \\
ALICE \cite{peng2022few} & \textbf{80.60} & 70.60 & 67.40 & 64.50 & 62.50 & 60.00 & 57.80 & \textbf{56.80} & \textbf{55.70} & +\textbf{7.39}  \\
C-FSCIL \cite{hersche2022constrained} & 76.40 & 71.14 & 66.46 & 63.29 & 60.42 & 57.46 & 54.78 & 53.11 & 51.41 & +3.10 \\
Entropy-Reg \cite{liu2022few} & 71.84 & 67.12 & 63.21 & 59.77 & 57.01 & 53.95 & 51.55 & 49.52 & 48.21 & -0.10 \\
Subspace Reg. \cite{akyurek2021subspace} &  80.37 & 71.69 & 66.94 & 62.53 & 58.90 & 55.00 & 51.94 & 49.76 & 46.79 & -1.52\\
\midrule 
FSLL \cite{mazumder2021few} & 66.48 & 61.75 & 58.16 & 54.16 & 51.10 & 48.53 & 46.54 & 44.20 & 42.28 & -6.03 \\
HardNet (WSN), $c=80\%$  
& 78.70 & 72.55 & 68.26 & 64.45 & 61.74 & 58.93 & 55.99 & 54.09 & 52.74 & +4.43 \\
HardNet (WSN), $c=87\%$  
& 79.17 & 73.05 & 69.16 & 65.43 & 62.61 & 59.31 & 56.73 & 54.69 & 53.47 & +5.16 \\
HardNet (WSN), $c=90\%$  
& 79.15 & 72.03 & 68.76 & 65.32 & 62.00 & 58.21 & 56.52 & 53.66 & 53.07 & +4.76 \\
\midrule 


\textcolor{cyan}{SoftNet}, ~~$c=80\%$  
& 79.37 & 74.31  & 69.89  & 66.16  & 63.40  & 60.75  & 57.62  &  55.67 & 54.34  & +6.03  \\

\textcolor{cyan}{SoftNet}, ~~$c=87\%$  
& 79.77 & \textbf{75.08}  & \textbf{70.59}  & \textbf{66.93} & \textbf{64.00} & \textbf{61.00} & \textbf{57.81}  & 55.81  & 54.68  & +6.37 \\

\textcolor{cyan}{SoftNet}, ~~$c=90\%$  
& 79.72 & 74.25  & 70.00  & 66.35 &  63.19  & 60.04  & 57.36  & 55.38  & 54.14  & +5.83 \\

\bottomrule
\end{tabular}}
\label{tab:miniImageNet_5way_5shot_baseline_split}
\end{center}
\end{table*}

\begin{table*}[ht]
\begin{center}
\caption{Classification accuracy of ResNet18 on CUB-200-2011 for 10-way 5-shot incremental learning (TOPIC class split~\cite{tao2020few}). $^\ast$ denotes results reported from \cite{shi2021overcoming}. $^\dagger$ represents our reproduced results.}

\resizebox{0.96\textwidth}{!}{
\begin{tabular}{lcccccccccccc}
\toprule
\multicolumn{1}{c}{\multirow{2}{*}{\textbf{Method}}}&\multicolumn{11}{c}{\textbf{sessions}}& \multicolumn{1}{c}{\multirow{2}{*}{\thead{\textbf{The gap} \\ \textbf{with cRT}}}} \\ 
\cline{2-12}
& 1 & 2 & 3 & 4 & 5 & 6 & 7 & 8 & 9 & 10 & 11   \\
\midrule
cRT \cite{shi2021overcoming}$^\ast$ & 77.16 & 74.41 & 71.31 & 68.08 & 65.57 & 63.08 & 62.44 & 61.29 & 60.12 & 59.85 & 59.30 & - \\
\midrule
TOPIC \cite{cheraghian2021semantic} & 68.68 & 62.49 & 54.81 & 49.99 & 45.25 & 41.40 & 38.35 & 35.36 & 32.22 & 28.31 & 26.28 & -34.80 \\
SPPR \cite{zhu2021self} &  68.68 & 61.85 & 57.43 & 52.68 & 50.19 & 46.88 & 44.65 & 43.07 & 40.17 & 39.63 & 37.33 & -21.97 \\
CEC \cite{zhang2021few} & 75.85 & 71.94 & 68.50 & 63.50 & 62.43 & 58.27 & 57.73 & 55.81 & 54.83 & 53.52 & 52.28 & -7.02 \\
F2M \cite{shi2021overcoming} & 77.13 & 73.92 & 70.27 & 66.37 & 64.34 & 61.69 & 60.52 & 59.38 & 57.15 & 56.94 & 55.89 & -3.41 \\ 
LIMIT \cite{zhou2022few}  & 75.89 & 73.55 & \textbf{71.99} & \textbf{68.14} & \textbf{67.42} & \textbf{63.61} & 62.40 & 61.35 & 59.91 & 58.66 & 57.41 & -1.89 \\
MetaFSCIL \cite{chi2022metafscil} & 75.90 & 72.41 & 68.78 & 64.78 & 62.96 & 59.99 & 58.30 & 56.85 & 54.78 & 53.82 & 52.64 & -6.66 \\

ALICE \cite{peng2022few} & 77.40 & 72.70 & 70.60 & 67.20 & 65.90 & 63.40 & \textbf{62.90} & \textbf{61.90} & \textbf{60.50} & \textbf{60.60} & \textbf{60.10} & \textbf{-0.02} \\
Entropy-Reg \cite{liu2022few} & 75.90 & 72.14 & 68.64 & 63.76 & 62.58 & 59.11 & 57.82 & 55.89 & 54.92 & 53.58 & 52.39 & -6.91 \\
\midrule
FSLL \cite{mazumder2021few} & 72.77 & 69.33 & 65.51 & 62.66 & 61.10 & 58.65 & 57.78 & 57.26 & 55.59 & 55.39 & 54.21 & -6.87 \\
HardNet (WSN), $c=88\%$ & 76.89 & 73.40 & 69.77 & 66.15 &	64.00 & 60.98 &	59.56 &	58.05 &	56.05 &	55.84 &	55.20 &	-4.10 \\
HardNet (WSN), $c=90\%$ & 77.23 & 73.62 & 70.20 & 66.36 &	64.32 & 61.40 &	59.86 & 58.28 &	56.36 &	55.88 &	55.30 &	-4.00 \\
HardNet (WSN), $c=93\%$ & 77.76 & 73.97 & 70.41 & 66.60 &	64.47 & 61.35 &	59.80 &	58.18 &	56.17 &	55.73 &	55.18 &	-4.12 \\

\midrule

\textcolor{cyan}{SoftNet}, ~$c=88\%$ & \textbf{78.14} & \textbf{74.61} & 71.28 & 67.46 & 65.14 & 62.39 & 60.84 &	59.17 &	57.41 & 57.12 &	56.64 &	-2.66 \\
\textcolor{cyan}{SoftNet}, ~$c=90\%$ & 78.07 & 74.58 & 71.37 & 67.54 & 65.37 & 62.60 & 61.07 &	59.37 &	57.53 &	57.21 &	56.75 &	-2.55 \\
\textcolor{cyan}{SoftNet}, ~$c=93\%$ & 78.11 & 74.51 & 71.14 & 62.27 & 65.14 & 62.27 & 60.77 &	59.03 &	57.13 &	56.77 &	56.28 &	-3.02 \\

\bottomrule
\end{tabular}
}
\label{tab:main_cub200_10way_5shot}
\end{center}
\end{table*}

\end{document}